\documentclass{article}

 \usepackage[preprint,nonatbib]{neurips_2025}

\usepackage[utf8]{inputenc} 
\usepackage[T1]{fontenc}    
\usepackage{hyperref}       
\usepackage{url}            
\usepackage{booktabs}       
\usepackage{amsfonts}       
\usepackage{nicefrac}       
\usepackage{microtype}      
\usepackage[table]{xcolor}         
\usepackage{multirow,multicol}
\usepackage{amsmath}
\usepackage{tabularx}
\usepackage{caption}
\usepackage{subcaption}
\usepackage{wrapfig}
\usepackage[observedlatentcolor]{cv-dags}
\usepackage{enumitem}
\usepackage{soul}
\usepackage{xr}
\usepackage[backend=biber, defernumbers=true, style=nature, maxnames=99, maxcitenames=2, date=year, isbn=false, url=false, doi=true, natbib=true]{biblatex}
\newcommand{\revst}[1]{{\color{violet} \st{#1}}}

\renewcommand{\revst}[1]{}

\title{A Causal Framework for Aligning Image Quality Metrics and Deep Neural Network Robustness}

\author{%
  Nathan Drenkow$^{1,2, \ast}$ \\
  $^1$The Johns Hopkins University Applied Physics Laboratory \\
  Laurel, MD USA \\
  $^\ast$ e-mail: ndrenko1@jhu.edu \\
  \And 
  Mathias Unberath$^2$ \\
  $^2$The Johns Hopkins University \\
  Baltimore, MD USA   
}

\begin{document}

\newcommand{\figleft}{{\em (Left)}}
\newcommand{\figcenter}{{\em (Center)}}
\newcommand{\figright}{{\em (Right)}}
\newcommand{\figtop}{{\em (Top)}}
\newcommand{\figbottom}{{\em (Bottom)}}
\newcommand{\captiona}{{\em (a)}}
\newcommand{\captionb}{{\em (b)}}
\newcommand{\captionc}{{\em (c)}}
\newcommand{\captiond}{{\em (d)}}

\newcommand{\newterm}[1]{{\bf #1}}

\def\figref#1{figure~\ref{#1}}
\def\Figref#1{Figure~\ref{#1}}
\def\twofigref#1#2{figures \ref{#1} and \ref{#2}}
\def\quadfigref#1#2#3#4{figures \ref{#1}, \ref{#2}, \ref{#3} and \ref{#4}}
\def\secref#1{section~\ref{#1}}
\def\Secref#1{Section~\ref{#1}}
\def\twosecrefs#1#2{sections \ref{#1} and \ref{#2}}
\def\secrefs#1#2#3{sections \ref{#1}, \ref{#2} and \ref{#3}}
\def\eqref#1{equation~\ref{#1}}
\def\Eqref#1{Equation~\ref{#1}}
\def\plaineqref#1{\ref{#1}}
\def\chapref#1{chapter~\ref{#1}}
\def\Chapref#1{Chapter~\ref{#1}}
\def\rangechapref#1#2{chapters\ref{#1}--\ref{#2}}
\def\algref#1{algorithm~\ref{#1}}
\def\Algref#1{Algorithm~\ref{#1}}
\def\twoalgref#1#2{algorithms \ref{#1} and \ref{#2}}
\def\Twoalgref#1#2{Algorithms \ref{#1} and \ref{#2}}
\def\partref#1{part~\ref{#1}}
\def\Partref#1{Part~\ref{#1}}
\def\twopartref#1#2{parts \ref{#1} and \ref{#2}}

\def\ceil#1{\lceil #1 \rceil}
\def\floor#1{\lfloor #1 \rfloor}
\def\1{\bm{1}}
\newcommand{\train}{\mathcal{D}}
\newcommand{\valid}{\mathcal{D_{\mathrm{valid}}}}
\newcommand{\test}{\mathcal{D_{\mathrm{test}}}}

\def\eps{{\epsilon}}

\def\reta{{\textnormal{$\eta$}}}
\def\ra{{\textnormal{a}}}
\def\rb{{\textnormal{b}}}
\def\rc{{\textnormal{c}}}
\def\rd{{\textnormal{d}}}
\def\re{{\textnormal{e}}}
\def\rf{{\textnormal{f}}}
\def\rg{{\textnormal{g}}}
\def\rh{{\textnormal{h}}}
\def\ri{{\textnormal{i}}}
\def\rj{{\textnormal{j}}}
\def\rk{{\textnormal{k}}}
\def\rl{{\textnormal{l}}}
\def\rn{{\textnormal{n}}}
\def\ro{{\textnormal{o}}}
\def\rp{{\textnormal{p}}}
\def\rq{{\textnormal{q}}}
\def\rr{{\textnormal{r}}}
\def\rs{{\textnormal{s}}}
\def\rt{{\textnormal{t}}}
\def\ru{{\textnormal{u}}}
\def\rv{{\textnormal{v}}}
\def\rw{{\textnormal{w}}}
\def\rx{{\textnormal{x}}}
\def\ry{{\textnormal{y}}}
\def\rz{{\textnormal{z}}}

\def\rvepsilon{{\mathbf{\epsilon}}}
\def\rvtheta{{\mathbf{\theta}}}
\def\rva{{\mathbf{a}}}
\def\rvb{{\mathbf{b}}}
\def\rvc{{\mathbf{c}}}
\def\rvd{{\mathbf{d}}}
\def\rve{{\mathbf{e}}}
\def\rvf{{\mathbf{f}}}
\def\rvg{{\mathbf{g}}}
\def\rvh{{\mathbf{h}}}
\def\rvu{{\mathbf{i}}}
\def\rvj{{\mathbf{j}}}
\def\rvk{{\mathbf{k}}}
\def\rvl{{\mathbf{l}}}
\def\rvm{{\mathbf{m}}}
\def\rvn{{\mathbf{n}}}
\def\rvo{{\mathbf{o}}}
\def\rvp{{\mathbf{p}}}
\def\rvq{{\mathbf{q}}}
\def\rvr{{\mathbf{r}}}
\def\rvs{{\mathbf{s}}}
\def\rvt{{\mathbf{t}}}
\def\rvu{{\mathbf{u}}}
\def\rvv{{\mathbf{v}}}
\def\rvw{{\mathbf{w}}}
\def\rvx{{\mathbf{x}}}
\def\rvy{{\mathbf{y}}}
\def\rvz{{\mathbf{z}}}

\def\erva{{\textnormal{a}}}
\def\ervb{{\textnormal{b}}}
\def\ervc{{\textnormal{c}}}
\def\ervd{{\textnormal{d}}}
\def\erve{{\textnormal{e}}}
\def\ervf{{\textnormal{f}}}
\def\ervg{{\textnormal{g}}}
\def\ervh{{\textnormal{h}}}
\def\ervi{{\textnormal{i}}}
\def\ervj{{\textnormal{j}}}
\def\ervk{{\textnormal{k}}}
\def\ervl{{\textnormal{l}}}
\def\ervm{{\textnormal{m}}}
\def\ervn{{\textnormal{n}}}
\def\ervo{{\textnormal{o}}}
\def\ervp{{\textnormal{p}}}
\def\ervq{{\textnormal{q}}}
\def\ervr{{\textnormal{r}}}
\def\ervs{{\textnormal{s}}}
\def\ervt{{\textnormal{t}}}
\def\ervu{{\textnormal{u}}}
\def\ervv{{\textnormal{v}}}
\def\ervw{{\textnormal{w}}}
\def\ervx{{\textnormal{x}}}
\def\ervy{{\textnormal{y}}}
\def\ervz{{\textnormal{z}}}

\def\rmA{{\mathbf{A}}}
\def\rmB{{\mathbf{B}}}
\def\rmC{{\mathbf{C}}}
\def\rmD{{\mathbf{D}}}
\def\rmE{{\mathbf{E}}}
\def\rmF{{\mathbf{F}}}
\def\rmG{{\mathbf{G}}}
\def\rmH{{\mathbf{H}}}
\def\rmI{{\mathbf{I}}}
\def\rmJ{{\mathbf{J}}}
\def\rmK{{\mathbf{K}}}
\def\rmL{{\mathbf{L}}}
\def\rmM{{\mathbf{M}}}
\def\rmN{{\mathbf{N}}}
\def\rmO{{\mathbf{O}}}
\def\rmP{{\mathbf{P}}}
\def\rmQ{{\mathbf{Q}}}
\def\rmR{{\mathbf{R}}}
\def\rmS{{\mathbf{S}}}
\def\rmT{{\mathbf{T}}}
\def\rmU{{\mathbf{U}}}
\def\rmV{{\mathbf{V}}}
\def\rmW{{\mathbf{W}}}
\def\rmX{{\mathbf{X}}}
\def\rmY{{\mathbf{Y}}}
\def\rmZ{{\mathbf{Z}}}

\def\ermA{{\textnormal{A}}}
\def\ermB{{\textnormal{B}}}
\def\ermC{{\textnormal{C}}}
\def\ermD{{\textnormal{D}}}
\def\ermE{{\textnormal{E}}}
\def\ermF{{\textnormal{F}}}
\def\ermG{{\textnormal{G}}}
\def\ermH{{\textnormal{H}}}
\def\ermI{{\textnormal{I}}}
\def\ermJ{{\textnormal{J}}}
\def\ermK{{\textnormal{K}}}
\def\ermL{{\textnormal{L}}}
\def\ermM{{\textnormal{M}}}
\def\ermN{{\textnormal{N}}}
\def\ermO{{\textnormal{O}}}
\def\ermP{{\textnormal{P}}}
\def\ermQ{{\textnormal{Q}}}
\def\ermR{{\textnormal{R}}}
\def\ermS{{\textnormal{S}}}
\def\ermT{{\textnormal{T}}}
\def\ermU{{\textnormal{U}}}
\def\ermV{{\textnormal{V}}}
\def\ermW{{\textnormal{W}}}
\def\ermX{{\textnormal{X}}}
\def\ermY{{\textnormal{Y}}}
\def\ermZ{{\textnormal{Z}}}

\def\vzero{{\bm{0}}}
\def\vone{{\bm{1}}}
\def\vmu{{\bm{\mu}}}
\def\vtheta{{\bm{\theta}}}
\def\va{{\bm{a}}}
\def\vb{{\bm{b}}}
\def\vc{{\bm{c}}}
\def\vd{{\bm{d}}}
\def\ve{{\bm{e}}}
\def\vf{{\bm{f}}}
\def\vg{{\bm{g}}}
\def\vh{{\bm{h}}}
\def\vi{{\bm{i}}}
\def\vj{{\bm{j}}}
\def\vk{{\bm{k}}}
\def\vl{{\bm{l}}}
\def\vm{{\bm{m}}}
\def\vn{{\bm{n}}}
\def\vo{{\bm{o}}}
\def\vp{{\bm{p}}}
\def\vq{{\bm{q}}}
\def\vr{{\bm{r}}}
\def\vs{{\bm{s}}}
\def\vt{{\bm{t}}}
\def\vu{{\bm{u}}}
\def\vv{{\bm{v}}}
\def\vw{{\bm{w}}}
\def\vx{{\bm{x}}}
\def\vy{{\bm{y}}}
\def\vz{{\bm{z}}}

\def\evalpha{{\alpha}}
\def\evbeta{{\beta}}
\def\evepsilon{{\epsilon}}
\def\evlambda{{\lambda}}
\def\evomega{{\omega}}
\def\evmu{{\mu}}
\def\evpsi{{\psi}}
\def\evsigma{{\sigma}}
\def\evtheta{{\theta}}
\def\eva{{a}}
\def\evb{{b}}
\def\evc{{c}}
\def\evd{{d}}
\def\eve{{e}}
\def\evf{{f}}
\def\evg{{g}}
\def\evh{{h}}
\def\evi{{i}}
\def\evj{{j}}
\def\evk{{k}}
\def\evl{{l}}
\def\evm{{m}}
\def\evn{{n}}
\def\evo{{o}}
\def\evp{{p}}
\def\evq{{q}}
\def\evr{{r}}
\def\evs{{s}}
\def\evt{{t}}
\def\evu{{u}}
\def\evv{{v}}
\def\evw{{w}}
\def\evx{{x}}
\def\evy{{y}}
\def\evz{{z}}

\def\mA{{\bm{A}}}
\def\mB{{\bm{B}}}
\def\mC{{\bm{C}}}
\def\mD{{\bm{D}}}
\def\mE{{\bm{E}}}
\def\mF{{\bm{F}}}
\def\mG{{\bm{G}}}
\def\mH{{\bm{H}}}
\def\mI{{\bm{I}}}
\def\mJ{{\bm{J}}}
\def\mK{{\bm{K}}}
\def\mL{{\bm{L}}}
\def\mM{{\bm{M}}}
\def\mN{{\bm{N}}}
\def\mO{{\bm{O}}}
\def\mP{{\bm{P}}}
\def\mQ{{\bm{Q}}}
\def\mR{{\bm{R}}}
\def\mS{{\bm{S}}}
\def\mT{{\bm{T}}}
\def\mU{{\bm{U}}}
\def\mV{{\bm{V}}}
\def\mW{{\bm{W}}}
\def\mX{{\bm{X}}}
\def\mY{{\bm{Y}}}
\def\mZ{{\bm{Z}}}
\def\mBeta{{\bm{\beta}}}
\def\mPhi{{\bm{\Phi}}}
\def\mLambda{{\bm{\Lambda}}}
\def\mSigma{{\bm{\Sigma}}}

\newcommand{\tens}[1]{\bm{\mathsfit{#1}}}
\def\tA{{\tens{A}}}
\def\tB{{\tens{B}}}
\def\tC{{\tens{C}}}
\def\tD{{\tens{D}}}
\def\tE{{\tens{E}}}
\def\tF{{\tens{F}}}
\def\tG{{\tens{G}}}
\def\tH{{\tens{H}}}
\def\tI{{\tens{I}}}
\def\tJ{{\tens{J}}}
\def\tK{{\tens{K}}}
\def\tL{{\tens{L}}}
\def\tM{{\tens{M}}}
\def\tN{{\tens{N}}}
\def\tO{{\tens{O}}}
\def\tP{{\tens{P}}}
\def\tQ{{\tens{Q}}}
\def\tR{{\tens{R}}}
\def\tS{{\tens{S}}}
\def\tT{{\tens{T}}}
\def\tU{{\tens{U}}}
\def\tV{{\tens{V}}}
\def\tW{{\tens{W}}}
\def\tX{{\tens{X}}}
\def\tY{{\tens{Y}}}
\def\tZ{{\tens{Z}}}

\def\gA{{\mathcal{A}}}
\def\gB{{\mathcal{B}}}
\def\gC{{\mathcal{C}}}
\def\gD{{\mathcal{D}}}
\def\gE{{\mathcal{E}}}
\def\gF{{\mathcal{F}}}
\def\gG{{\mathcal{G}}}
\def\gH{{\mathcal{H}}}
\def\gI{{\mathcal{I}}}
\def\gJ{{\mathcal{J}}}
\def\gK{{\mathcal{K}}}
\def\gL{{\mathcal{L}}}
\def\gM{{\mathcal{M}}}
\def\gN{{\mathcal{N}}}
\def\gO{{\mathcal{O}}}
\def\gP{{\mathcal{P}}}
\def\gQ{{\mathcal{Q}}}
\def\gR{{\mathcal{R}}}
\def\gS{{\mathcal{S}}}
\def\gT{{\mathcal{T}}}
\def\gU{{\mathcal{U}}}
\def\gV{{\mathcal{V}}}
\def\gW{{\mathcal{W}}}
\def\gX{{\mathcal{X}}}
\def\gY{{\mathcal{Y}}}
\def\gZ{{\mathcal{Z}}}

\def\sA{{\mathbb{A}}}
\def\sB{{\mathbb{B}}}
\def\sC{{\mathbb{C}}}
\def\sD{{\mathbb{D}}}
\def\sF{{\mathbb{F}}}
\def\sG{{\mathbb{G}}}
\def\sH{{\mathbb{H}}}
\def\sI{{\mathbb{I}}}
\def\sJ{{\mathbb{J}}}
\def\sK{{\mathbb{K}}}
\def\sL{{\mathbb{L}}}
\def\sM{{\mathbb{M}}}
\def\sN{{\mathbb{N}}}
\def\sO{{\mathbb{O}}}
\def\sP{{\mathbb{P}}}
\def\sQ{{\mathbb{Q}}}
\def\sR{{\mathbb{R}}}
\def\sS{{\mathbb{S}}}
\def\sT{{\mathbb{T}}}
\def\sU{{\mathbb{U}}}
\def\sV{{\mathbb{V}}}
\def\sW{{\mathbb{W}}}
\def\sX{{\mathbb{X}}}
\def\sY{{\mathbb{Y}}}
\def\sZ{{\mathbb{Z}}}

\def\emLambda{{\Lambda}}
\def\emA{{A}}
\def\emB{{B}}
\def\emC{{C}}
\def\emD{{D}}
\def\emE{{E}}
\def\emF{{F}}
\def\emG{{G}}
\def\emH{{H}}
\def\emI{{I}}
\def\emJ{{J}}
\def\emK{{K}}
\def\emL{{L}}
\def\emM{{M}}
\def\emN{{N}}
\def\emO{{O}}
\def\emP{{P}}
\def\emQ{{Q}}
\def\emR{{R}}
\def\emS{{S}}
\def\emT{{T}}
\def\emU{{U}}
\def\emV{{V}}
\def\emW{{W}}
\def\emX{{X}}
\def\emY{{Y}}
\def\emZ{{Z}}
\def\emSigma{{\Sigma}}

\newcommand{\etens}[1]{\mathsfit{#1}}
\def\etLambda{{\etens{\Lambda}}}
\def\etA{{\etens{A}}}
\def\etB{{\etens{B}}}
\def\etC{{\etens{C}}}
\def\etD{{\etens{D}}}
\def\etE{{\etens{E}}}
\def\etF{{\etens{F}}}
\def\etG{{\etens{G}}}
\def\etH{{\etens{H}}}
\def\etI{{\etens{I}}}
\def\etJ{{\etens{J}}}
\def\etK{{\etens{K}}}
\def\etL{{\etens{L}}}
\def\etM{{\etens{M}}}
\def\etN{{\etens{N}}}
\def\etO{{\etens{O}}}
\def\etP{{\etens{P}}}
\def\etQ{{\etens{Q}}}
\def\etR{{\etens{R}}}
\def\etS{{\etens{S}}}
\def\etT{{\etens{T}}}
\def\etU{{\etens{U}}}
\def\etV{{\etens{V}}}
\def\etW{{\etens{W}}}
\def\etX{{\etens{X}}}
\def\etY{{\etens{Y}}}
\def\etZ{{\etens{Z}}}

\newcommand{\pdata}{p_{\rm{data}}}
\newcommand{\ptrain}{\hat{p}_{\rm{data}}}
\newcommand{\Ptrain}{\hat{P}_{\rm{data}}}
\newcommand{\pmodel}{p_{\rm{model}}}
\newcommand{\Pmodel}{P_{\rm{model}}}
\newcommand{\ptildemodel}{\tilde{p}_{\rm{model}}}
\newcommand{\pencode}{p_{\rm{encoder}}}
\newcommand{\pdecode}{p_{\rm{decoder}}}
\newcommand{\precons}{p_{\rm{reconstruct}}}

\newcommand{\laplace}{\mathrm{Laplace}} 

\newcommand{\E}{\mathbb{E}}
\newcommand{\Ls}{\mathcal{L}}
\newcommand{\R}{\mathbb{R}}
\newcommand{\emp}{\tilde{p}}
\newcommand{\lr}{\alpha}
\newcommand{\reg}{\lambda}
\newcommand{\rect}{\mathrm{rectifier}}
\newcommand{\softmax}{\mathrm{softmax}}
\newcommand{\sigmoid}{\sigma}
\newcommand{\softplus}{\zeta}
\newcommand{\KL}{D_{\mathrm{KL}}}
\newcommand{\Var}{\mathrm{Var}}
\newcommand{\standarderror}{\mathrm{SE}}
\newcommand{\Cov}{\mathrm{Cov}}
\newcommand{\normlzero}{L^0}
\newcommand{\normlone}{L^1}
\newcommand{\normltwo}{L^2}
\newcommand{\normlp}{L^p}
\newcommand{\normmax}{L^\infty}

\newcommand{\parents}{Pa} 

\newcommand{\argmax}{arg\,max}
\newcommand{\argmin}{arg\,min}

\newcommand{\sign}{sign}
\newcommand{\Tr}{Tr}
\let\ab\allowbreak

\newcommand{\ie}{i.\,e.,~}
\newcommand{\eg}{e.\,g.,~}

\maketitle

\begin{abstract}
    Image quality plays an important role in the performance of deep neural networks (DNNs) that have been widely shown to exhibit sensitivity to changes in imaging conditions. 
    Conventional image quality assessment (IQA) seeks to measure and align quality relative to human perceptual judgments, but we often need a metric that is not only sensitive to imaging conditions but also well-aligned with DNN sensitivities.
    We first ask whether conventional IQA metrics are also informative of DNN performance. 
    We show theoretically and empirically that conventional IQA metrics are weak predictors of DNN performance for image classification. 
    Using our causal framework, we then develop metrics that exhibit strong correlation with DNN performance, thus enabling us to effectively estimate the quality distribution of large image datasets relative to targeted vision tasks.
\end{abstract}

\section{Introduction}
\label{sec:intro}
Ensuring the robustness of deep neural networks (DNNs) to real-world imaging conditions is crucial for safety- and cost-critical applications. Extensive research has shown that DNNs remain sensitive to natural distortions~\citep{Taori2020-xa, Djolonga2021-eo, Ibrahim2022-lk, Geirhos2021-bp} despite efforts to close the gap between performance on clean and naturally-distorted images.  While much effort has focused primarily on the design and optimization of robust DNNs, there is now growing interest in developing a deeper understanding of how the properties of the image data itself influence robustness during training and evaluation~\citep{Ilyas2022-eh, Lin2022-sn, Pavlak2023-vh, Drenkow2025-xx}.  

Since image quality is known to influence DNN behavior, a first step in analyzing image data is to examine the relationship between image quality (IQ) and DNN robustness. Here \textit{quality} describes the absence of distortion but more generally relates to the ability to extract task-relevant information from the image. Image \textit{quality} and \textit{difficulty} are closely related where quality measures properties of the imaging conditions while difficulty involves content and composition in addition to the conditions.
While image content and quality remain intimately related, prior work in evaluating DNN robustness~\cite{Hendrycks2019-ye, Dodge2016-fw, Laugros2019-lv, Drenkow2024-gl, Drenkow2024-hh} has shown that changes in image quality can result in diverse DNN behavior even when content is held fixed.  For this reason, examining the relationship between IQ metrics and DNN behavior remains an important question that is theoretically and empirically unanswered prior to this work.  

In an ideal case, image quality metrics will strongly correlate with task DNN performance  while remaining relatively independent of knowledge or assumptions about specific downstream task models that will consume the data. 
Despite decades of research into IQ metrics~\cite{wang2004image, zhang2011fsim, Xu2017-rp, Agnolucci2024-ti}, we examine for the first time in-depth the explicit connection between IQ metrics and task DNN performance.  
Due to the lack of prior theoretical work in this area, we first analyze this link empirically. Then, \textit{the primary goal of this work is to provide a mathematical framework for identifying and analyzing the conditions under which IQ metrics and DNN performance are correlated.} 

In providing a mathematical means to establish a link between IQ and DNN performance, our method enables new pathways for quantitatively analyzing dataset composition. Using IQ metrics as a proxy for DNN performance, we can estimate the quality distribution of datasets to understand how the range of ``\textit{easy}'' to ``\textit{hard}'' images influences observed DNN robustness.  This becomes increasingly useful as image datasets continue to grow in size to the extent that the cost and feasibility of using human annotators to assess and annotate properties of each data point is becoming intractable.  With pre-training datasets for foundation models and other large-scale vision models approaching hundreds of millions to billions of images~\citep{Radford2021-sr, Sun2017-nn, Schuhmann2022-ea}, new automated methods for analysis are needed for quantitatively assessing dataset composition. 

In this work, we focus specifically on \textit{natural robustness} which considers how images are distorted due to real-world factors such as lighting, weather, sensor settings, and/or motion. Image quality assessment (IQA) metrics have been developed over several decades of research~\citep{Wang2022-vx, Xu2017-rp, zhang2011fsim, wang2004image, Agnolucci2024-ti, Mittal2013-dq, Ye2012-iu, Xu2016-wz, Zhang2019-gz} and provide quantitative measures of quality calibrated with respect to human perceptual judgments. To the best of our knowledge, little work has been done to understand how these IQA metrics can help relate image difficulty and DNN performance.  To make this connection explicit, we state our primary research question: \textbf{What is the extent of the relationship between IQ and DNN performance metrics?} 

Our primary motivation is to identify image quality metrics that allow us to assess the distribution of image quality in large-scale datasets and establish quality-driven priors for DNN performance independent of any specific trained task models.  We propose the following desiderata for IQ metrics towards achieving these objectives.

\begin{itemize}[leftmargin=*]
    \item \textbf{D1 - Sensitive}: IQ metrics should be sufficiently sensitive to changes in image conditions
    \item \textbf{D2 - Blind}: IQ metrics should work in No Reference IQA (NR-IQA) settings where images are assessed \textit{without} knowledge of a reference image captured under ``clean'' conditions
    \item \textbf{D3 - Predictive}: IQ metrics should be correlated with DNN task performance 
    \item \textbf{D4 - Task Model Agnostic}: IQ metrics should be designed/trained/calibrated without \textit{a priori} knowledge of the downstream DNN models/architectures to be trained or evaluated on the data under consideration
\end{itemize}

The first criterion (\textbf{D1}) is a baseline condition requiring that the metric is actually sensitive to the natural conditions likely in the imaging domain. \textbf{D2} operates under the assumption that real-world datasets will not consist of pairs of clean/distorted images and will instead contain images collected in diverse conditions. \textbf{D3} stems from the idea that quality metrics should measure general properties of the data that influence performance metrics (e.g., if the quality metric decreases, then the performance metric should also decrease, although not necessarily at the same rate).  Lastly, \textbf{D4} comes from the desire to use image quality to assess the composition of the dataset \textit{independent} of any task-specific model training and without making assumptions about the type of DNN to be trained downstream. In other words, we want to avoid IQ metrics that are biased towards specific task models and/or require pre-training on each dataset to be analyzed.  

To determine the extent to which IQA metrics satisfy the above desiderata, our work makes the following contributions:
\begin{itemize}[leftmargin=*]
    \setlength\itemsep{0pt}
    \item Our primary contribution is a causal framework for analyzing the relationship between image quality and DNN performance in a range of IQA settings
    \item We use the framework to establish theoretically and empirically the independence of image quality and DNN performance under general conditions
    \item We identify specific conditions under which IQA metrics can be predictive of DNN performance
    \item We conduct a first-of-its-kind evaluation of the relation between conventional IQ metrics and DNN performance and find that conventional metrics are weakly predictive of performance
    \item We use the framework in the context of image classification tasks to develop a new task-guided IQA metric that enables quantitative assessments of image quality that are strongly predictive of downstream DNN task performance
    \item We show that our framework and novel metric provide the means to expose subtle differences in dataset quality distributions that directly impact on DNN performance
\end{itemize}

\section{Results}
\label{sec:results}

\subsection{Experiment setup}
\label{sub:exp-setup}
Given the causal interpretation of IQA and the IQ metric desiderata (\textbf{D1-D4}), we examine how conventional NR-IQA metrics relate to DNN performance.  Our primary hypothesis is that if quality ($Q$) and task model performance ($M$) metrics are sensitive to a common set of visual features $Z$ derived from images $X$ (Fig.~\ref{fig:latents-dag}), then we should observe that $Q$ is correlated with $M$ and even predictive of $M$ given $X$. Plainly stated, if image quality is high in general, then DNN performance should be similarly high (and vice versa). 

We focus the following experiments on image classification tasks since they have available benchmark datasets and have been well-studied within the deep learning field. We show how our framework can be used to identify the relationship between IQA metrics and DNN performance as well as how it can guide the development of new metrics that satisfy all desiderata.  For image classification, our experiments show that common NR-IQA methods are very weakly predictive of DNN performance, and while they satisfy \textbf{D1, D2, D4} of our IQ desiderata, they fail to satisfy \textbf{D3} and may not be suitable for estimating priors on DNN performance.

In the experiments in this section we use the following basic setup.  In order to have precise control and knowledge of the type and severity of image distortion, we use the ImageNet validation (IN-val) and ImageNet-C (IN-C)~\citep{Hendrycks2019-ye} datasets for evaluating IQ/performance on clean and corrupted images respectively.  For reference, we provide a common corruptions causal Directed Acyclic Graph (DAG) in Appendix~\ref{app:common-corruptions} for comparison with the ones in Figures~\ref{fig:latents-dag} and~\ref{fig:standard-dag}.

For each experiment, we compute the IQ metric ($Q$) and DNN correctness ($M$) for each image of the IN-C evaluation dataset.  We use the following common and high-performing NR-IQA metrics ($Q$): CLIP-IQA~\citep{Wang2022-vx}, ARNIQA~\citep{Agnolucci2024-ti}, BRISQUE~\citep{mittal2012no}, and Total Variation (TV). 
Here, CLIP-IQA and ARNIQA represent the state-of-the-art in learning-based IQA metrics while BRISQUE and TV represent conventional non-deep learning baselines.  
For DNNs, we evaluate the correctness ($M$) using pretrained ResNet34~\citep{he2016deep}, ConvNext-B~\citep{Liu2022-os}, EfficientNet-V2-M~\cite{Tan2021-xf}, MobileNet-V3-L~\cite{Howard2019-ds}, Vision Transformers~\cite{Dosovitskiy2020-tb}, and Swin-B~\citep{Liu2021-pj} models provided via the \verb|torchvision| 
package~\citep{Marcel2010-er}. 
This set of models covers a wide a range of architectural design, scale, efficiency, and performance characteristics in order to capture potential variability in the relationship between DNN performance and IQ metrics.
Across all experiments, 95\% confidence intervals (CI) are obtained via bootstrapping with 1000 resamples.

\subsection{Correlation and Predictability of $Q,~M$ (\textbf{D3})}  \label{sec:exp1}

\begin{figure}[h!]
    \centering
    \includegraphics[width=0.9\linewidth]{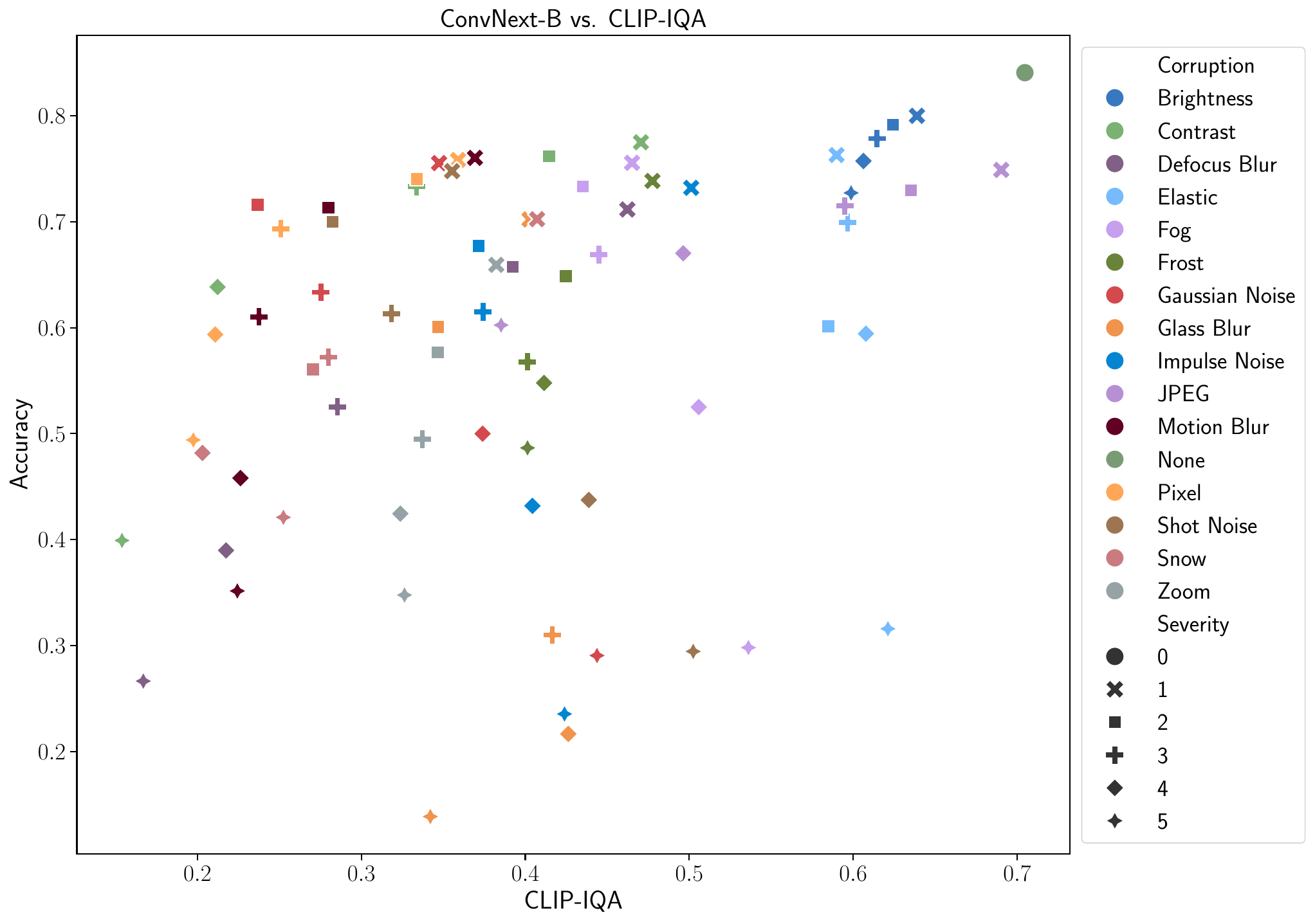}
    \caption{\textbf{Relationship between conventional NR-IQA metric and classification accuracy.} Accuracy ($M$) vs. IQ ($Q$) for ConvNext-B and CLIP-IQA respectively.  Each point represents the average accuracy over all images in the ImageNet val set corrupted with the corresponding corruption/severity. Little correlation is observed between $M, Q$ across all corruptions/severities and $Q$ is weakly predictive of $M$.}
    \label{fig:acc-vs-iq}
\end{figure}

We start by examining the correlation between $Q, M$ for NR-IQA metrics.  Figure~\ref{fig:acc-vs-iq} shows the general relationship between $Q, M$ where each point in the figure is the average accuracy (over 50k images) for each corruption and severity in IN-C.  Similarly, Table~\ref{tab:acc-iq} computes the Kendall Rank Correlation Coefficient (KRCC), Spearman Rank Correlation Coefficient (SRCC), and Pearson Linear Correlation Coefficient (PLCC) between IQ and average accuracy across all corruption/severity pairs (75 total).

These results provide a look at the group-wise association between $Q, M$ where the groups capture general trends in performance/IQ based on corruption type and severity.  The low correlation between $Q, M$ suggests that these NR-IQA metrics likely fall under the model described by Figure~\ref{fig:standard-dag} where $Q, M$ are conditionally independent given $X$.

We also examine the point-wise relationship between $Q, M$.  We aggregate DNN predictions and IQ values for all images in IN-C across all corruptions/severities and then randomly split the dataset (by image ID) into 80\% training and 20\% testing.  We train a logistic regression classifier to predict $P(M|Q)$ and test on the hold-out set. We measure the predictability of $M$ using Area Under the Curve (AUC) and average cross-entropy (CE).

Table~\ref{tab:acc-iq} shows that at the per-image level, $Q$ is still weakly predictive of $M$ (i.e., AUC $\approx 0.5$). This result is consistent with the theoretical analysis in Sec.~\ref{sub:base-iqa} and the weak correlation observed empirically between $Q, M$ measured at the group level. 

While the causal DAG in Figure~\ref{fig:standard-dag} would suggest that conditioning on the class label $Y$ should not change the result, we test this empirically as follows. 
We re-run the logistic regression for each label value in $\gY$ separately (1000 total) and compute the mean AUC ($mAUC$) and CE ($mCE$) across all labels.  While we observe some variability in results when fixing $Y$, we find $mAUC=0.5652~(\sigma=0.08)$ and $mCE=0.6176~(\sigma=0.1094)$ suggesting that even when we control for $Y$, the predictability of DNN performance from the NR-IQA metrics remains weak. 
More detailed results from experiments controlling for image content can be found in Appendix~\ref{app:control-for-content}.  

These results suggest that NR-IQA metrics are likely sensitive to a different set of image features than task DNNs (i.e., no shared $Z$) and thus are barely, if at all, predictive of performance (i.e., they do not satisfy criterion \textbf{D3}).  
In particular, Table~\ref{tab:acc-iq} shows that the AUC for predicting DNN performance given conventional NR-IQA metrics is close to chance (i.e., $0.5$). We also observe that this result holds across a range of NR-IQA metrics including those that are minimally tunable (TV, BRISQUE) to optimized through large-scale training (CLIP-IQA, ARNIQA), but all of which are calibrated against human judgments. 
This provides strong evidence that this form of calibration is a likely cause of the lack of correlation with DNN performance.  
\textit{The primary implication of this result is that if we intend to use IQ metrics to measure image quality/difficulty from the DNN perspective, common NR-IQA metrics may not be well-suited to this task and alternative approaches are needed.}

Lastly, we acknowledge that estimating image quality independent of the image content remains a challenging task. In our experiment, we found that by holding image- and class-level content fixed (Appendix~\ref{app:control-for-content}), we find that NR-IQA metrics are still weakly predictive of task DNN performance.  While this does not completely disentangle the quality/content relationship, it provides clear evidence that the relationship between conventional NR-IQA metrics is weak. 

\subsection{Restoring the association between $Q, M$ via strong task-guidance (\textbf{D3})}
\label{sec:strong-tg-iqa}

The previous results indicated that existing NR-IQA meet desiderata \textbf{D1, D2, D4}, but the lack of predictability (\textbf{D3}) between NR-IQA metrics and DNN accuracy/correctness is a major limitation in using these metrics for assessing dataset quality relative to potential downstream task models.
Focusing specifically on \textbf{D3}, we next consider an alternative formulation of the causal model that will allow us to recover a dependence between $M, Q$ when conditioning on $X$.  

\begin{figure}[h!]
    \centering
    \resizebox{0.65\linewidth}{!}{%
    \includegraphics[width=\linewidth]{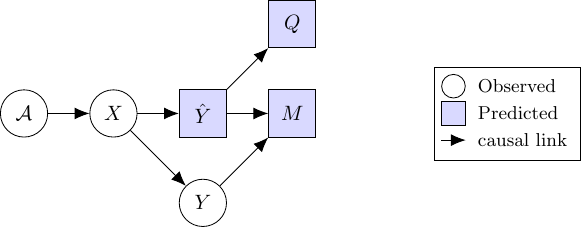}
    }%
    \caption{\textbf{Strong task-guided IQA.} Causal diagram relating model accuracy ($M$) with IQ metrics ($Q$) with the additional dependence $\hat{Y} \rightarrow Q$.}
    \label{fig:preds-dag}
\end{figure}

In the case where a pre-trained DNN $f_\theta$ is given, Figure~\ref{fig:preds-dag} describes a scenario where the predictions from this DNN may also be used as indicators of quality.  This parallels other work~\citep{Hendrycks2019-rw} which shows that uncertainty in the output predictions is often a good predictor of the OOD nature of the input.  Note here that while $Q, M$ both depend on $\hat{Y}$, $Q$ requires no knowledge of the labels.  In this case, it is possible that $\hat{Y}$ can be incorrect from the perspective of the ground truth label $Y$ but still provide information about $Q$ (e.g., via a low confidence prediction).  

Because this approach uses a model for $Q$ that is already trained for the classification task, we consider this \textbf{strong} task-guided IQA (TG-IQA).  Clearly, this provides an alternative to the conventional NR-IQA metrics but now violates \textbf{D4} since $Q$ is informed directly by the same model trained for the task and measured by $M$.  Nonetheless, our (temporary) goal here is to use the causal framework to show there exists a case where $Q, M$ are associated through a common set of features $Z$.  Our hypothesis is that with \textbf{strong} TG-IQA we should observe a clear correlation between $M, Q$.

We examine the case where $Q$ is determined directly from predictions generated by a pre-trained task DNN. In this case, let $f_\theta$ be pre-trained to predict $P(Y|X)$. Then, let $z \in \R^k$ be the pre-softmax logits obtained from $f_\theta$ and $\hat{y} = \operatorname{softmax}(z)$ where each $\hat{y}_i = P(Y = i|X)$ for $i \in {1,\dots, K}$.
We consider three possible variants of $Q$ in this setting: (1) Max probability: $Q_p := \max_i \hat{y_i}$, (2) Entropy: $Q_h := H(\hat{y}) = - \sum_i \hat{y_i} \log \hat{y_i}$, and (3) Max logit: $Q_l := \max_i z_i$.  While all three cases are inherently tied to the underlying label set $\gY$, the values of $Q$ do \textbf{not} have access to the ground truth label $Y$. Each of these $Q$ implicitly capture a DNN's confidence about its prediction and the natural underlying hypothesis is that confidence and image quality are positively correlated (i.e., as quality decreases, confidence also tends to decrease). These choices for $Q$ are driven by their use in out-of-distribution~\citep{Hendrycks2016-nk, Hendrycks2019-rw, Hendrycks2020-rk} and distribution shift detection~\citep{Wang2020-ck}.

Using the same setup as in Section~\ref{sub:exp-setup}, we now replace the NR-IQA metrics with $Q_p, Q_h, Q_l$.  As in Section~\ref{sec:exp1}, we examine the group-wise correlation and point-wise predictability of $M$ from $Q$. To ensure our test of predictability is fair, we use separate models for obtaining $M$ and $Q$ (\eg ConvNext-B and Swin-B respectively).  We provide additional results for other model pairs in Appendix~\ref{app:strong-tg-iqa}.  Figure~\ref{fig:acc-prob-entropy} shows the group-wise relationship between $Q, M$ where groups are averages over all images for the corresponding corruption, severity.

The results in Figure~\ref{fig:acc-prob-entropy} and Table~\ref{tab:acc-iq-strong-tg} show that strong task-guidance for $Q$ results in high correlation between $Q,M$ and predictability of $M$ from $Q$ (\textbf{D3}). \textit{This result is important to show that by using the causal framework it is possible to find a metric $Q$ that relies on a similar set of features as separate task models and is predictive of $M$.}

The high correlation between $M$, $Q$ exists across all task DNN architectures and variants of the metric which is consistent with the scenario shown in Fig.~\ref{fig:preds-dag}. In particular, given the DAG in Fig.~\ref{fig:preds-dag}, we see that without placing strong requirements on the specific task model architecture/design (e.g., convnet vs. transformer), if the task DNN producing predictions $\hat{Y}$ is sensitive to the imaging conditions, $Q$ will be strongly correlated with $M$. 

However, like the previous section, this approach is only a partial solution since it satisfies \textbf{D1, D2, D3} but clearly violates \textbf{D4} by requiring a model already trained for the classification task. In this case, modeling/optimization choices for the model used for $Q$ may result in unintended biases (e.g., performance disparities across (image class, distortion type) pairs) that might skew estimates of $Q$. 

\subsection{Restoring the association between $Q, M$ via weak task-guidance (\textbf{D3, D4})}
\label{sec:weak-tg-iqa}

So far, Sec.~\ref{sec:exp1} showed that common NR-IQA metrics are weakly predictive of DNN performance and are therefore not viable candidates for supporting image/dataset-level analysis given our desiderata. Then, we were able to address the predictability issue (\textbf{D3}) in Sec.~\ref{sec:strong-tg-iqa} using strong task-guidance, but at the cost of requiring a task model already trained for the classification task (a violation of \textbf{D4}). To satisfy all desiderata, we use the causal framework to design a task-guided metric (ZSCLIP-IQA) that uses the weaker zero-shot formulation of the image classification task to align $Q$ and $M$ (Sec.~\ref{sub:zsclip-iqa}).

\begin{figure}[h!]
    \centering
    \includegraphics[width=0.9\linewidth]{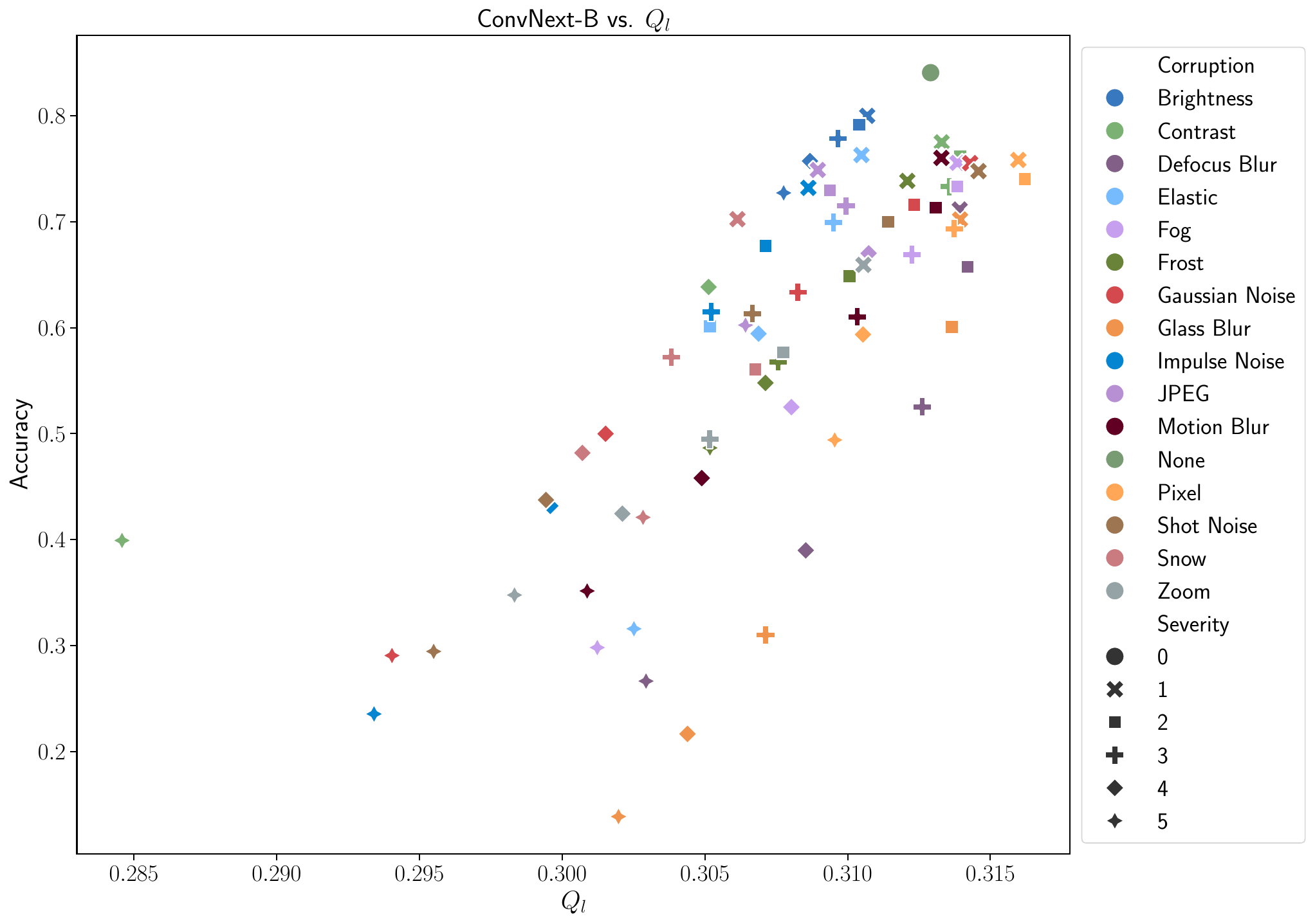}
    \caption{\textbf{Relationship of weak task-guided IQA and classification accuracy.} Accuracy vs. IQ with ConvNext-B as the task model $M$ and ZSCLIP-IQA max-logit as the quality metric $Q_l$ which generally outperforms the other variants.}
    \label{fig:acc-iq-weak-tg}
\end{figure}

Using the setup from Sec.~\ref{sec:exp1} we now replace the NR-IQA metrics with $Q_p, Q_h, Q_l$ based on the weak task-guided ZSCLIP-IQA method described in Section~\ref{sub:zsclip-iqa}.  We again examine the group-wise correlation and point-wise predictability of $M$ from $Q$. Figure~\ref{fig:acc-iq-weak-tg} and Table~\ref{tab:acc-iq-weak-tg} show that weak task-guidance is enough to restore the association between $Q$ and $M$ without requiring a new task model to be trained on the dataset of interest.  

We also note that while the CLIP backbone for ZSCLIP-IQA is pre-trained on a self-supervised task that resembles classification, it was not exposed to ImageNet (or IN-C) data during its training (see Sec.~5 in~\citep{Radford2021-sr}) and can be effectively used here in a zero-shot setting to satisfy \textbf{D4}.  In fact, while methods like CLIP-IQA and ARNIQA also rely on pre-trained backbones, the results of Tables~\ref{tab:acc-iq} and~\ref{tab:acc-iq-weak-tg} show that only ZSCLIP-IQA is ``guided'' (via our causal framework) to be a stronger predictor of DNN performance compared to other methods calibrated to human perceptual judgment.

The results of this experiment confirm that using the causal framework we can design an IQA metric that satisfies all of the desiderata. We note that while the group-wise correlation between $M, Q$ is high, the predictability (measured via AUC) is not as significant given that the DNN used as the backbone for the IQ metric is no longer the same as that used for the task.  
Nonetheless, the weak task-guided metric still achieves above chance AUC which is a significant improvement over conventional NR-IQA metrics (see Table~\ref{tab:acc-iq}).  Future research may be able to further use the weak task-guided formulation and causal framework to improve this result.  
\textit{The primary implication of this result is that the causal framework enabled us to develop an IQ metric that satisfies all of our desiderata and thus allows us to effectively predict task DNN performance on an image dataset by analyzing its quality distribution.  We perform one such analysis in Section~\ref{sub:exp3a-mild-corrupt}.}

\subsection{Predictability of DNN performance for mildly corrupted datasets}
\label{sub:exp3a-mild-corrupt}

In the previous experiments, the use of IN-C allowed us to investigate the large-scale effect of image corruptions on the predictability of performance by using multiple corrupted versions of the validation set with multiple levels of severity.  In real-world datasets, we expect that only a small fraction of images will be corrupted.  
We next examine the extent to which IQA metrics can be used to show differences between the quality distributions of datasets containing varying levels of corruption while still satisfying \textbf{D1-D4} in these more realistic settings. 

To answer this question, we generate new variants of IN-val consisting of mixtures of clean and corrupted images.  For each variant, we specify a set of valid corruptions $\gC$, severities $\gS$, and a corruption probability $p_{c}$.  We choose a fraction $1 - p_c$ of the original IN-val image IDs to remain as clean images and a fraction $p_c$ to be corrupted.  The corrupted images are sampled uniformly amongst the corruptions $c \in \gC$ and severities $s \in \gS$.  The resulting variant consists of the original 50k image IDs with a mixture of clean and corrupted images. We choose $\gC$ to consist of all 15 corruptions in the IN-C dataset and limit severity to $\gS = \{1,2,3\}$ in order to further test the sensitivity of the IQA metrics (\textbf{D1}).  We create variants of the IN-val dataset for $p_c = N/100$ for $N \in [1,\dots,20]$ .  We evaluate the DNNs on these dataset variants and estimate predictability using logistic regression as in previous experiments.  

We compute $mAUC$ over all $p_c$ variants and find ZSCLIP-IQA ($Q_l$) outperforms all other NR-IQA metrics with $mAUC=0.64$ with the next best (CLIP-IQA) achieving only $mAUC=0.57$. The full results are found in Appendix~\ref{app:pred-real-world-data} and show that predictability with ZSCLIP-IQA is stable with respect to changes in the proportion of clean/corrupted images in the dataset whereas more traditional NR-IQA metrics remain near random chance $AUC$ and exhibit higher variance as $p_c$ changes.

This experiment examines how IQ metrics can be used to measure the variability of image quality within and across image \emph{datasets} and to provide insight into the overall difficulty of the dataset for downstream task DNNs. 
The results show that while all metrics in this experiment can distinguish between differences in the quality distributions of the dataset variants, only ZSCLIP-IQA achieves high predictability over all variants. Conventional IQA metrics improve only as the number of distorted images in the dataset increases (where it becomes easier to separate clean and corrupted images).  

The broader impact of these results is that we can use IQ metrics that satisfy the desiderata as a means to audit datasets to provide hypotheses about how their composition relates to task DNN performance.  For instance, if an image quality distribution is estimated using a weak task-guided metric and found to skew towards higher-quality images, it may be difficult to draw strong conclusions about task DNN robustness evaluated on that data. In contrast, if the quality distribution skews towards lower-quality images, then there is stronger evidence that a task DNN that performs well on the data is a robust model.  \textit{The primary implication of this experiment is that we show how the weak task-guided metric, developed using our causal framework, provides a means to effectively estimate and compare dataset-level image quality distributions in this way.}

\vspace{-8pt}
\section{Discussion}
\label{sec:discussion}

In this work, we were motivated to identify measures of image quality that allow us to produce IQ-driven priors on DNN performance.  We presented a causal inference framework for this problem and proposed a set of IQ metric desiderata to guide our analysis (Sec.~\ref{sec:causal-iqa}).
Using our causal framework, we show conditions where image quality measures can be predictive of DNN performance.
We then provide a first-of-its-kind detailed examination of the relationship between conventional NR-IQA metrics and DNN performance. We use our causal framework and extensive empirical evaluations in the context of image classification to demonstrate that common NR-IQA do not satisfy our desired IQ criteria (Sec.~\ref{sec:exp1}).  
We then use the causal approach to develop introduce the notion of task-guided IQA metrics (Sec.~\ref{sec:strong-tg-iqa}).  In particular, the ZSCLIP-IQA (Sec.~\ref{sub:zsclip-iqa}) metric
provides a causality-driven proof-of-concept that satisfies all IQ desiderata and paves the way for future research to improve the alignment between IQA metrics and DNN performance.

One broad implication from this work is that our framework exposed the existence of cases where IQ metrics and task DNNs are not sensitive to the same features of image quality.  In the context of using IQ metrics as priors for DNN performance (e.g., to estimate image understanding difficulty), we demonstrated for the first time that conventional NR-IQA metrics are generally not appropriate for this task.  However, while our task-guided IQA metrics were shown to be suitable in contexts where task DNNs are the downstream consumers of the image data, conventional NR-IQA metrics (e.g., BRISQUE, TV, CLIP-IQA, ARNIQA) may still be more appropriate for applications where humans are the target consumers.

Similarly, while the task-guided metrics were found to be predictive of task DNN performance, there was not much disparity in the degree of predictability observed across different classes of task DNN architectures (e.g., convolutional nets, transformers).  As such, the task-guided metrics could be used to aid in making choices between task model architectures in downstream applications by first predicting baseline performance on targeted image datasets using the TG-IQA metric and then comparing actual task DNN performance against those predictions to determine which architecture exhibits the greatest improvement over the baseline.

Lastly, the results of Section~\ref{sub:exp3a-mild-corrupt} demonstrate how our work lays the foundation for future research to examine how to analyze dataset-level properties of images. Such analyses can be used to determine how properties of datasets (e.g., image quality distribution) impact downstream consumers of the data.  For instance, by first understanding the quality distribution of robustness benchmark datasets, downstream evaluation results obtained using these benchmarks can be better contextualized.  Such findings will allow researchers to better de-couple properties of datasets from properties of their models, thus enabling them to make more targeted improvements to either.  This capability is of growing interest as imaging datasets continue to grow in size beyond the limits of qualitative assessment by humans.

As a tool for analysis, the proposed causal framework poses minimal broader risks. While causal models require assumptions about the data generating process, these assumptions are made explicitly in the causal graph and improve the overall transparency of the analysis.  Of greater concern is the possibility that using quality metrics to prune/resample datasets may lead to unintended consequences such as removing poor-quality images in a way that disparately affects protected groups. While our work does not address the question of dataset pruning/resampling, we mention this to help ensure that future researchers consider these possibilities in their own work.

We also recognize that image quality alone is insufficient to predict task model performance as both image content and composition play a role in task difficulty and the relationship between IQA and performance may be confounded by other factors. 
While the experiment in Appendix~\ref{app:control-for-content} examined two approaches to controlling for content, future work should more deeply address this question.
Nonetheless, we show that our causal framework still allows us to analyze the conditions where quality properties of our dataset may be correlated with DNN performance. Our ZSCLIP-IQA method provides one solution that satisfies the proposed IQ desiderata but we believe there are many opportunities for improving on this approach in future research. 

We also acknowledge that our experiments only addressed image classification tasks. We focused initially on classification since it is well-studied and clearly defined, with many public benchmarks available for evaluation. Even in this context, we are the first to show that the notion of quality is task-dependent (i.e., perceptual judgment vs. classification). Our primary contribution in this work is the causal framework for IQA and we believe this provides a strong foundation for supporting future research that examines similar questions for a wider range of vision tasks.

\section{Methods}
\label{sec:causal-iqa}

Image quality assessment has been long-studied in the computer vision and image processing literature. Full Reference IQA (FR-IQA)~\citep{wang2004image, zhang2011fsim, zhang2018unreasonable} assume the availability of a reference or ``clean'' image against which the test image is compared and the quality is measured.  In contrast, the NR-IQA~\citep{Xu2017-rp, zhang2011fsim, wang2004image, mittal2012no, Wang2022-vx, Agnolucci2024-ti, Mittal2013-dq, Ye2012-iu, Xu2016-wz} setting (aka Blind IQA) uses only features of the test image to estimate a quality score. In both settings, conventional IQA methods are calibrated and compared against human perceptual judgments of quality such as Mean Opinion Scores (MOS).  These measures are task-agnostic and humans are not required to make judgments about the content of the image but only to measure subjective ``quality'' (typically on a scale of 1-5, Poor-High). This motivates our investigation into whether these metrics can also provide task-relevant assessments of image quality.

In addition, separate from IQA research, considerable attention has been paid to understanding and characterizing the impact of image quality on DNN robustness~\cite{Drenkow2021-pw}.  In conventional settings, robustness is evaluated by exposing DNNs to images degraded by precisely controlled synthetic corruptions that target specific isolated imaging conditions~\cite{Dodge2016-fw, Hendrycks2019-ye, Michaelis2019-gi, Laugros2019-lv, Laugros2021-fd, Taori2020-xa, Djolonga2021-eo, Varga2024-ef, Oguine2024-md}. Later evaluations of robustness have targeted imaging conditions that capture more challenging, naturally-occurring degradations~\cite{Ibrahim2022-lk, Drenkow2024-gl, Zhao2021-gl, Drenkow2024-hh}. In all of these works, quantitative measures of IQ were not explicitly correlated with DNN robustness.  Our work is the first to our knowledge to systematically examine this relationship and paves the way for correlating image quality and DNN robustness when control over image degradation is not possible.

A key question of this work centers on whether IQA metrics calibrated against human MOS are sensitive to any of the same image features that DNNs use for downstream tasks.  Outside of the IQA literature, prior works have shown differences in humans and DNNs in the context of shape/texture bias~\citep{Geirhos2018-ct, Hermann2019-gg}, shortcut learning~\citep{Geirhos2020-ag, Zech2018-hs, Brown2022-tf, Ong_Ly2024-jh}, and error consistency~\citep{Geirhos2021-bp, Wichmann2023-gl}, but the question remains open whether IQA metrics aligned with human MOS correlate with DNN performance.  

Our work is strongly motivated by the growing interest in automated methods for dataset analysis.  In particular, new methods focus on dataset pruning~\citep{Tan2023-tv, He2023-ev, Abbas2024-lo}, identifying difficult/important examples~\citep{Kwok2024-qm, Ilyas2022-eh} or data slices~\citep{Eyuboglu2022-eh, Chung2019-qn, Sohoni2020-ae, Chen2019-vn}, and dataset auditing for shortcuts~\citep{Pavlak2023-vh, Drenkow2025-xx}.  Other results have shown that understanding dataset composition matters for analyzing model robustness~\citep{Ibrahim2022-lk, Drenkow2024-gl}.  Our work takes a positive step towards automated methods for analyzing the quality distribution of image datasets and establishing priors on DNN performance.

\subsection{Causal Framework for IQA}
\label{sub:framework}
To understand the relationship between IQA metrics and DNN performance, we provide here a causal inference perspective on the IQA problem.  We use causal directed acyclic graphs (DAGs) to illustrate our assumptions about the imaging generating process, quality metric, and performance metric as well as the interactions between all associated variables.  This causal framework provides a means for identifying the specific conditions under which quality metrics are predictive of DNN performance.

We specify a causal DAG $\gG$ via a set of nodes $\gV$ and directed edges $\gE$.  To obtain the causal interpretation, directed edges imply a causal relationship such that for a variable/node $V \in \gV$, $V$ is a function of its parents ($V = f_V(pa(V), U)$ where $U$ is an exogenous noise term).

For defining causal models in the IQA context, we start from a set of factors $A \in \gA$ that capture the key variables in the data generating process affecting the image conditions (e.g., lighting, focal length, aperture, exposure, weather). Let $X \in \gX$ be the resulting images, and for a task $T$, let $Y \in \gY$ be the label associated with $X$ for the task.  For this work, we focus on classification tasks where $\gY$ consists of a discrete set of $K$ classes ($\gY = \{1,\dots,K\}$).  Our quality metric $Q: \gX \rightarrow \R$ maps images to real number scores (typically in $[0, 1]$ where $1$ is the highest quality).  We also assume a downstream task DNN $f_\theta: \gX \rightarrow \gY$ that maps images to class probabilities and is parameterized by $\theta$.  We write the predicted probabilities $\hat{Y} = f_\theta(X)$ where $\hat{Y} \in \R^K$.  Given one-hot encoded labels $Y$ and predictions $\hat{Y}$, we can compute a performance metric (e.g., accuracy) $M: \gY \times \gY \rightarrow \R$.  In the general case and without loss of generality, we assume that $\hat{Y}$ is the prediction from a deterministic DNN.  Similarly, we also assume that $Q, M$ are both deterministic functions of their parents in the causal DAG.

\subsection{Baseline IQA formulation}
\label{sub:base-iqa}

We start with the baseline formulation of the IQA problem as shown in Figure~\ref{fig:standard-dag}. We assume the labels $Y$ are determined from interpreting $X$ and that an oracle labeling function exists such that $Y$ can always be determined from $X$.

This model is a general formulation and makes no assumptions about the nature of the functions that compute $Q, \hat{Y}$. This causal model is also consistent with conventional NR-IQA settings~\citep{mittal2012no, Ma2018-mx} where the determination of $Q$ given $X$ is based on a function calibrated to human perceptual judgment and without knowledge of the task or labels.

\paragraph{Conditional independence of $Q, M$:}
The causal graph of Figure~\ref{fig:standard-dag} illustrates that under the baseline formulation, $Q \perp M | X$ and $X$ is said to \textit{d-separate} $Q, M$.  The interpretation here is that given any image $X$, there is no expected relationship between $Q$ and $M$ by construction.  This is not to say that a relationship cannot exist, but simply that there is nothing in this model that ensures it directly.  

In addition to observing the d-separation of $Q,M$ by $X$, we can also compute the average causal effect (ACE) of $Q\rightarrow M$. We use the potential outcomes notation $M(Q=q)$ (or $M(q)$) to indicate the value of $M$ if $Q$ had been set to the value of $q$.
{\small
\begin{align*}
    ACE(Q \rightarrow M) &= \E[M(Q=q) - M(Q=q')] = \E[M(q)] - \E[M(q')]  & \text{linearity of expectation} \\
    &= \E_X[\E[M(q)|X] - \E[M(q')|X]] & \text{law of total expectation} \\
    &= \E_X[\E[M(q)|Q=q, X] - \E[M(q')|Q=q', X]] & \text{unconfoundedness} \\
    &= \E_X[\E[M|Q=q, X] - \E[M|Q=q', X]] & \text{consistency} \\
    &= \E_X[\E[M|X] - \E[M|X]] = 0.  & \text{conditional independence}
\end{align*}}
The absence of causal effect and association between $Q,M$ in this formulation suggests that, without further assumptions, traditional IQA metrics should not be predictive of DNN performance.  Furthermore, while No Reference and Full Reference (FR) IQA differ in their assumptions and setup, we provide their causal models as special cases of Figure~\ref{fig:standard-dag} in Appendix~\ref{app:iqa-dags} to show that $Q \perp M | X$ holds in both cases.

\subsection{Shared features IQA formulation}
\label{sub:latents-iqa}

Ideally, we would like $Q$ and $M$ to become dependent when conditioning on $X$ (i.e., we can learn about $M$ by observing only $Q$). This occurs in the case where there exists a common set of features $Z$ that are utilized both for the prediction function of $\hat{Y}$ and the quality score $Q$ as shown in Figure~\ref{fig:latents-dag}. This scenario does not presume a singular set of $Z$ that serves the task model and quality metric, but rather, the $Z$ shown here represents the intersection of features used by both.  The existence of $Z$ ensures $Q, M$ are no longer independent given $X$. The primary question is whether such a $Z$ exists or whether $Q, M$ are related only as shown in Figure~\ref{fig:standard-dag}. An expanded discussion relating the baseline and shared features models can be found in Appendix~\ref{app:latents-dag}.

\textbf{Remark: Correlation vs. Causation}~~~
While we typically use causal models to estimate cause-effect relationships, a key clarification here is that we seek a weaker criterion, namely to establish the conditions under which quality and performance are \textit{at least} correlated.  Since we know that $X$ is a common cause for both $Q, M$, we first want to ensure that when we estimate $Q, M$ from $X$ we know the conditions under which $Q, M$ will be related via the same features of $X$ (e.g., via $Z$).

To address the aforementioned issues, we consider instead a weaker form of task-guidance where quality metrics can be aligned with task-specific information without requiring the expense of training a new task model directly on the dataset of interest (\textbf{D4}).  In this setting shown in Figure~\ref{fig:zsclip-dag}, the computation of $Q$ is dependent not only on the image $X$ but on the label set $\gY$.  The task $T$ is used as a selection variable~\cite{Zadrozny2004-oh, Bareinboim2022-cf} on which the dataset is conditioned, and as a collider in the DAG, $T$ creates an association between $M, Q$.

\subsection{Zero-Shot CLIP IQA}
\label{sub:zsclip-iqa}

Using the causal framework, we propose a new quality metric that uses Zero-Shot (ZS) capabilities of the multi-modal CLIP foundation model~\cite{Radford2021-sr} in order to address all desiderata.  In particular, we derive a new image quality metric (ZSCLIP-IQA) based on a zero-shot classification problem for our data and task of interest (Figure~\ref{fig:zsclip-dag}).   

Let $\gD = \{ (x_i, y_i) \}_{i=1}^{N}$ be our dataset with images $x$ and labels $y$, $f: \gX \rightarrow \gZ$ be our CLIP image embedding network ($\gZ \in \R^d$), and $g: \gT \rightarrow \gW$ be our CLIP text/token-embedding network ($\gW \in \R^d$). 

We define a set of task-relevant classes/tokens $\gT = \{T_i\}_{i=1}^{K}$ that capture the text labels for concepts or entities likely to occur in the images (e.g., $K=1000$ classes in the ImageNet dataset).  We embed each of the text tokens $g(T_i)=w_i$ and normalize to get a unit vector representation for each token.  Note when using CLIP as our text embedding network, we may also augment $T_i$ to include additional words (e.g., \verb|"A picture of a <token>"|). The full set $\mathbf{W}=[w_0; w_1;~\cdots~w_K] \in \R^{d \times K}$ constitutes the ZS weights.

To evaluate image quality, we compute image embeddings $f_\theta(x)=z$ for $x \in \gD$ which we normalize to be unit length. For each image, we compute the cosine similarity between the image embedding and each of the tokens $s = z \mathbf{W}$ (with $z \in \R^d,~s \in \R^K$) and compute estimated class probabilities via a softmax over similarity scores ($\hat{y} = \operatorname{softmax}(s)$).  As in the strong TG-IQA scenario, we implement three variants of $Q$: (1) Max probability: $Q_p := \max_i \hat{y_i}$, (2) Entropy: $Q_h := H(\hat{y}) = - \sum_i \hat{y_i} \log \hat{y_i}$, and (3) Max-logit: $Q_l := \max_i s_i$.

\section*{Data Availability}
The datasets generated and/or analyzed during the current study are available in the ImageNet-C repository (\url{https://zenodo.org/records/2235448}).

\section*{Code Availability}
The code used for this study is available from the corresponding author
upon reasonable request.

\section*{Acknowledgments}
This work is funded by Independent Research \& Development (IR\&D) program funds in the Research and Exploratory Development Department (REDD) at the Johns
Hopkins University Applied Physics Laboratory.

\section*{Author Contributions}
ND developed the method, designed and ran the experiments, analyzed the results and wrote the main manuscript. MU supported the development of the method, design of experiments, and contributions to the main manuscript. All authors reviewed the manuscript.

\section*{Competing Interests}
The authors declare no competing interests.

\printbibliography[title=References]

\clearpage
\begin{figure}[h!]
    \centering
    \resizebox{0.65\linewidth}{!}{%
    \includegraphics[width=\linewidth]{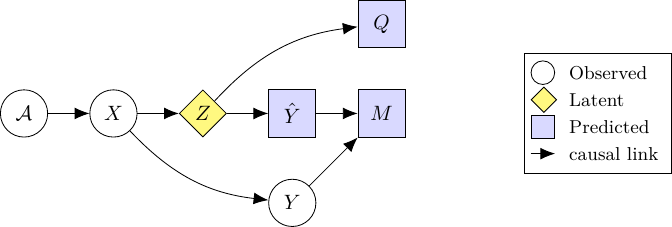}
    }%
    \caption{\textbf{Baseline IQA formulation with shared latent features.} Causal diagram relating model accuracy ($M$) with IQ metrics ($Q$). In this case, $Q$ and $M$ are related via a common cause $Z$ that represents features in the image $X$ that influence both the prediction and quality.}
    \label{fig:latents-dag}
\end{figure}

\begin{figure}[h!]
    \centering
    \resizebox{0.65\linewidth}{!}{%
    \includegraphics[width=\linewidth]{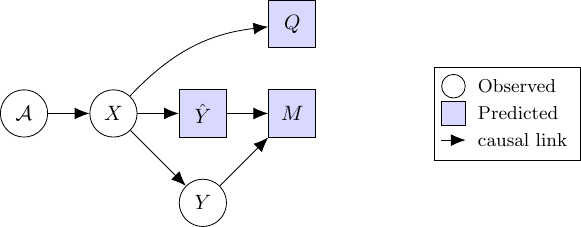}
    }%
    \caption{\textbf{Baseline IQA formulation.} Causal diagram relating model accuracy ($M$) with IQ metrics ($Q$).}
    \label{fig:standard-dag}
\end{figure}

\begin{figure}[h!]
    \centering
    \includegraphics[width=0.9\linewidth]{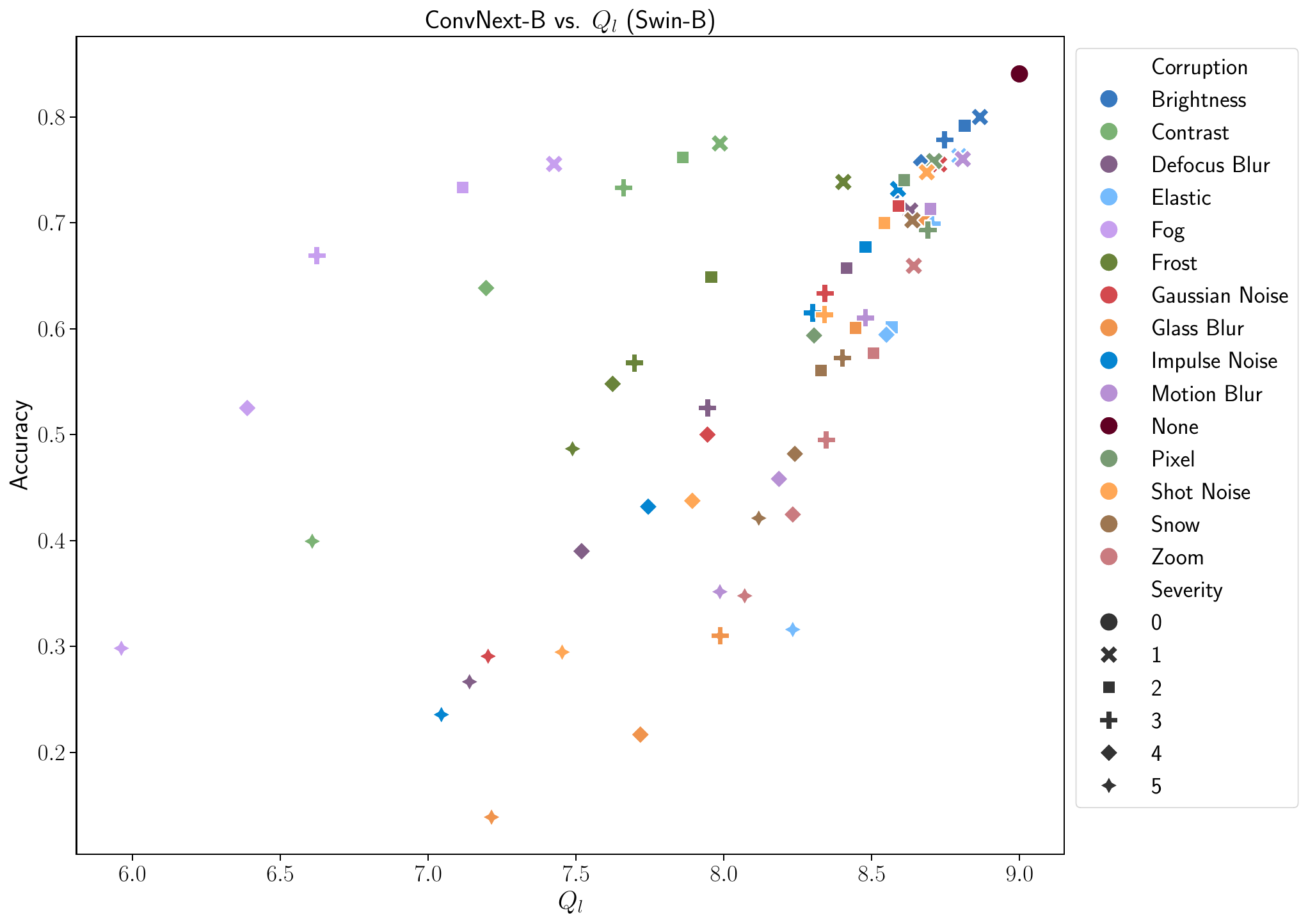}
    \caption{\textbf{Relationship of strong task-guided IQA and classification accuracy.} Accuracy vs. max logit using ConvNext-B for the task model and Swin-B for $Q_l$ which generally outperforms the other variants}
    \label{fig:acc-prob-entropy}
\end{figure}

\begin{figure}[h!]
    \centering
    \resizebox{0.55\linewidth}{!}{%
    \includegraphics[width=\linewidth]{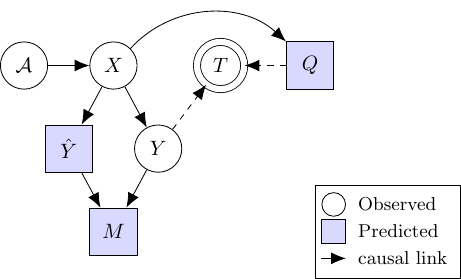}
    }%
    \caption{\textbf{Weak task-guided IQA.} Causal DAG relating model performance ($M$) with IQ metrics ($Q$). $Q$ uses information about the label set $\gY$. The task $T$ is viewed as a selection variable which influences both the labels $Y$ and $Q$.}
    \label{fig:zsclip-dag}
\end{figure}
\begin{table*}[h!]
    \centering
    \caption{Correlation between IQ and accuracy, correctness. SRCC, PLCC computed using average accuracy for each (corruption, severity). {\small AUC and CE based on point-wise predictions (95\% CI within $\pm0.001$). SRCC, PLCC values have $p < 0.05$.}}
    \resizebox{\linewidth}{!}{%
    \begin{tabular}{cl|ll||lll}
\toprule
Model & IQA Metric & AUC $\uparrow$ & CE $\downarrow$ & $\mid KRCC \mid~\uparrow$ & $\mid PLCC \mid$ $\uparrow$ & $\mid SRCC \mid$ $\uparrow$ \\
\midrule
\multirow[c]{4}{*}{ConvNext-B} & ARNIQA & 0.517 & 0.677 & 0.088$\pm$0.149 & 0.168$\pm$0.215 & 0.127$\pm$0.214 \\
 & BRISQUE & 0.568 & 0.670 & 0.255$\pm$0.129 & 0.398$\pm$0.190 & 0.374$\pm$0.182 \\
 & CLIP-IQA & 0.567 & 0.670 & 0.273$\pm$0.154 & 0.328$\pm$0.202 & 0.378$\pm$0.212 \\
 & TV & 0.477 & 0.676 & 0.108$\pm$0.183 & 0.138$\pm$0.294 & 0.151$\pm$0.255 \\
\cline{1-7}
\multirow[c]{4}{*}{ResNet34} & ARNIQA & 0.499 & 0.663 & 0.003$\pm$0.155 & 0.006$\pm$0.205 & 0.007$\pm$0.225 \\
 & BRISQUE & 0.552 & 0.658 & 0.175$\pm$0.140 & 0.278$\pm$0.186 & 0.254$\pm$0.202 \\
 & CLIP-IQA & 0.599 & 0.647 & 0.307$\pm$0.159 & 0.467$\pm$0.188 & 0.429$\pm$0.207 \\
 & TV & 0.500 & 0.657 & 0.051$\pm$0.194 & 0.291$\pm$0.256 & 0.047$\pm$0.264 \\
\cline{1-7}
\multirow[c]{4}{*}{EfficientNet-V2-M} & ARNIQA & 0.522 & 0.674 & 0.098$\pm$0.149 & 0.204$\pm$0.213 & 0.143$\pm$0.214 \\
 & BRISQUE & 0.579 & 0.665 & 0.271$\pm$0.137 & 0.431$\pm$0.186 & 0.393$\pm$0.188 \\
 & CLIP-IQA & 0.574 & 0.666 & 0.276$\pm$0.158 & 0.332$\pm$0.209 & 0.389$\pm$0.213 \\
 & TV & 0.468 & 0.675 & 0.109$\pm$0.181 & 0.075$\pm$0.275 & 0.158$\pm$0.254 \\
\cline{1-7}
\multirow[c]{4}{*}{MobileNet-V3-L} & ARNIQA & 0.503 & 0.682 & 0.008$\pm$0.148 & 0.024$\pm$0.215 & 0.019$\pm$0.216 \\
 & BRISQUE & 0.562 & 0.676 & 0.198$\pm$0.137 & 0.314$\pm$0.191 & 0.285$\pm$0.196 \\
 & CLIP-IQA & 0.586 & 0.669 & 0.300$\pm$0.159 & 0.412$\pm$0.192 & 0.415$\pm$0.215 \\
 & TV & 0.506 & 0.676 & 0.038$\pm$0.191 & 0.298$\pm$0.271 & 0.035$\pm$0.261 \\
\cline{1-7}
\multirow[c]{4}{*}{Swin-B} & ARNIQA & 0.510 & 0.675 & 0.069$\pm$0.155 & 0.118$\pm$0.230 & 0.098$\pm$0.222 \\
 & BRISQUE & 0.574 & 0.667 & 0.291$\pm$0.131 & 0.443$\pm$0.183 & 0.426$\pm$0.178 \\
 & CLIP-IQA & 0.571 & 0.667 & 0.290$\pm$0.161 & 0.361$\pm$0.199 & 0.410$\pm$0.211 \\
 & TV & 0.485 & 0.674 & 0.090$\pm$0.180 & 0.153$\pm$0.299 & 0.123$\pm$0.255 \\
\cline{1-7}
\multirow[c]{4}{*}{ViT-L16} & ARNIQA & 0.508 & 0.674 & 0.067$\pm$0.157 & 0.098$\pm$0.226 & 0.097$\pm$0.224 \\
 & BRISQUE & 0.562 & 0.668 & 0.274$\pm$0.125 & 0.430$\pm$0.172 & 0.407$\pm$0.175 \\
 & CLIP-IQA & 0.575 & 0.665 & 0.293$\pm$0.152 & 0.414$\pm$0.180 & 0.422$\pm$0.202 \\
 & TV & 0.485 & 0.673 & 0.103$\pm$0.182 & 0.181$\pm$0.289 & 0.136$\pm$0.247 \\
\cline{1-7}
\bottomrule
\end{tabular}

    }%
    \label{tab:acc-iq}
\end{table*}

\begin{table*}[h!]
    \centering
    \caption{Correlation between IQ and accuracy, correctness. KRCC, SRCC, PLCC computed using average accuracy for each (corruption, severity). {\small AUC and CE based on point-wise predictions (95\% CI within $\pm0.001$). KRCC, SRCC, PLCC values have $p < 0.05$.} Full table in Appendix~\ref{app:strong-tg-iqa}.}
    \resizebox{\linewidth}{!}{%
    
\begin{tabular}{cl|ll||lll}
\toprule
Model & IQA Metric & AUC $\uparrow$ & CE $\downarrow$ & $\mid KRCC \mid~\uparrow$ & $\mid PLCC \mid$ $\uparrow$ & $\mid SRCC \mid$ $\uparrow$ \\
\midrule
\multirow[c]{3}{*}{ConvNext-B} & $Q_h$ & 0.772 & 0.562 & 0.660$\pm$0.070 & 0.822$\pm$0.070 & 0.854$\pm$0.063 \\
 & $Q_l$ & 0.778 & 0.555 & 0.660$\pm$0.067 & 0.826$\pm$0.067 & 0.854$\pm$0.063 \\
 & $Q_p$ & 0.826 & 0.504 & 0.738$\pm$0.045 & 0.888$\pm$0.045 & 0.910$\pm$0.044 \\
\cline{1-7}
\multirow[c]{3}{*}{ResNet34} & $Q_h$ & 0.848 & 0.470 & 0.862$\pm$0.028 & 0.930$\pm$0.028 & 0.969$\pm$0.023 \\
 & $Q_l$ & 0.827 & 0.492 & 0.870$\pm$0.015 & 0.951$\pm$0.015 & 0.973$\pm$0.020 \\
 & $Q_p$ & 0.850 & 0.461 & 0.886$\pm$0.015 & 0.960$\pm$0.015 & 0.977$\pm$0.021 \\
\cline{1-7}
\multirow[c]{3}{*}{EfficientNet-V2-M} & $Q_h$ & 0.831 & 0.497 & 0.888$\pm$0.020 & 0.956$\pm$0.020 & 0.981$\pm$0.012 \\
 & $Q_l$ & 0.831 & 0.496 & 0.876$\pm$0.029 & 0.937$\pm$0.029 & 0.977$\pm$0.014 \\
 & $Q_p$ & 0.862 & 0.456 & 0.900$\pm$0.014 & 0.968$\pm$0.014 & 0.984$\pm$0.010 \\
\cline{1-7}
\multirow[c]{3}{*}{MobileNet-V3-L} & $Q_h$ & 0.833 & 0.496 & 0.906$\pm$0.019 & 0.973$\pm$0.019 & 0.984$\pm$0.013 \\
 & $Q_l$ & 0.835 & 0.496 & 0.895$\pm$0.031 & 0.951$\pm$0.031 & 0.982$\pm$0.015 \\
 & $Q_p$ & 0.853 & 0.470 & 0.922$\pm$0.010 & 0.983$\pm$0.010 & 0.988$\pm$0.010 \\
\cline{1-7}
\multirow[c]{3}{*}{Swin-B} & $Q_h$ & 0.766 & 0.578 & 0.532$\pm$0.207 & 0.483$\pm$0.207 & 0.654$\pm$0.174 \\
 & $Q_l$ & 0.732 & 0.597 & 0.485$\pm$0.203 & 0.458$\pm$0.203 & 0.611$\pm$0.181 \\
 & $Q_p$ & 0.807 & 0.529 & 0.603$\pm$0.184 & 0.620$\pm$0.184 & 0.732$\pm$0.142 \\
\cline{1-7}
\multirow[c]{3}{*}{ViT-L16} & $Q_h$ & 0.799 & 0.545 & 0.633$\pm$0.124 & 0.690$\pm$0.124 & 0.819$\pm$0.092 \\
 & $Q_l$ & 0.780 & 0.556 & 0.592$\pm$0.124 & 0.672$\pm$0.124 & 0.786$\pm$0.092 \\
 & $Q_p$ & 0.828 & 0.506 & 0.694$\pm$0.104 & 0.765$\pm$0.104 & 0.864$\pm$0.075 \\
\cline{1-7}
\bottomrule
\end{tabular}

    }%
    \label{tab:acc-iq-strong-tg}
\end{table*}

\begin{table*}[h!]
    \centering
    \caption{Correlation between IQ and accuracy.  KRCC, SRCC, PLCC based on average accuracy for each (corruption, severity) combination. AUC and CE are computed based on point-wise predictions and all have 95\% CI within $\pm0.001$. {\small All KRCC, SRCC, PLCC values have $p < 0.005$.}}
    \resizebox{\linewidth}{!}{%
    
\begin{tabular}{cc|cc||ccc}
\toprule
Model & ZSCLIP-IQA & AUC $\uparrow$ & CE $\downarrow$ & $\mid KRCC \mid~\uparrow$ & $\mid PLCC \mid~\uparrow$ & $\mid SRCC \mid~\uparrow$ \\
\midrule
\multirow[c]{3}{*}{ConvNext-B} & $Q_h$ & 0.349 & 0.677 & 0.738$\pm$0.067 & 0.869$\pm$0.059 & 0.906$\pm$0.048 \\
 & $Q_l$ & 0.675 & 0.632 & 0.573$\pm$0.084 & 0.764$\pm$0.080 & 0.783$\pm$0.085 \\
 & $Q_p$ & 0.602 & 0.677 & 0.788$\pm$0.056 & 0.884$\pm$0.048 & 0.937$\pm$0.034 \\
\cline{1-7}
\multirow[c]{3}{*}{ResNet34} & $Q_h$ & 0.666 & 0.663 & 0.800$\pm$0.059 & 0.936$\pm$0.028 & 0.945$\pm$0.032 \\
 & $Q_l$ & 0.692 & 0.609 & 0.630$\pm$0.087 & 0.814$\pm$0.055 & 0.820$\pm$0.084 \\
 & $Q_p$ & 0.368 & 0.663 & 0.834$\pm$0.049 & 0.949$\pm$0.024 & 0.959$\pm$0.026 \\
\cline{1-7}
\multirow[c]{3}{*}{EfficientNet-V2-M} & $Q_h$ & 0.344 & 0.675 & 0.721$\pm$0.074 & 0.844$\pm$0.069 & 0.890$\pm$0.061 \\
 & $Q_l$ & 0.674 & 0.631 & 0.545$\pm$0.084 & 0.717$\pm$0.079 & 0.757$\pm$0.089 \\
 & $Q_p$ & 0.604 & 0.675 & 0.759$\pm$0.062 & 0.860$\pm$0.051 & 0.920$\pm$0.042 \\
\cline{1-7}
\multirow[c]{3}{*}{MobileNet-V3-L} & $Q_h$ & 0.655 & 0.682 & 0.740$\pm$0.065 & 0.896$\pm$0.047 & 0.907$\pm$0.046 \\
 & $Q_l$ & 0.686 & 0.629 & 0.559$\pm$0.087 & 0.765$\pm$0.084 & 0.768$\pm$0.091 \\
 & $Q_p$ & 0.377 & 0.682 & 0.785$\pm$0.055 & 0.918$\pm$0.039 & 0.936$\pm$0.033 \\
\cline{1-7}
\multirow[c]{3}{*}{Swin-B} & $Q_h$ & 0.354 & 0.676 & 0.735$\pm$0.072 & 0.864$\pm$0.063 & 0.904$\pm$0.054 \\
 & $Q_l$ & 0.672 & 0.633 & 0.538$\pm$0.094 & 0.718$\pm$0.096 & 0.743$\pm$0.098 \\
 & $Q_p$ & 0.601 & 0.676 & 0.778$\pm$0.059 & 0.879$\pm$0.050 & 0.935$\pm$0.034 \\
\cline{1-7}
\multirow[c]{3}{*}{ViT-L16} & $Q_h$ & 0.363 & 0.674 & 0.789$\pm$0.053 & 0.915$\pm$0.042 & 0.938$\pm$0.032 \\
 & $Q_l$ & 0.673 & 0.630 & 0.592$\pm$0.086 & 0.775$\pm$0.088 & 0.791$\pm$0.083 \\
 & $Q_p$ & 0.599 & 0.674 & 0.845$\pm$0.045 & 0.920$\pm$0.042 & 0.962$\pm$0.023 \\
\cline{1-7}
\bottomrule
\end{tabular}

    }%
    \label{tab:acc-iq-weak-tg}
\end{table*}

\clearpage
\appendix
\section{Causal model for NR-/FR-IQA}
\label{app:iqa-dags}
\begin{figure}[h!]
    \centering
    \begin{subfigure}[b]{0.4\textwidth}
        \centering
        \resizebox{0.9\linewidth}{!}{%
        \includegraphics[width=\linewidth]{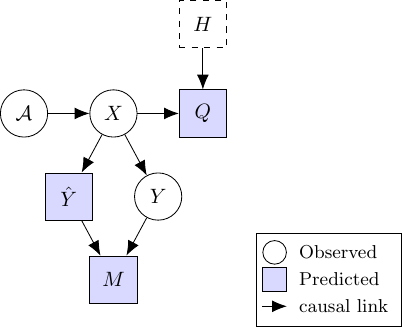}
        }%
        \caption{Causal model for NR-IQA}
        \label{fig:nr-iqa-dag}
     \end{subfigure}
     \hspace{2cm}
    \begin{subfigure}[b]{0.4\textwidth}
        \centering
        \resizebox{0.9\linewidth}{!}{%
        \includegraphics[width=\linewidth]{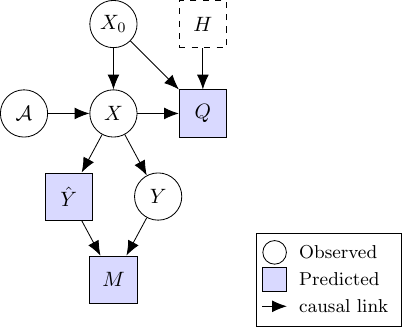}
        }%
        \caption{Causal model for FR-IQA}
        \label{fig:fr-iqa-dag}
     \end{subfigure}
    \caption{Causal model for the NR-/FR-IQA settings. The FR-IQA setting includes $X_0$ which is the reference image.  In both models, $H$ indicates human annotator guidance which reflects that IQ metrics are typically calibrated against human perceptual judgements (dashed-line box indicates $H$ is not used directly in the calculation of $Q$).}
    \label{fig:iqa-dags}
\end{figure}

We provide here the causal models for the NR-IQA and FR-IQA settings in Figure~\ref{fig:iqa-dags}.

While the main manuscript focused on the NR-IQA setting, we can see from the causal models here that the results generalize to the FR-IQA case as well.  In particular, the independence of $Q, M$ given $X$ is not affected by whether a ``clean'' reference image ($X_0$) is available for computing $Q$.  Also, these models also account for the influence of human annotators $H$ in calibrating the function for computing $Q$, but not that this does not change the relationship between $M, Q$.

\section{Causal model for IQA with latent features}
\label{app:latents-dag}
\begin{figure}[h!]
    \centering
    \resizebox{0.6\linewidth}{!}{%
    \includegraphics[width=\linewidth]{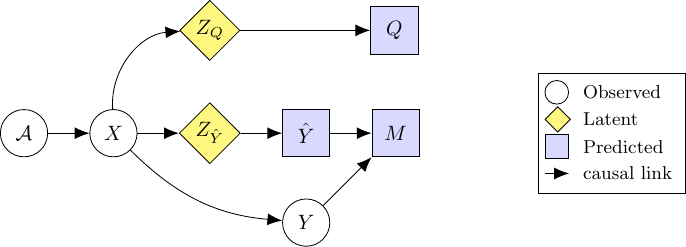}
    }%
    \caption{Causal model for IQA that accounts for the use of latent features by the task DNN and IQ metric towards computing $M$ and $Q$ respectively.}
    \label{fig:standard-latents-dag}
\end{figure}

In understanding the difference between the baseline IQA formulation in Figure~\ref{fig:standard-dag} and the shared features formulation of Figure~\ref{fig:latents-dag} in Section~\ref{sec:causal-iqa}, we provide an expanded version of the baseline DAG in Figure~\ref{fig:standard-latents-dag}.  Here we show that the task DNN for computing $\hat{Y}$ and the function for computing $Q$ rely on latent features $Z_{\hat{Y}}$ and $Z_Q$ respectively.  

In this expanded model, $Z_{\hat{Y}}$ and $Z_Q$ are independent given $X$ and not ``shared'', and therefore $Q \perp M \mid X$ as discussed in \S\ref{sec:causal-iqa}.  In contrast, Figure~\ref{fig:latents-dag} considers the case where $Z$ represents the features derived from $X$ that are common between $Z_{\hat{Y}}$ and $Z_Q$ shown in the baseline case above.  Thus, Figure~\ref{fig:latents-dag} shows the case where $X$ does not block all paths between $Q, M$ since a path exists from $Q$ to $M$ through $Z$.  This ensures that $Q$ and $M$ will be correlated given $X$ unlike in the baseline case above.

\section{Causal model for common corruptions robustness evaluation}
\label{app:common-corruptions}
\begin{figure}[h!]
    \centering
    \resizebox{0.6\linewidth}{!}{%
    \includegraphics[width=\linewidth]{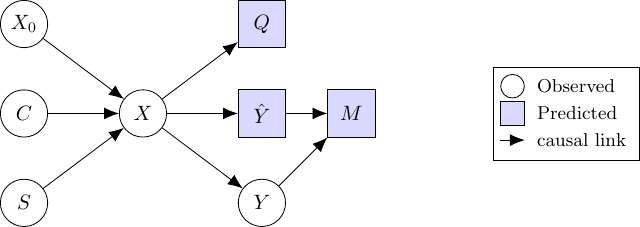}
    }%
    \caption{Causal model for the common corruptions framework where $C$ refers to the corruption type, $S$ refers to the corruption severity, and $X_0$ is the unperturbed, ``clean'' image.}
    \label{fig:common-corruptions-dag}
\end{figure}

The common corruptions framework~\citep{Hendrycks2019-ye} is used in our experiments to ensure full control of the image distortion types and severity.  Figure~\ref{fig:common-corruptions-dag} shows a version of the baseline IQA causal model customized to account for the corruption process used by this evaluation framework.  Here, the corrupted image $X$ is determined by the corruption function (e.g., Gaussian noise, defocus blur, fog, contrast, brightness, JPEG compression), the severity ($S \in \{1, 2, 3, 4, 5\}$), and the ``clean'' image $X_0$.

In this setting, we see that $C, S$ replace the original set of imaging factors $\gA$ in the graph in Figure~\ref{fig:standard-dag}. As such, the analysis from \S\ref{sec:causal-iqa} holds in the common corruptions framework and allows us to study the relationship between $Q, M$ in a setting where we can precisely control the imaging conditions.

\section{Relationship of NR-IQA and DNN performance metrics}
\begin{figure}[h!]
    \centering
    \includegraphics[width=0.48\linewidth]{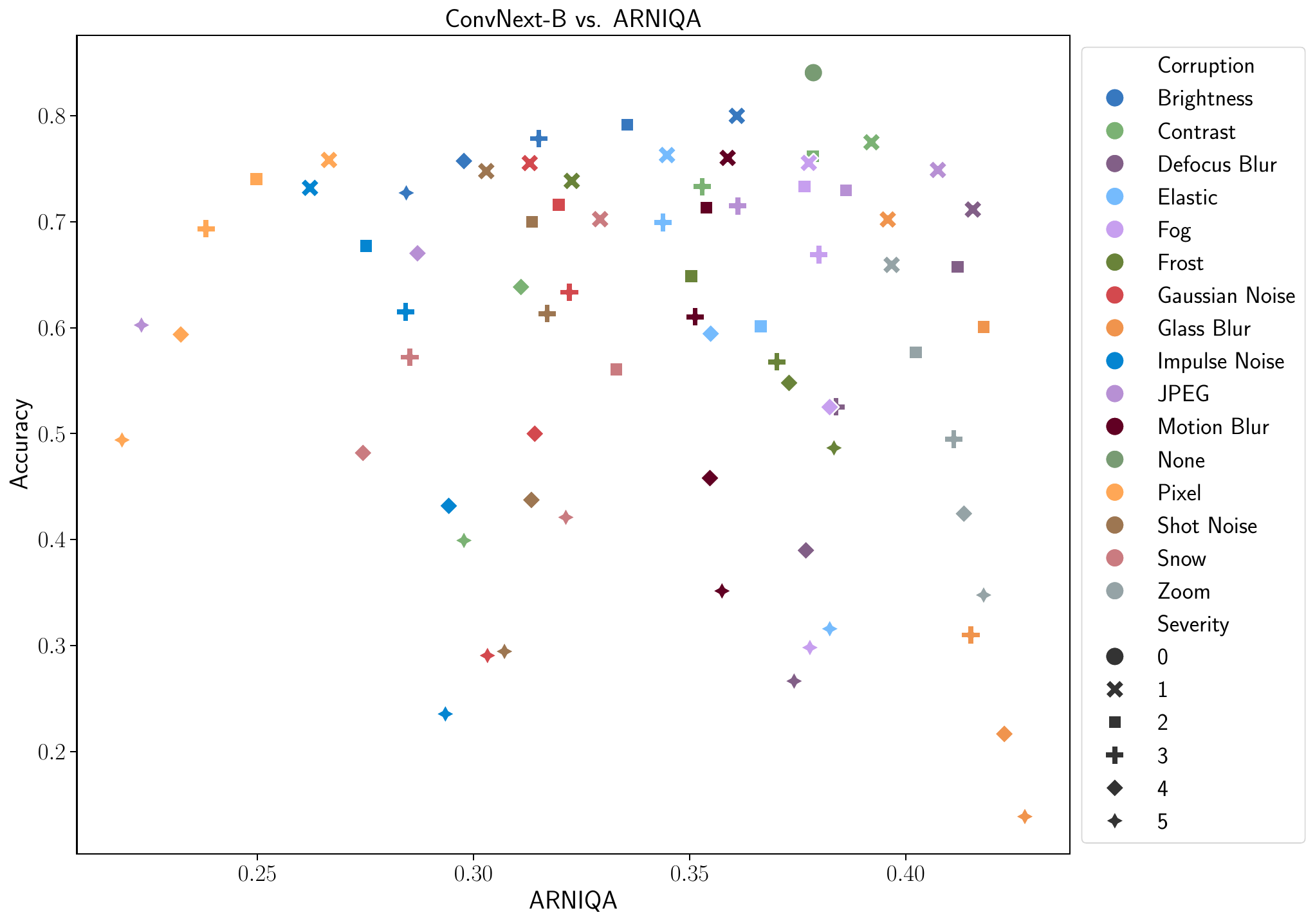} 
    \includegraphics[width=0.48\linewidth]{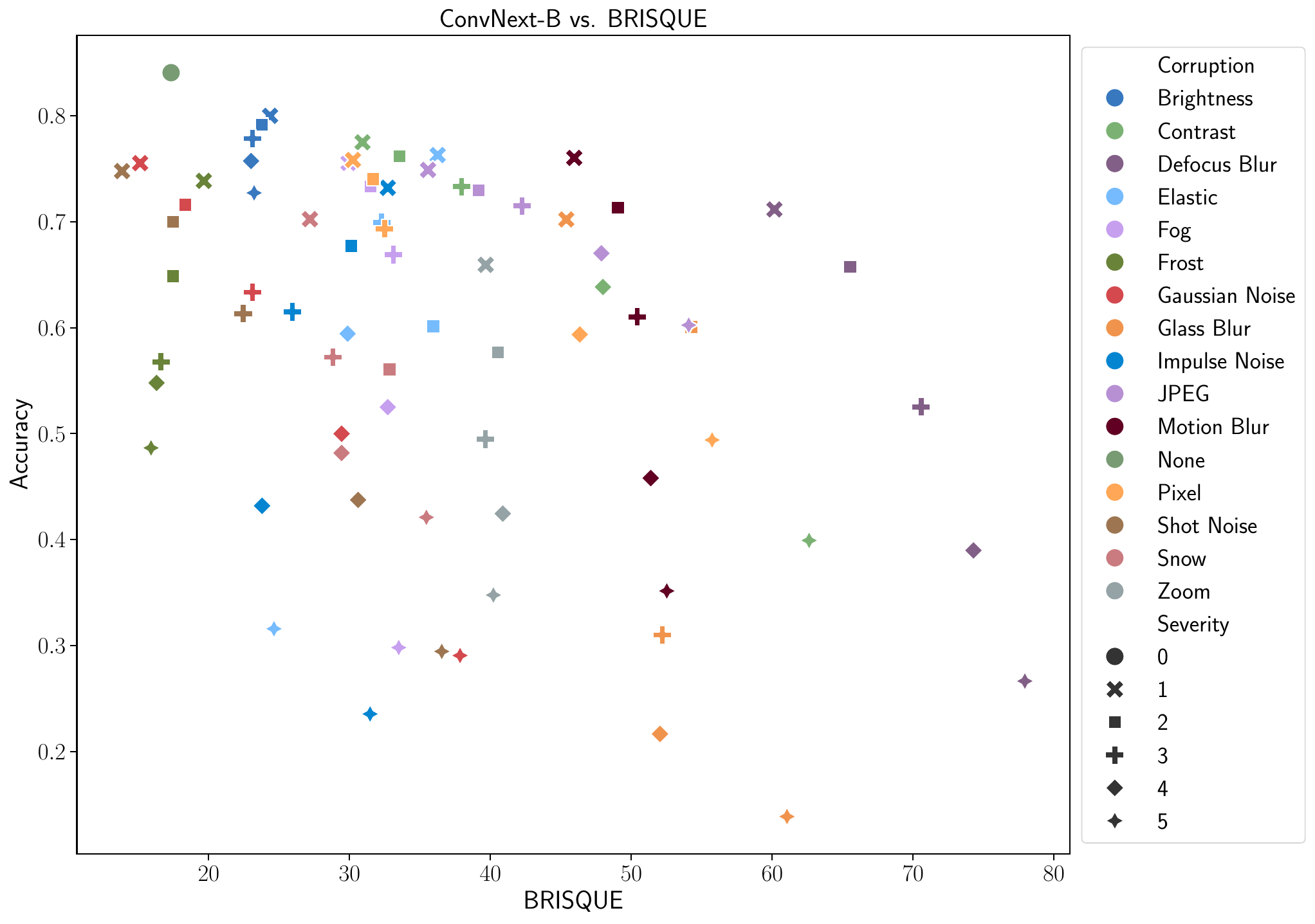} \\
    \includegraphics[width=0.48\linewidth]{figures/Figure-3-acc-convnext-clipiqa.pdf} 
    \includegraphics[width=0.48\linewidth]{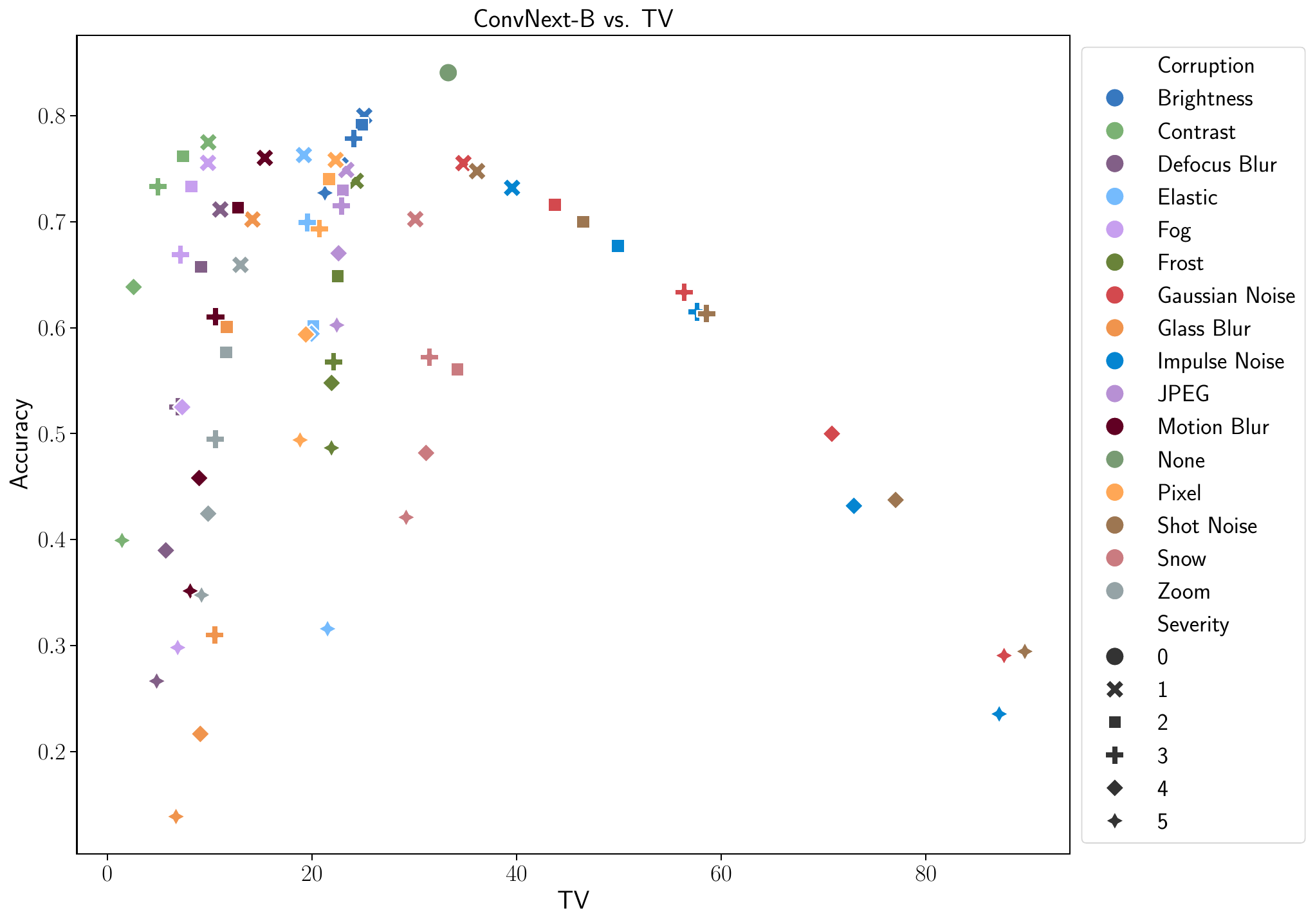} \\
    \caption{Comparison of ConvNext-B accuracy with (clockwise) ARNIQA, BRISQUE, CLIP-IQA, and TV. Little correlation is observed between group-wise accuracy and each NR-IQA metric.}
    \label{fig:app-convnext}
\end{figure}

\begin{figure}[h!]
    \centering
    \includegraphics[width=0.48\linewidth]{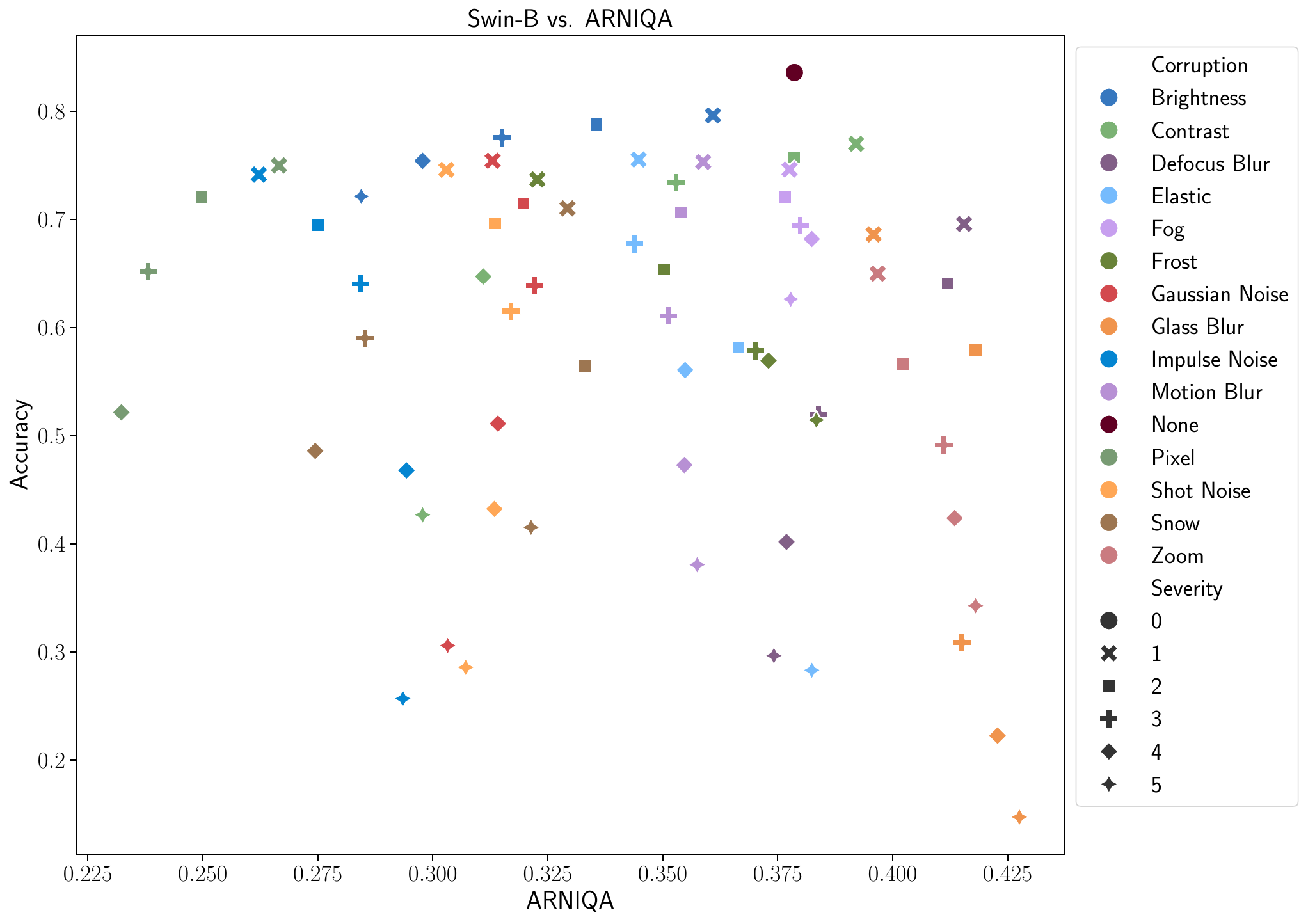} 
    \includegraphics[width=0.48\linewidth]{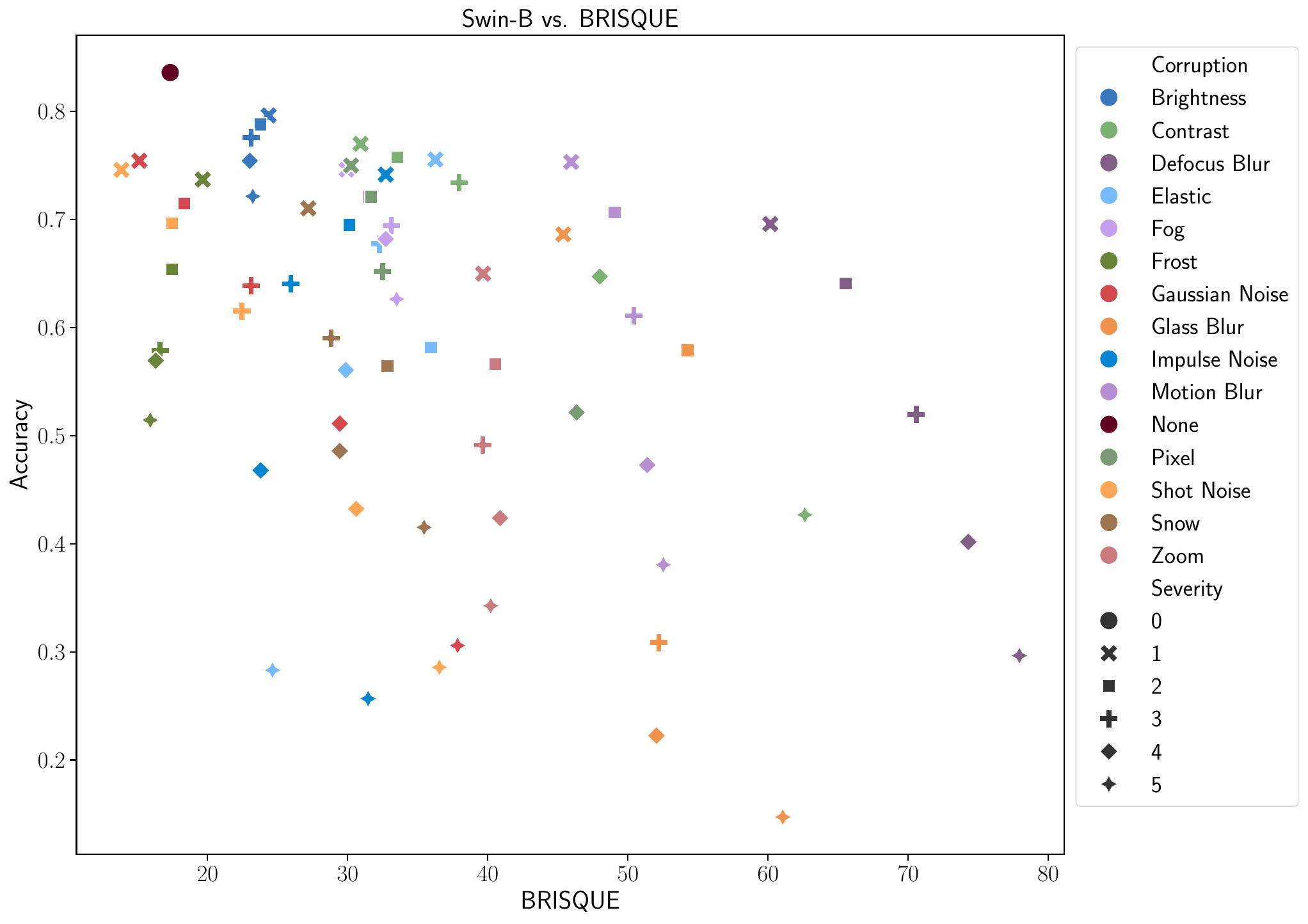} \\
    \includegraphics[width=0.48\linewidth]{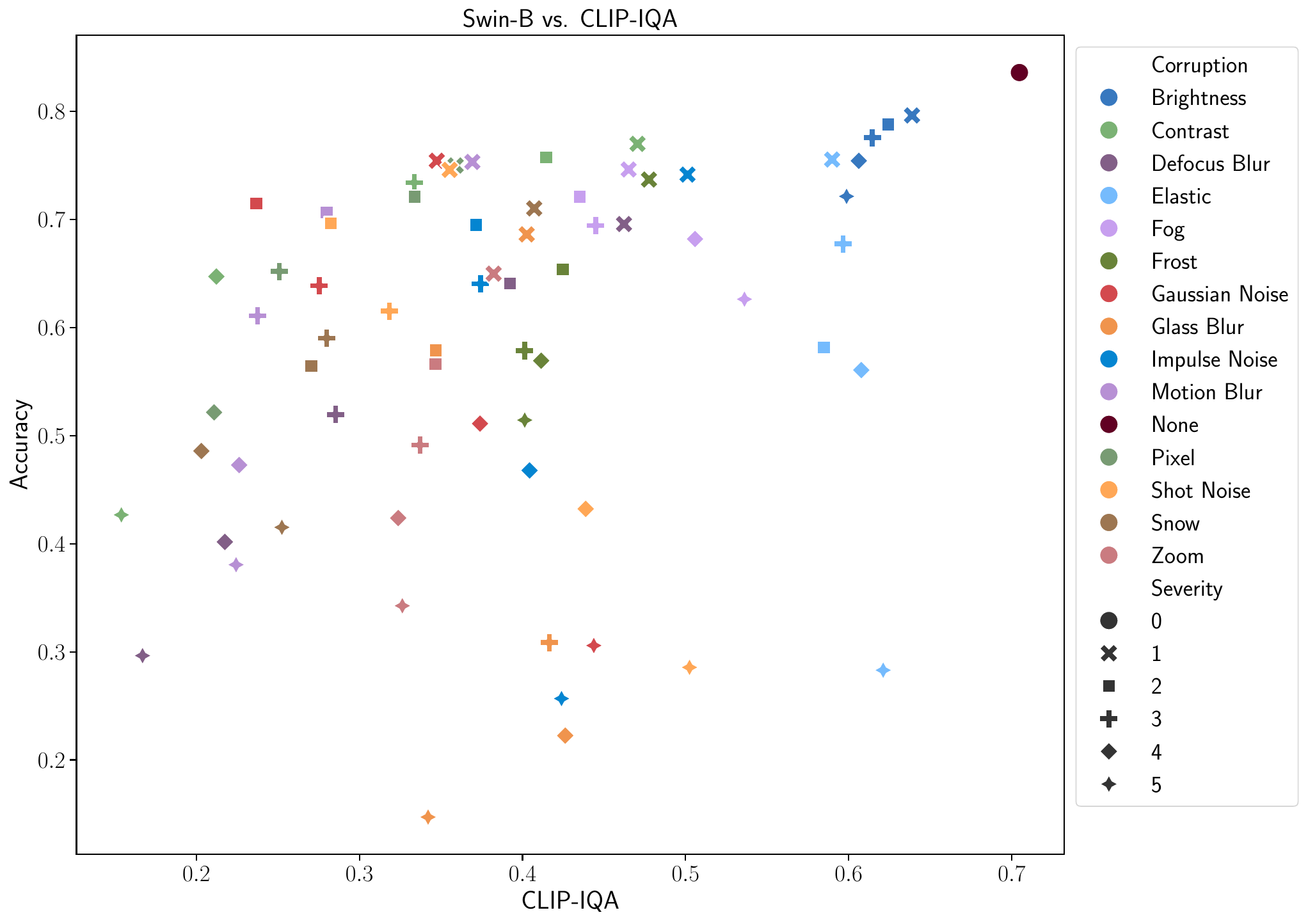}
    \includegraphics[width=0.48\linewidth]{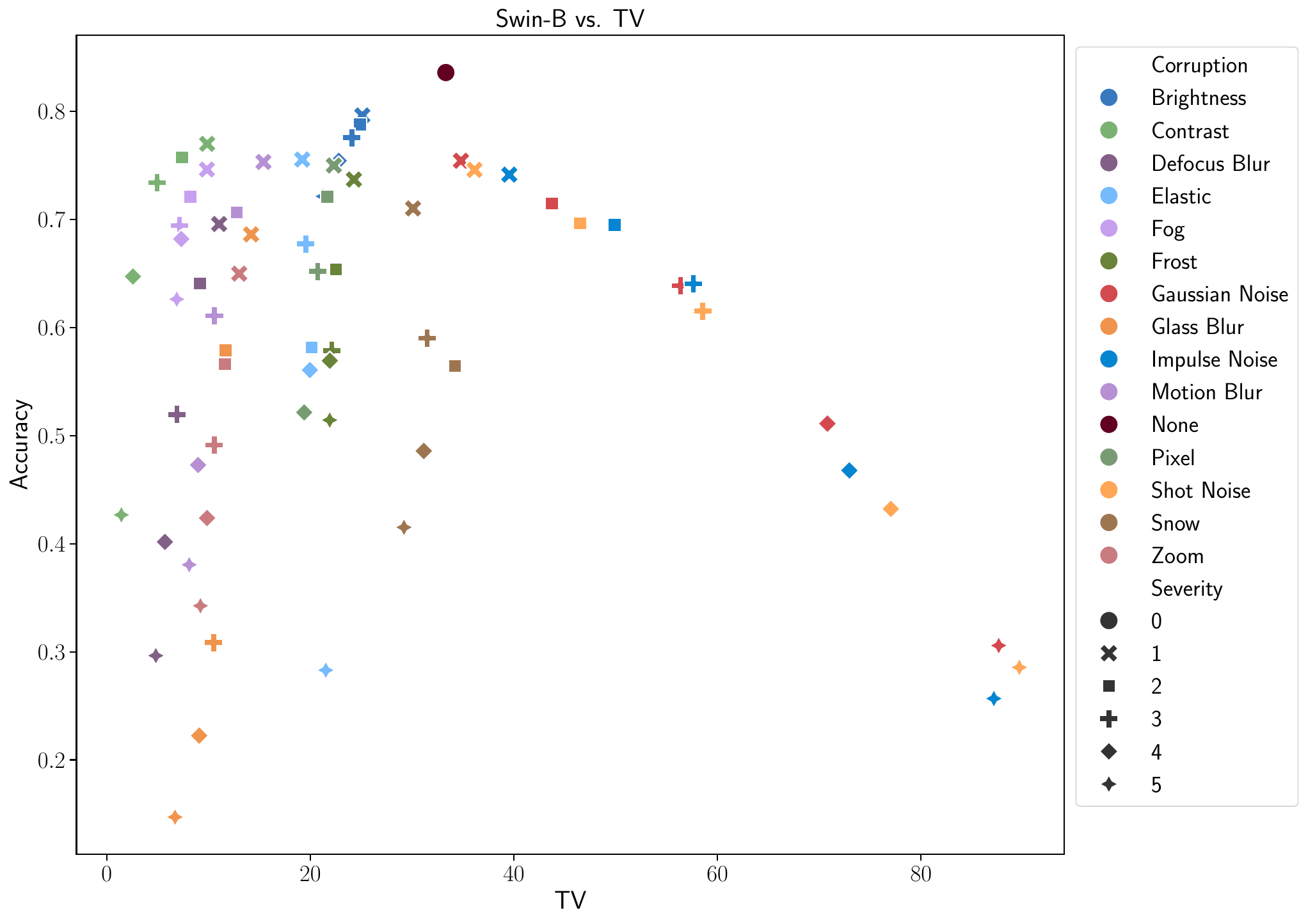}
    \caption{Comparison of Swin-B accuracy with (clockwise) ARNIQA, BRISQUE, CLIP-IQA, and TV. Little correlation is observed between group-wise accuracy and each NR-IQA metric.}
    \label{fig:app-swin}
\end{figure}

\begin{figure}[h!]
    \centering
    \includegraphics[width=0.48\linewidth]{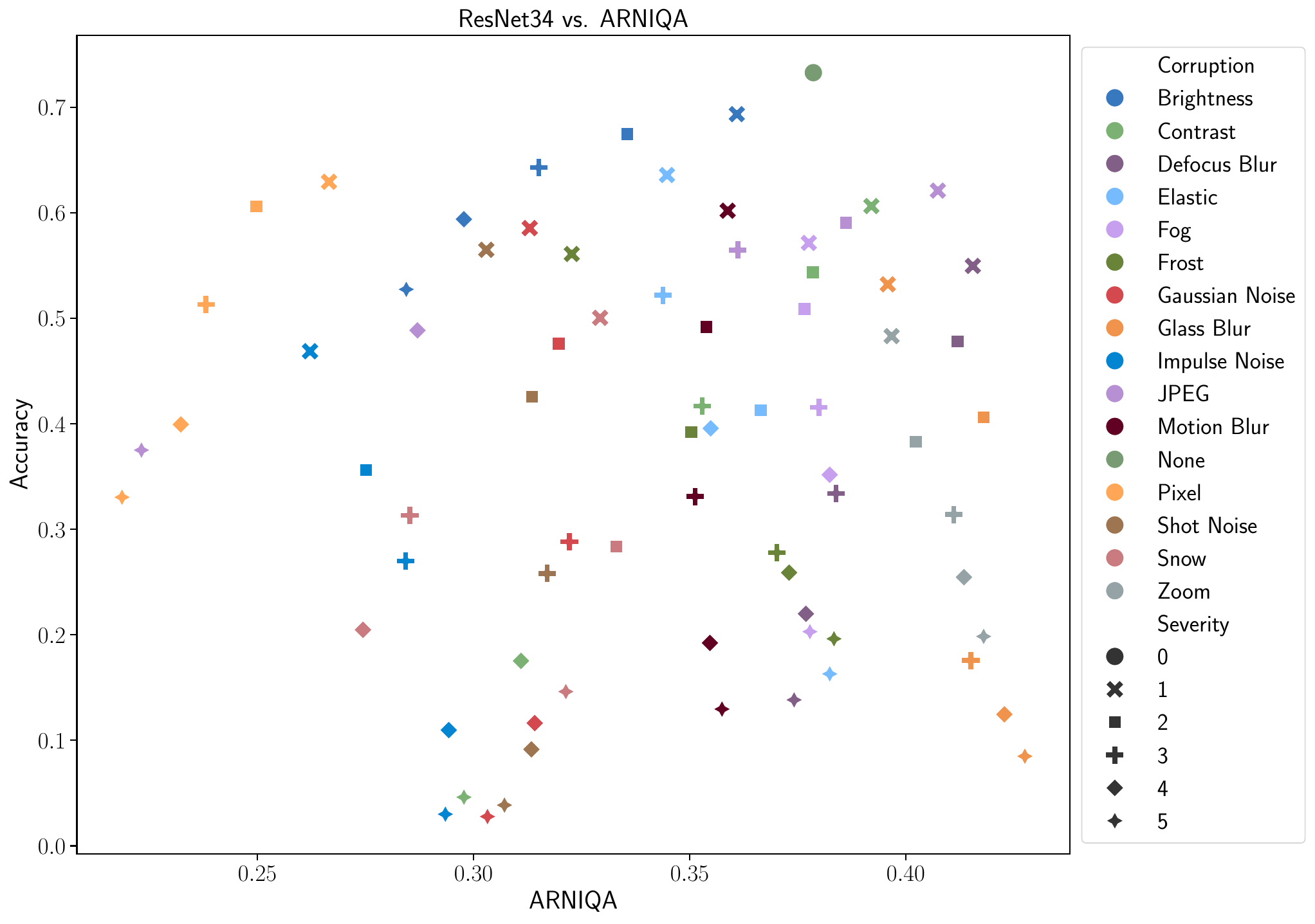}
    \includegraphics[width=0.48\linewidth]{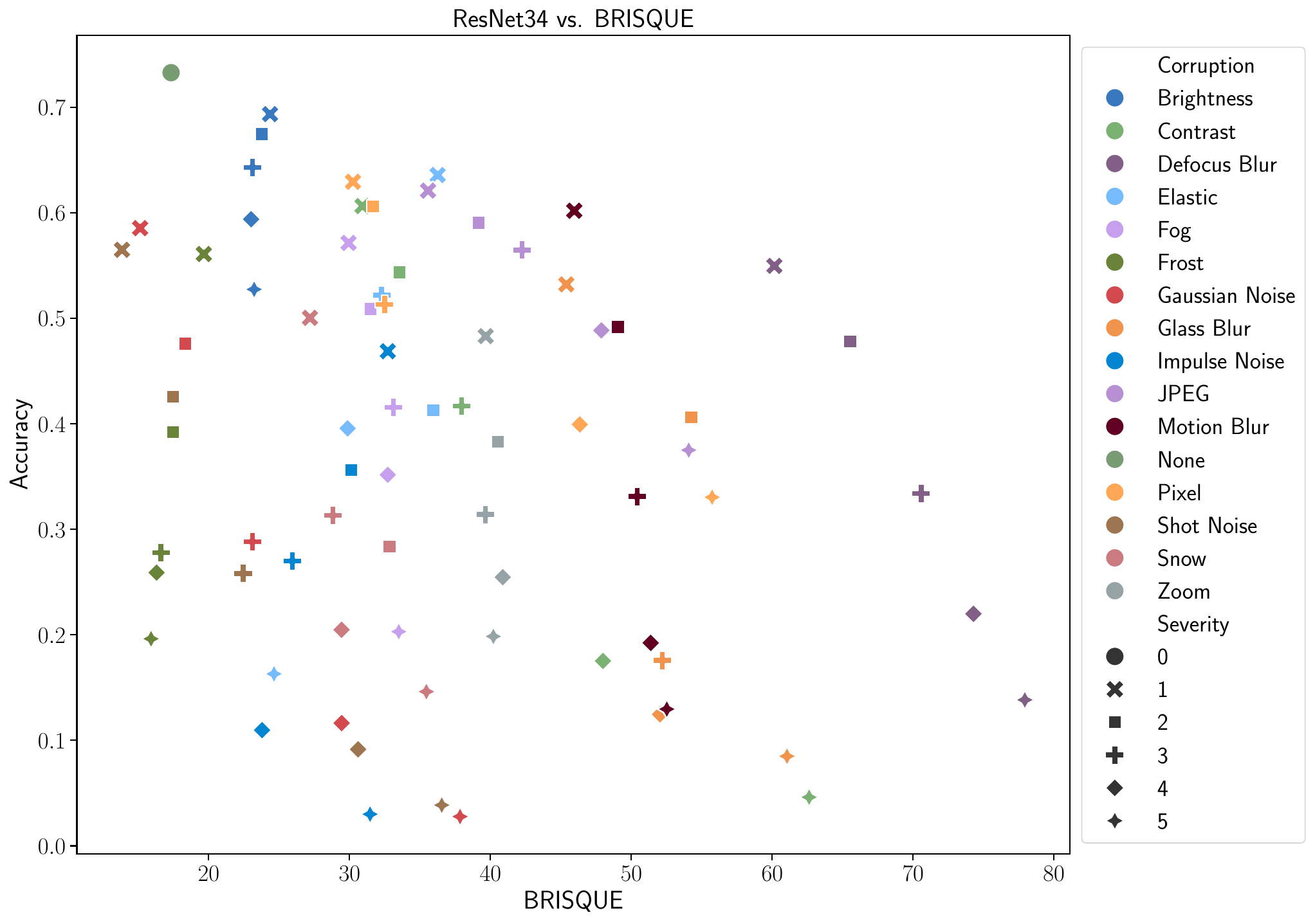} \\
    \includegraphics[width=0.48\linewidth]{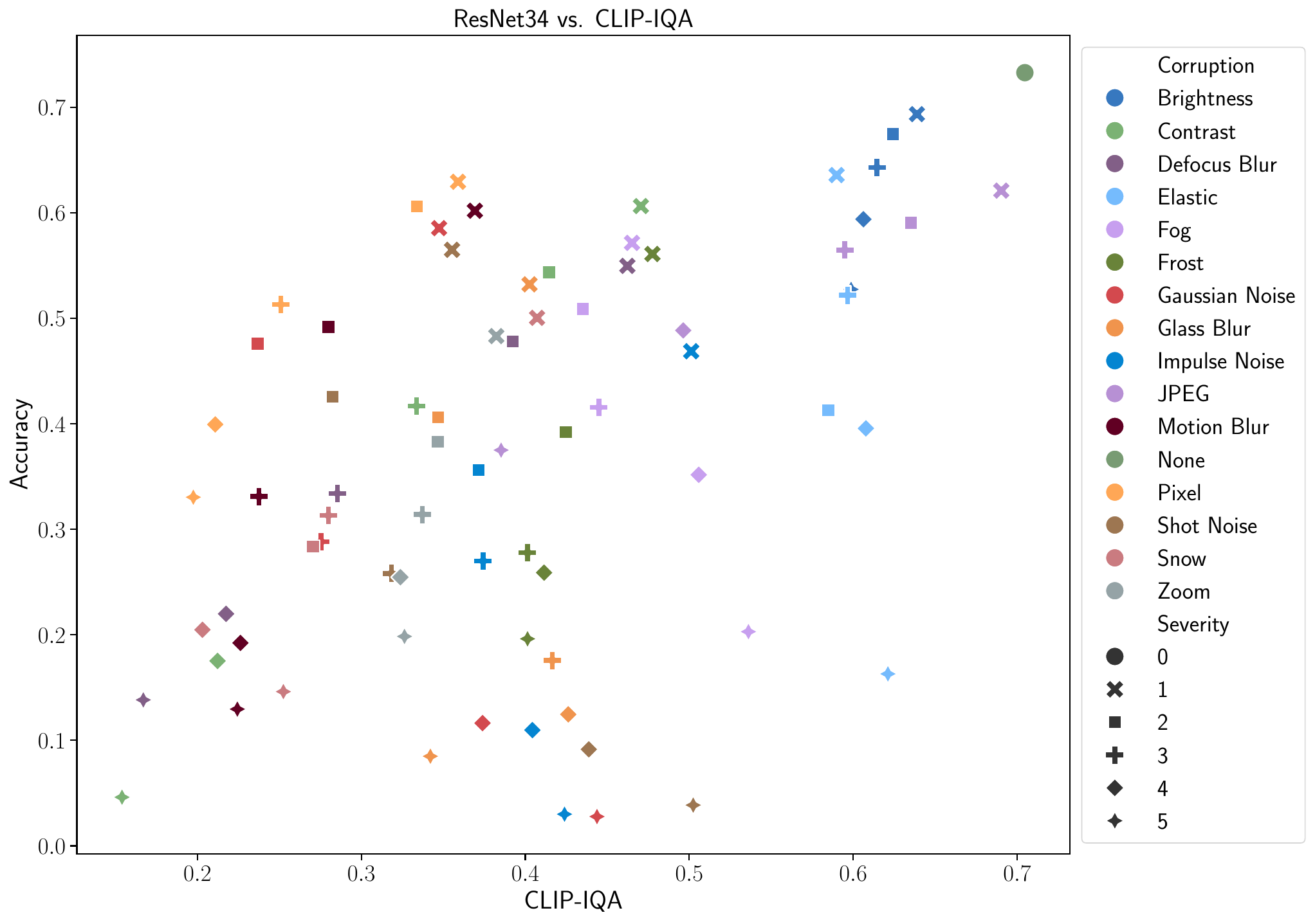}
    \includegraphics[width=0.48\linewidth]{figures/supplemental/exp1/acc-convnext-tv.pdf} \\
    \caption{Comparison of ResNet34 accuracy with (clockwise) ARNIQA, BRISQUE, CLIP-IQA, and TV. Little correlation is observed between group-wise accuracy and each NR-IQA metric.}
    \label{fig:app-resnet}
\end{figure}

In Figures~\ref{fig:app-convnext},~\ref{fig:app-swin}, and~\ref{fig:app-resnet} we show the relationship of additional NR-IQA metrics with DNN performance for additional architectures and metrics.  In general, we see weak trends in accuracy vs. average IQ suggesting that these metrics are most consistent with the causal model in Figure~\ref{fig:standard-dag}.

\section{Relationship of strong task-guided IQA and DNN performance metrics}
\label{app:strong-tg-iqa}
\begin{figure}[h!]
    \centering
    \includegraphics[width=0.48\linewidth]{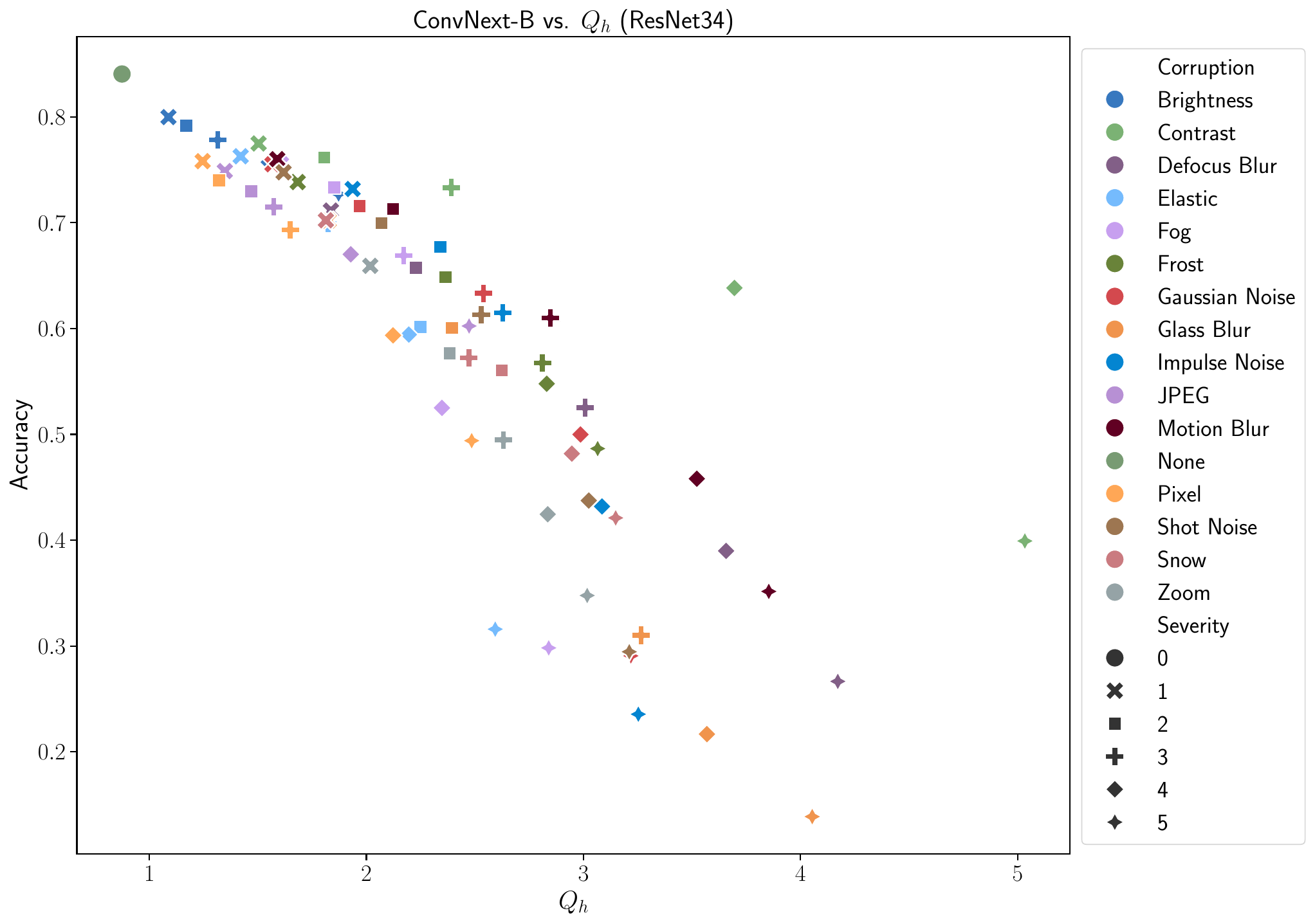}
    \includegraphics[width=0.48\linewidth]{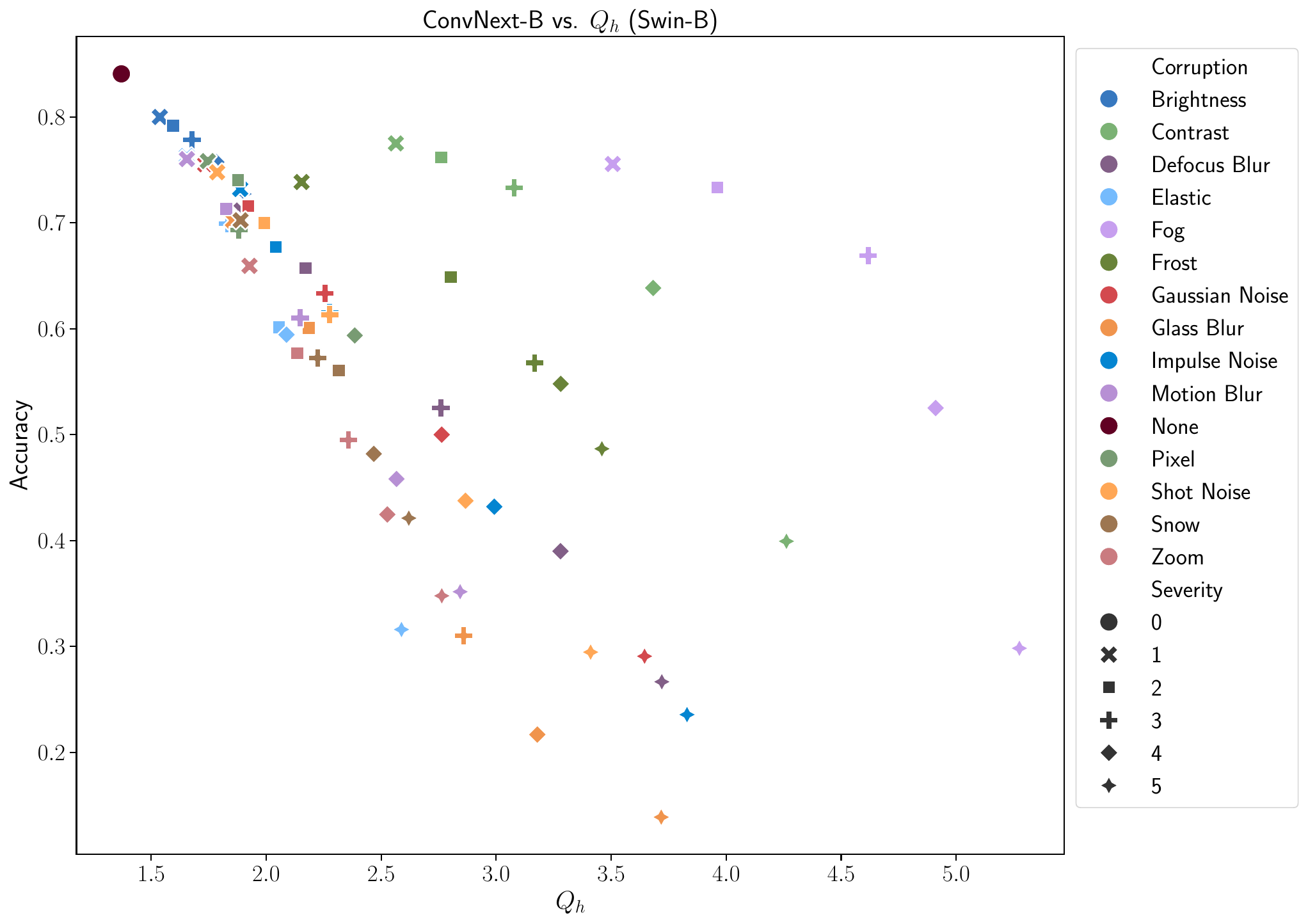} \\
    \includegraphics[width=0.48\linewidth]{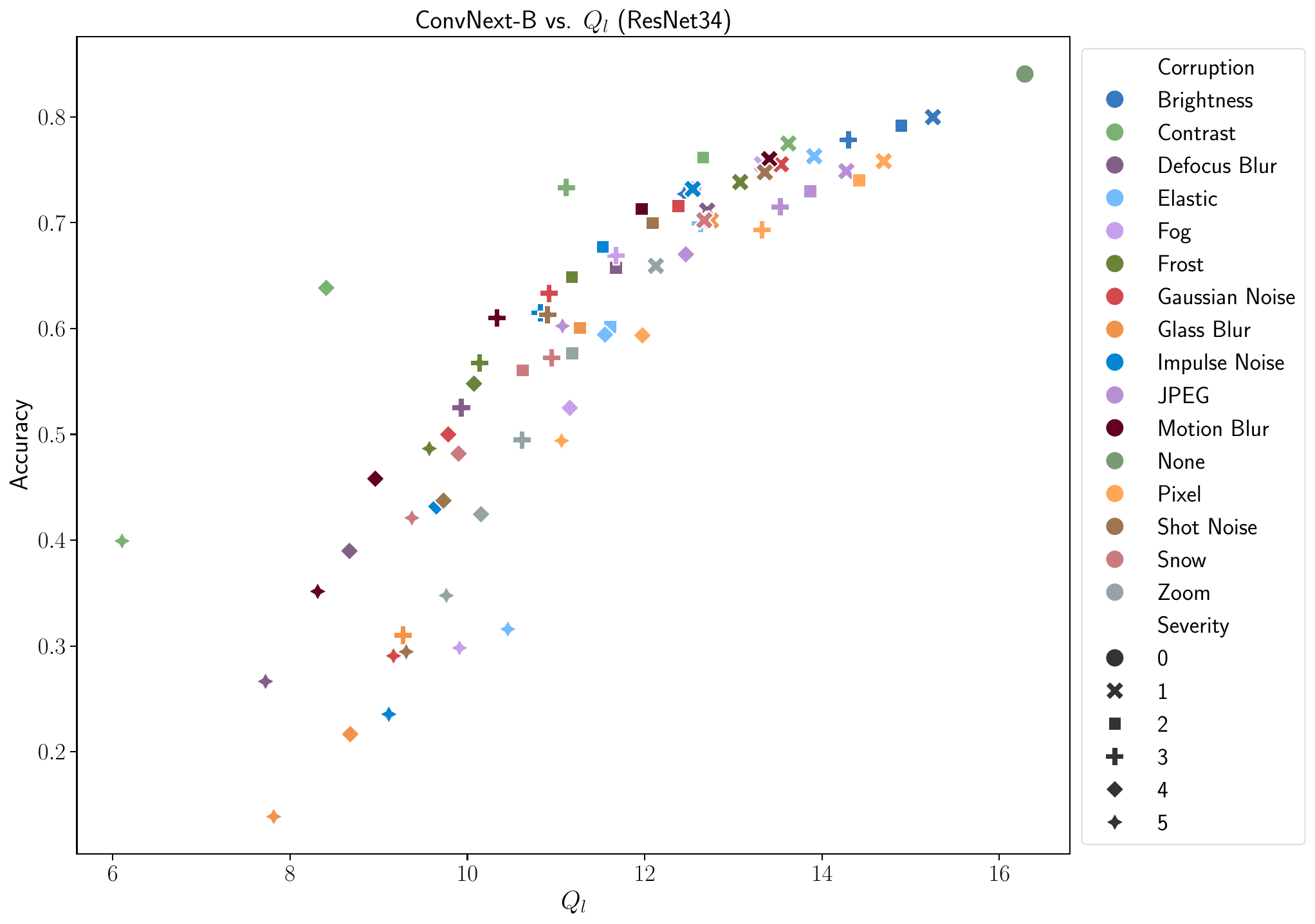}
    \includegraphics[width=0.48\linewidth]{figures/Figure-5-acc-convnext-logit-swin.pdf} \\
    \includegraphics[width=0.48\linewidth]{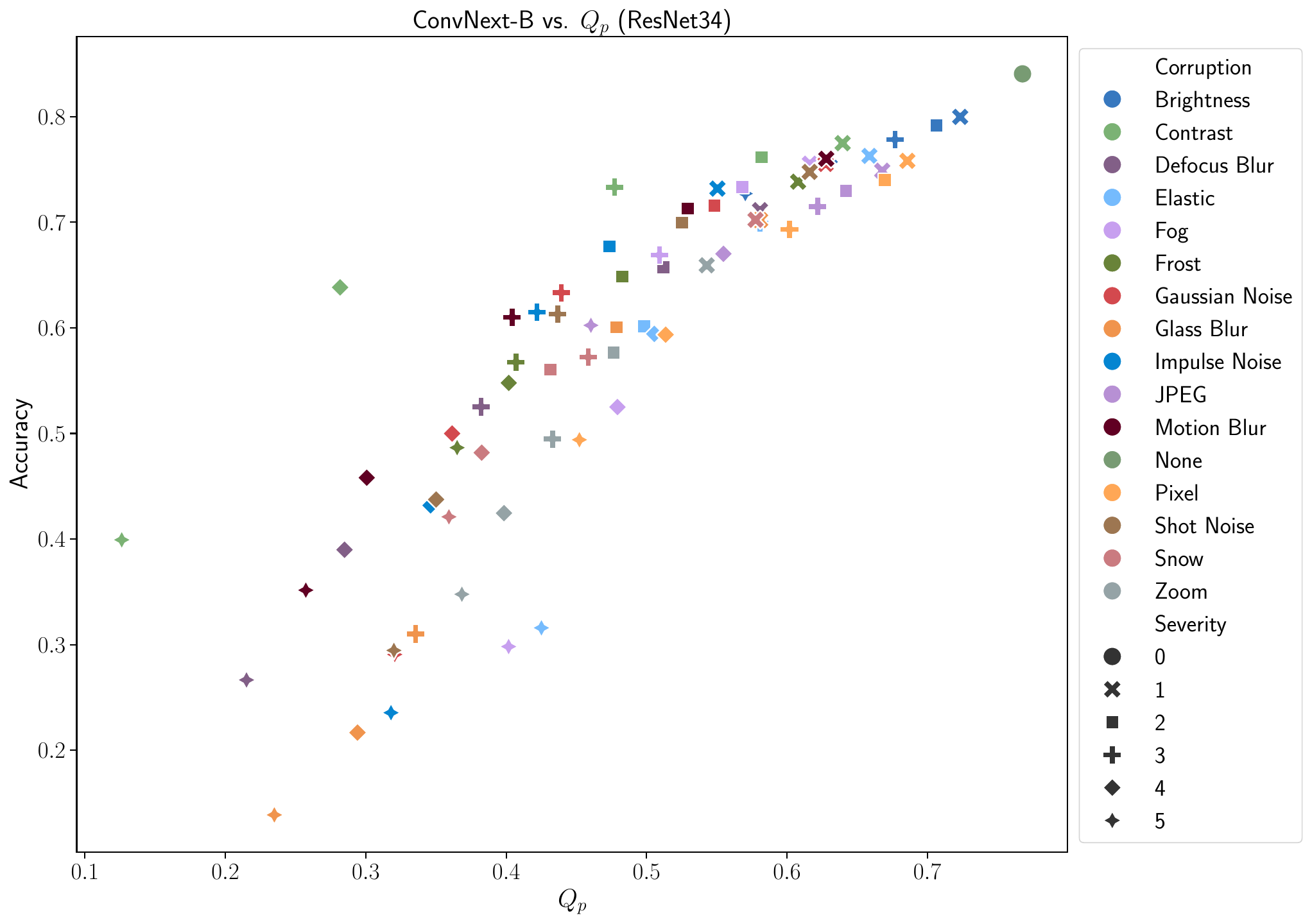}
    \includegraphics[width=0.48\linewidth]{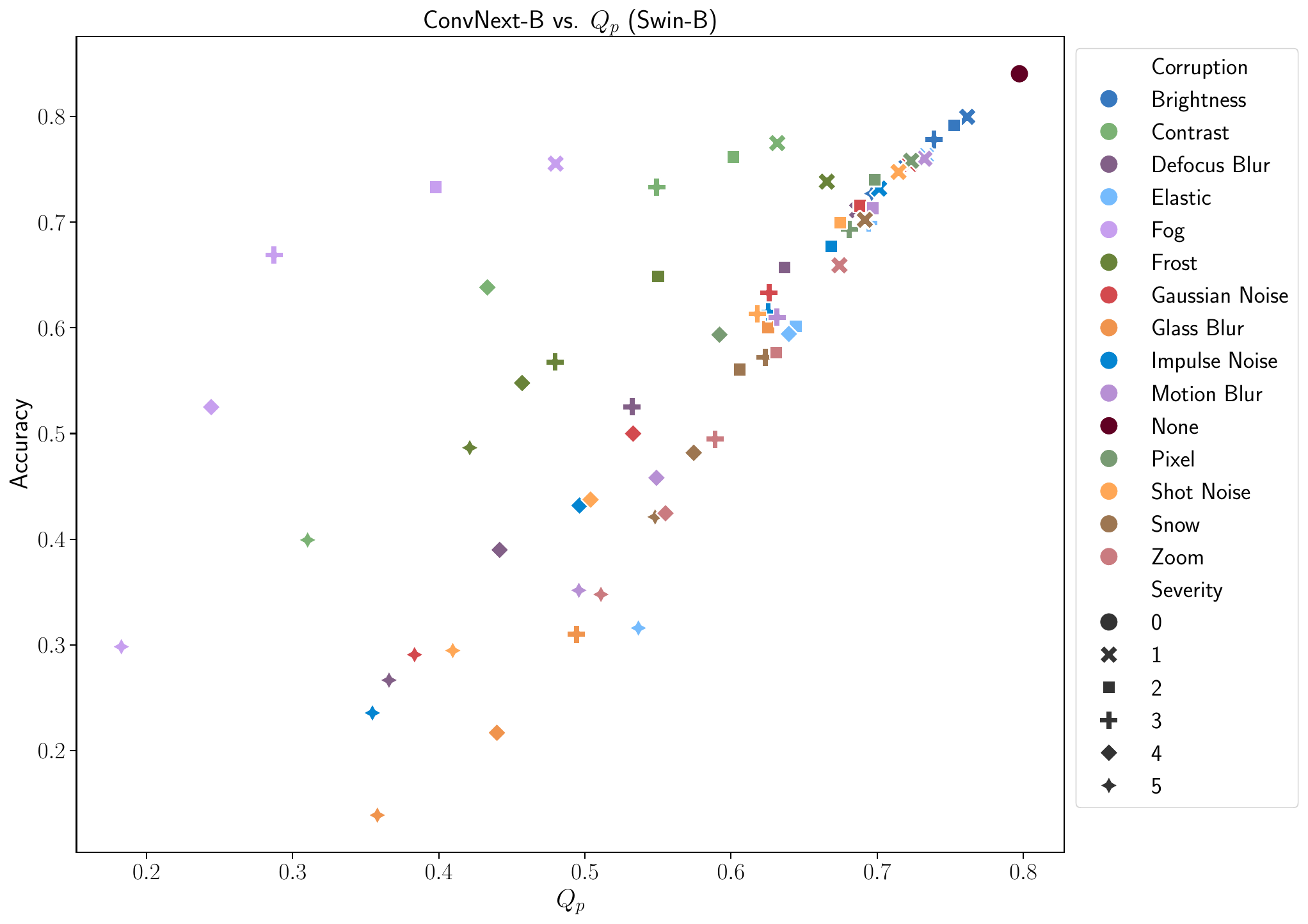} \\
    \caption{Comparison of ConvNext-B accuracy with (row) $Q_h, Q_l, Q_p$ computed using (col) ResNet34, Swin-B.  High correlation is observed between each IQA metric and accuracy. }
    \label{fig:app-stg-convnext}
\end{figure}

\begin{figure}[h!]
    \centering
    \includegraphics[width=0.48\linewidth]{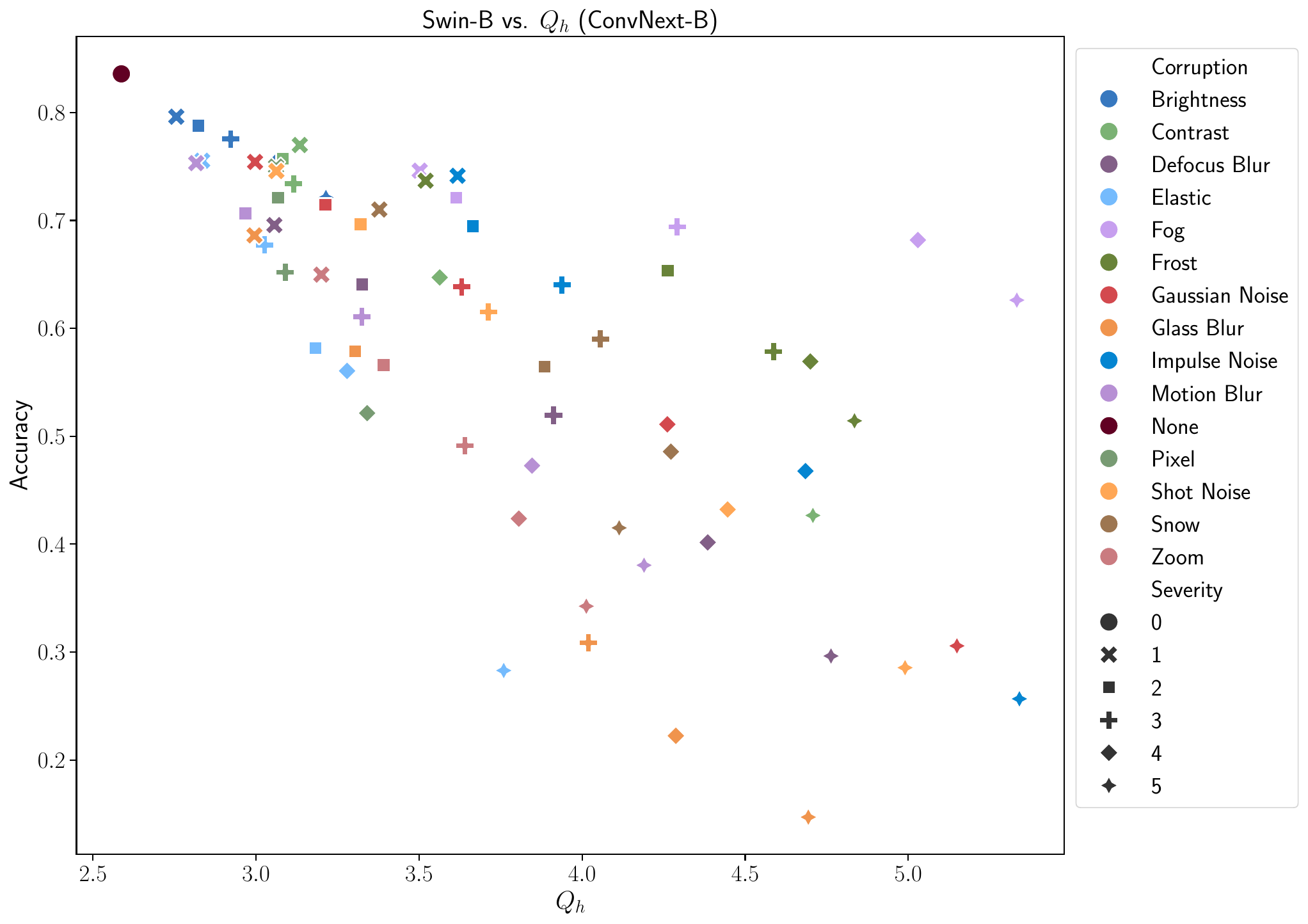}
    \includegraphics[width=0.48\linewidth]{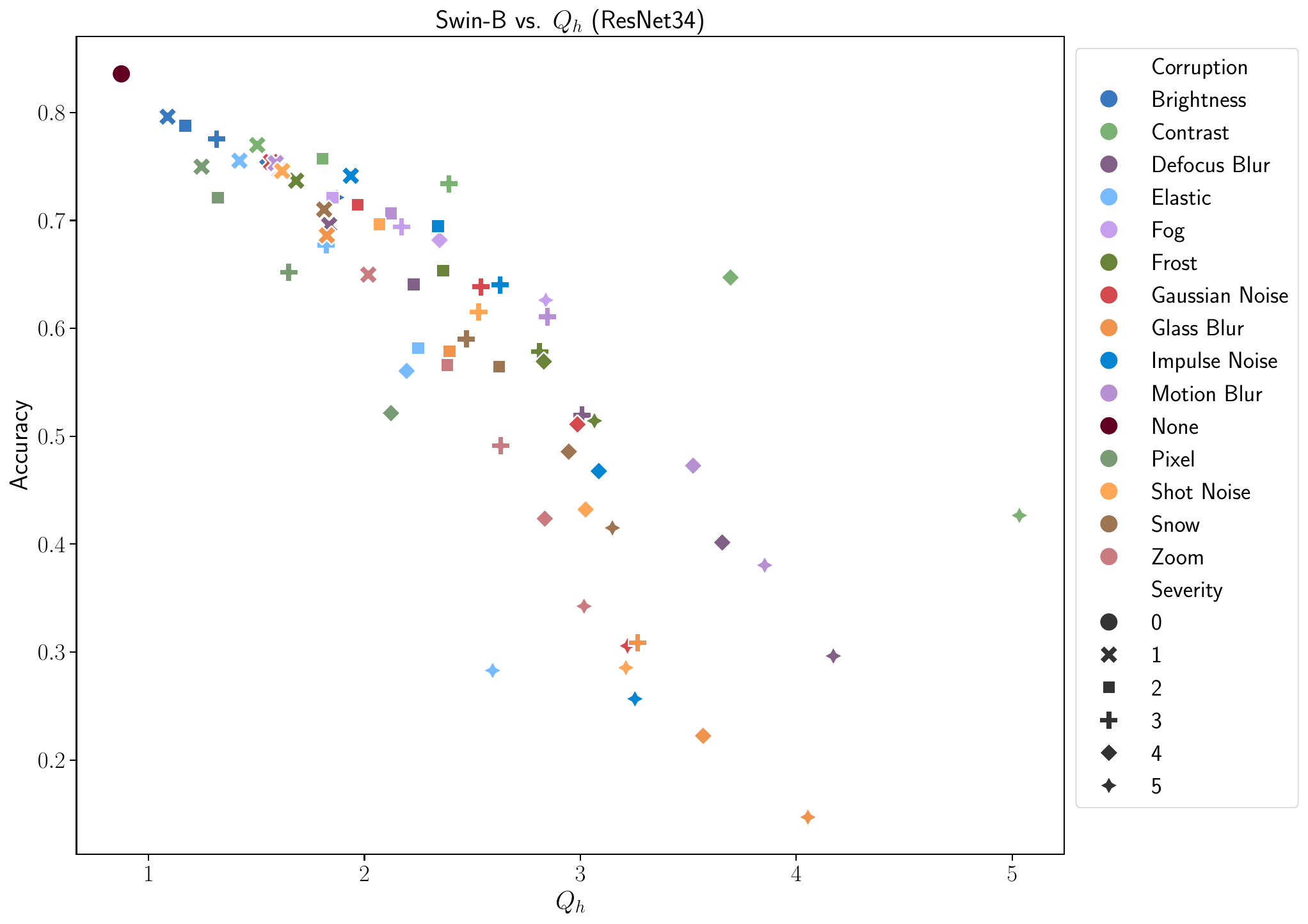} \\
    \includegraphics[width=0.48\linewidth]{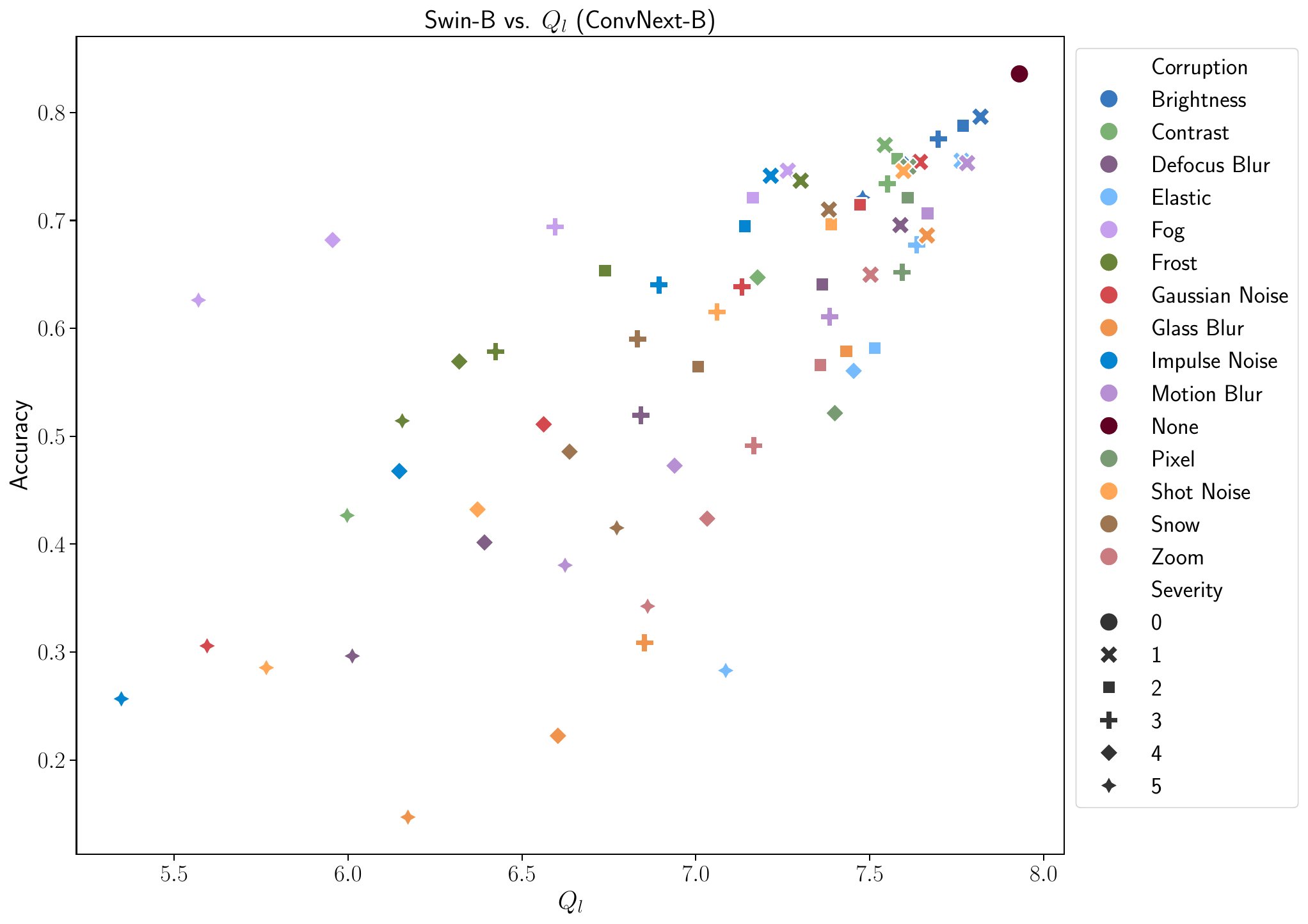}
    \includegraphics[width=0.48\linewidth]{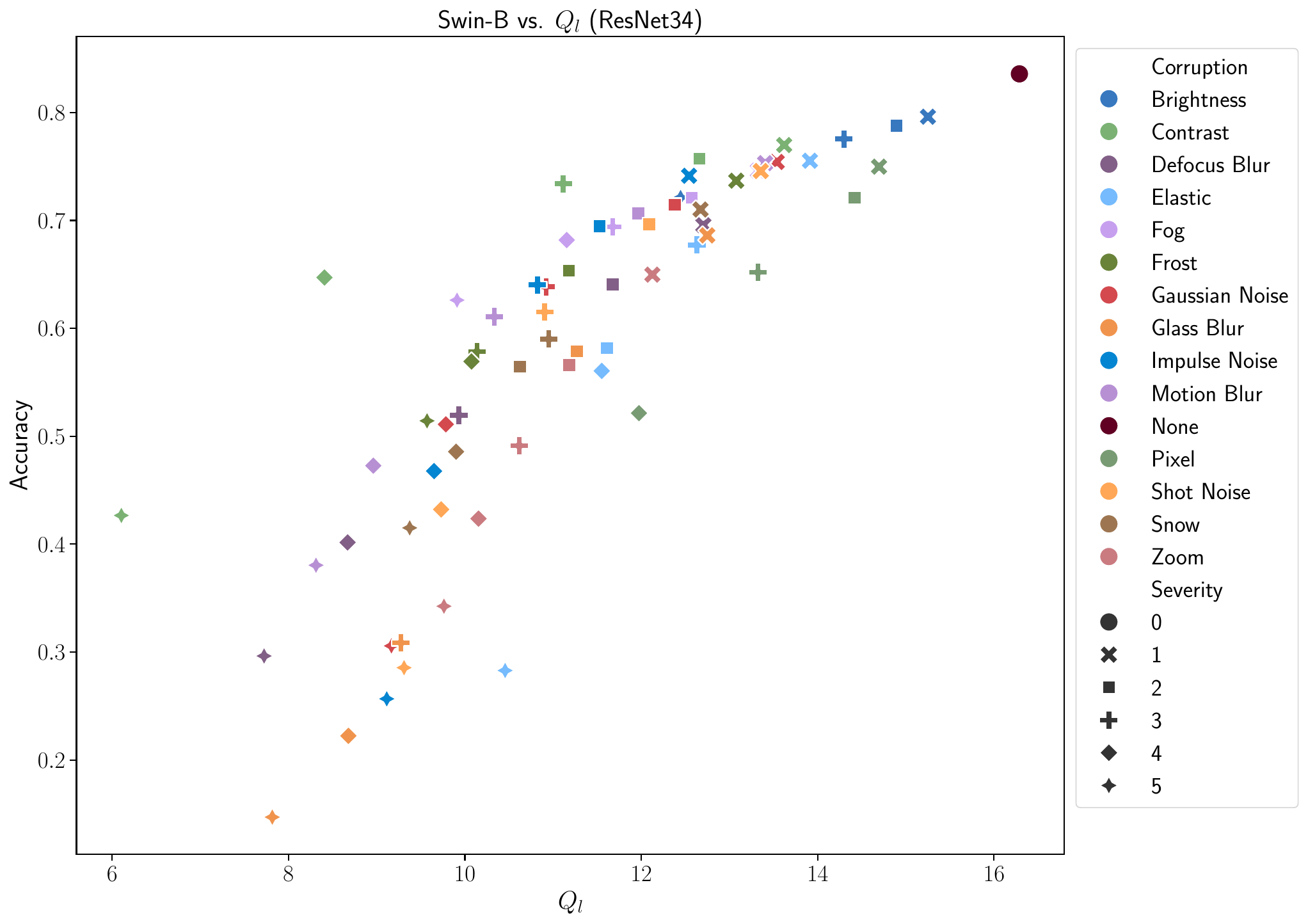} \\
    \includegraphics[width=0.48\linewidth]{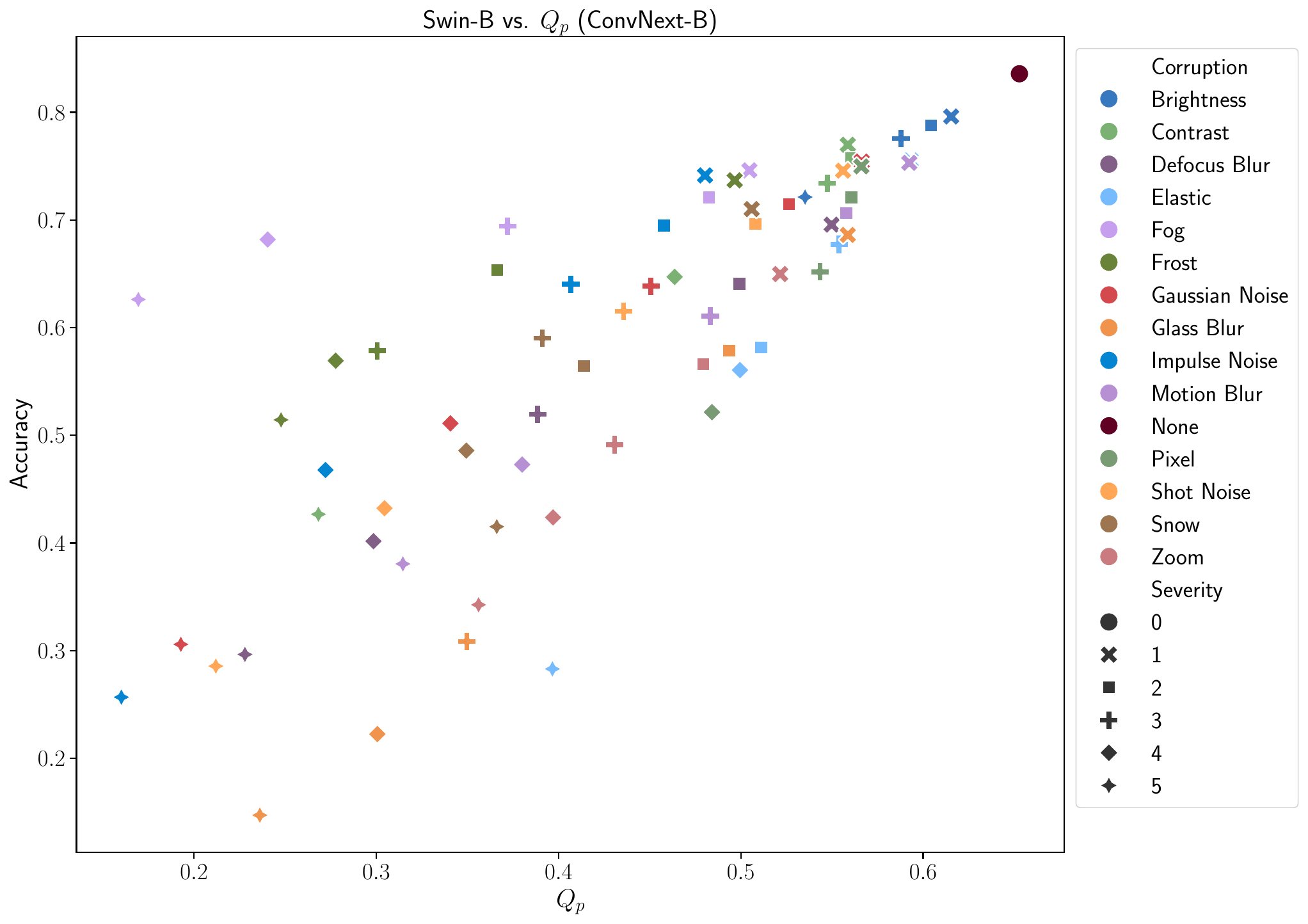}
    \includegraphics[width=0.48\linewidth]{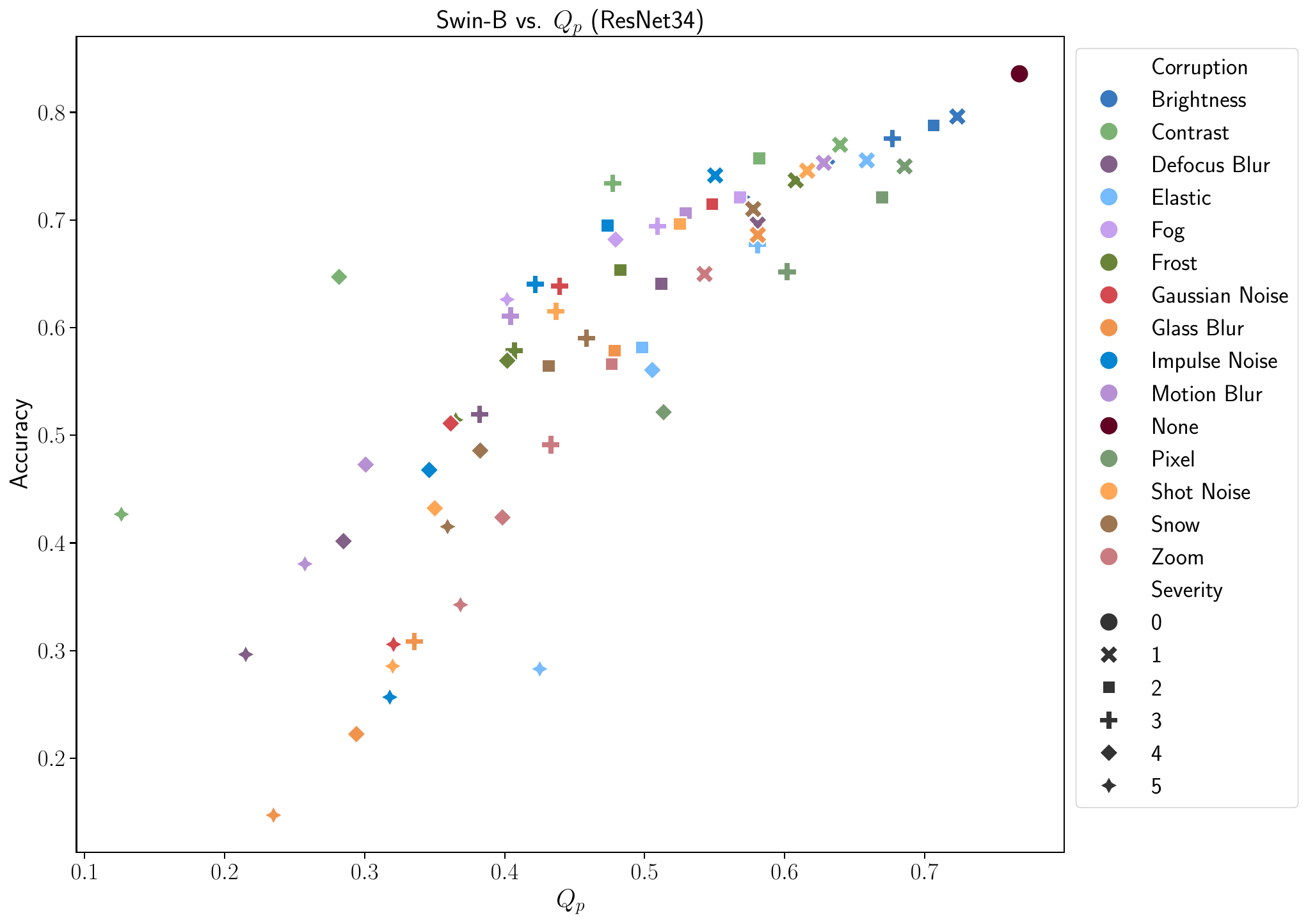} \\
    \caption{Comparison of Swin-B accuracy with (row) $Q_h, Q_l, Q_p$ computed using (col) ConvNext-B, ResNet34. High correlation is observed between each IQA metric and accuracy. }
    \label{fig:app-stg-swin}
\end{figure}

\begin{figure}[h!]
    \centering
    \includegraphics[width=0.48\linewidth]{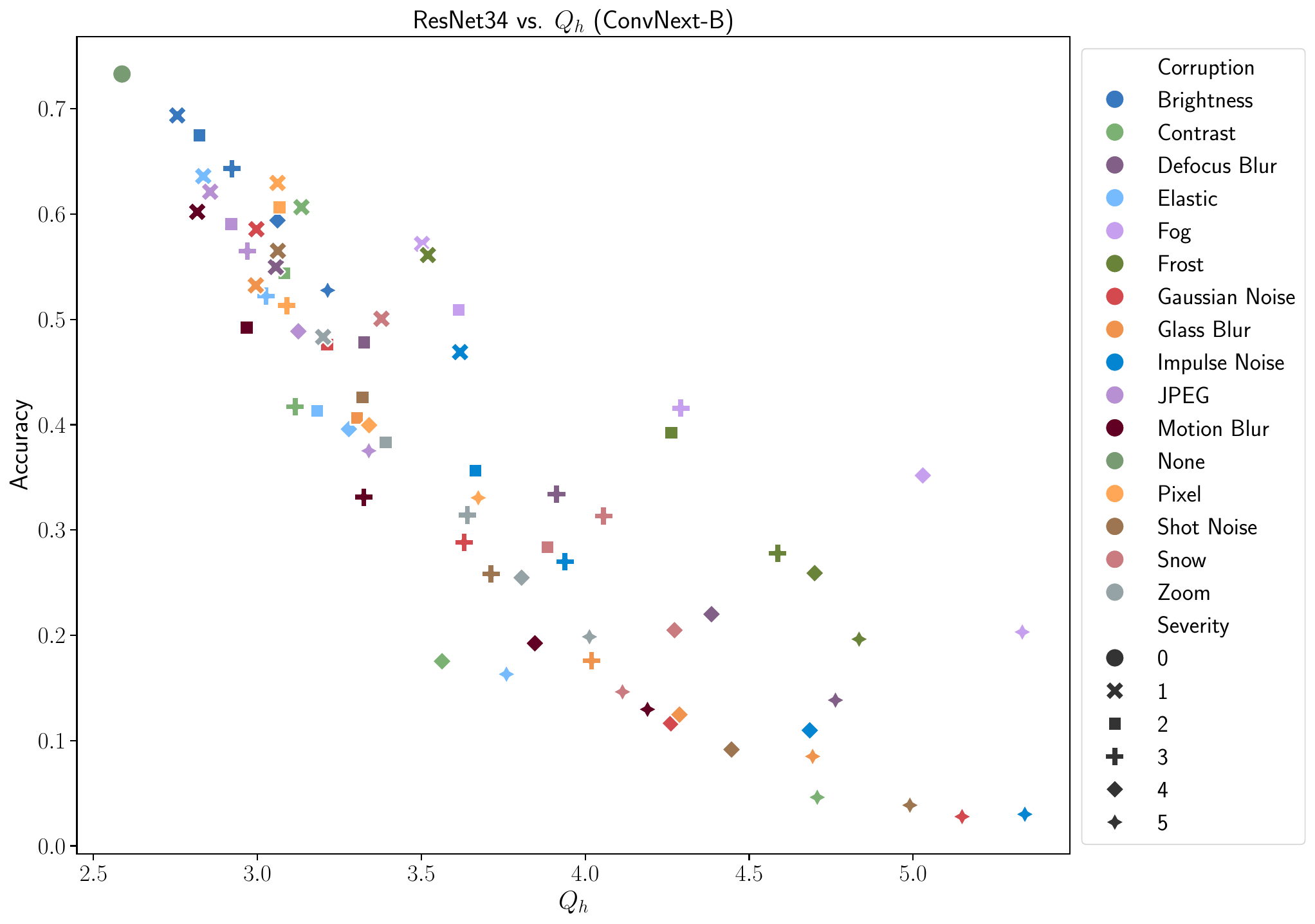}
    \includegraphics[width=0.48\linewidth]{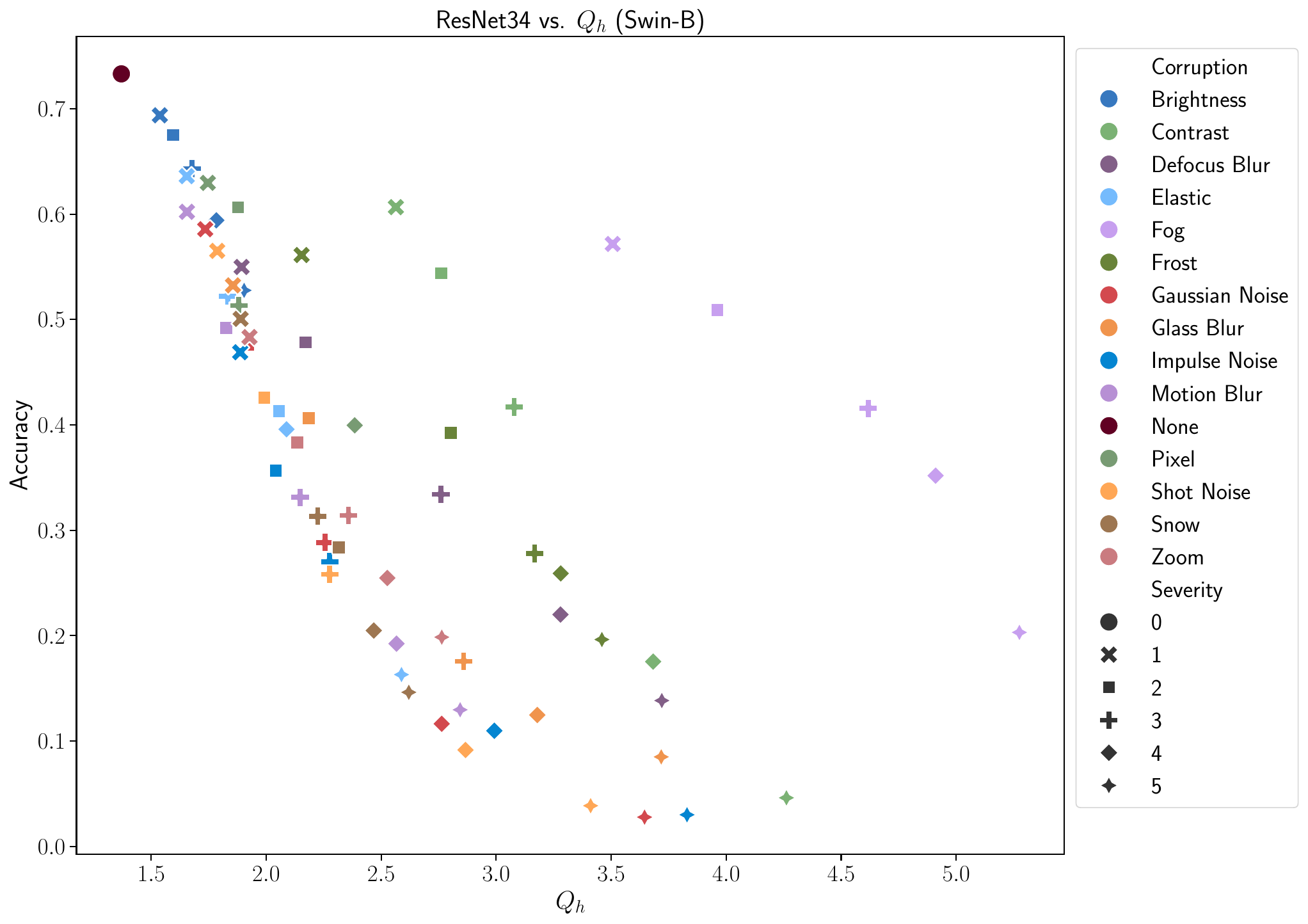} \\
    \includegraphics[width=0.48\linewidth]{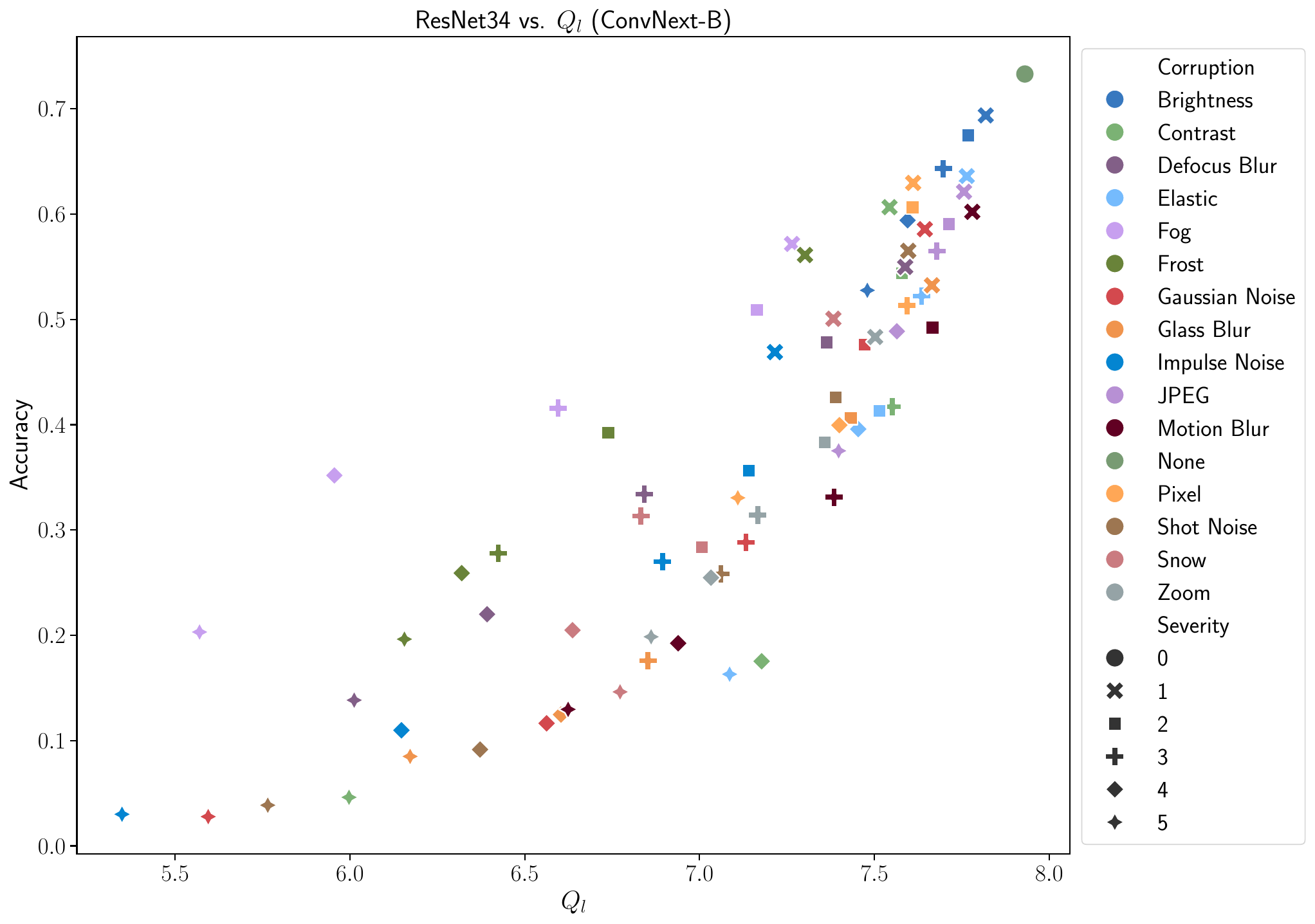}
    \includegraphics[width=0.48\linewidth]{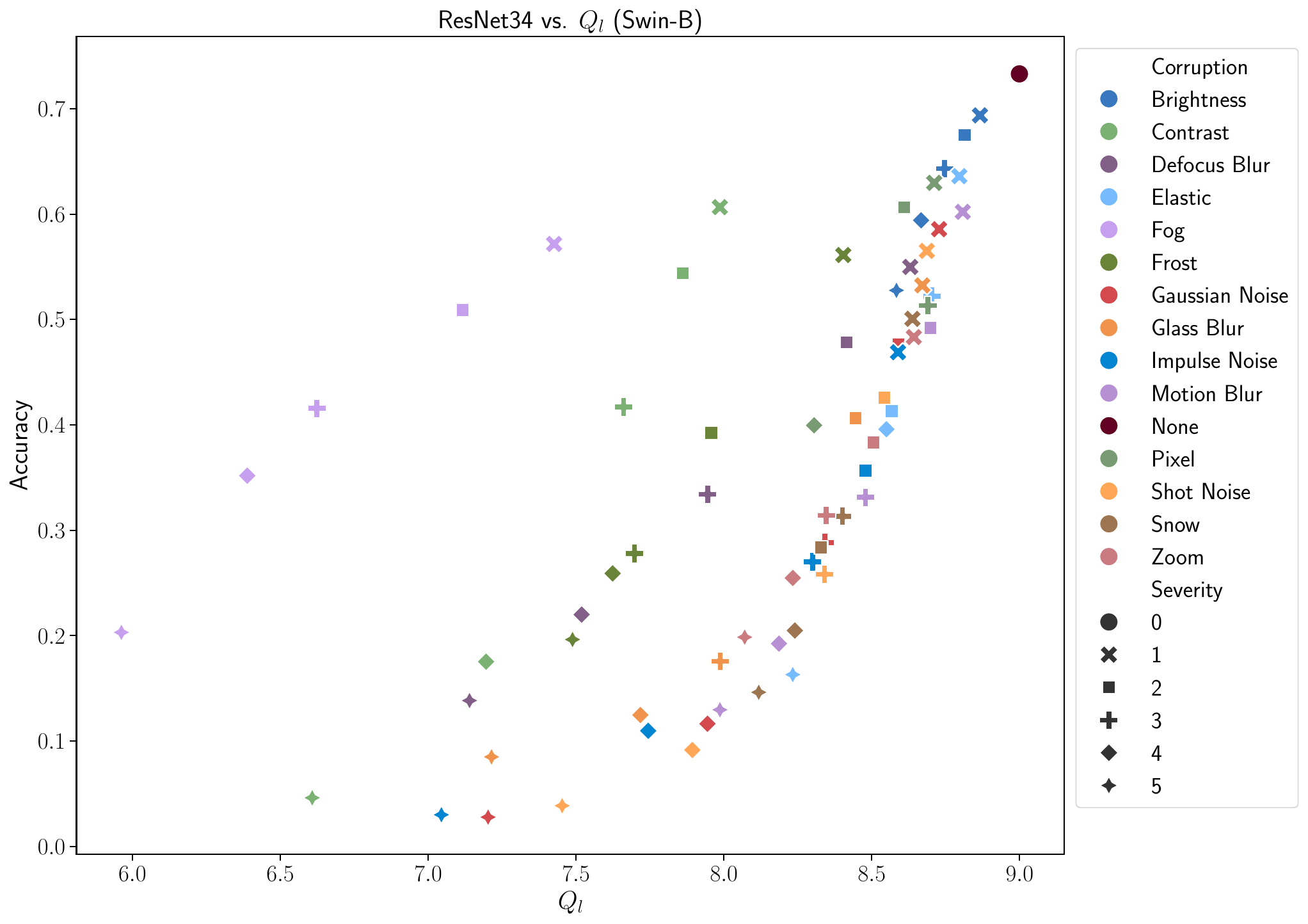} \\
    \includegraphics[width=0.48\linewidth]{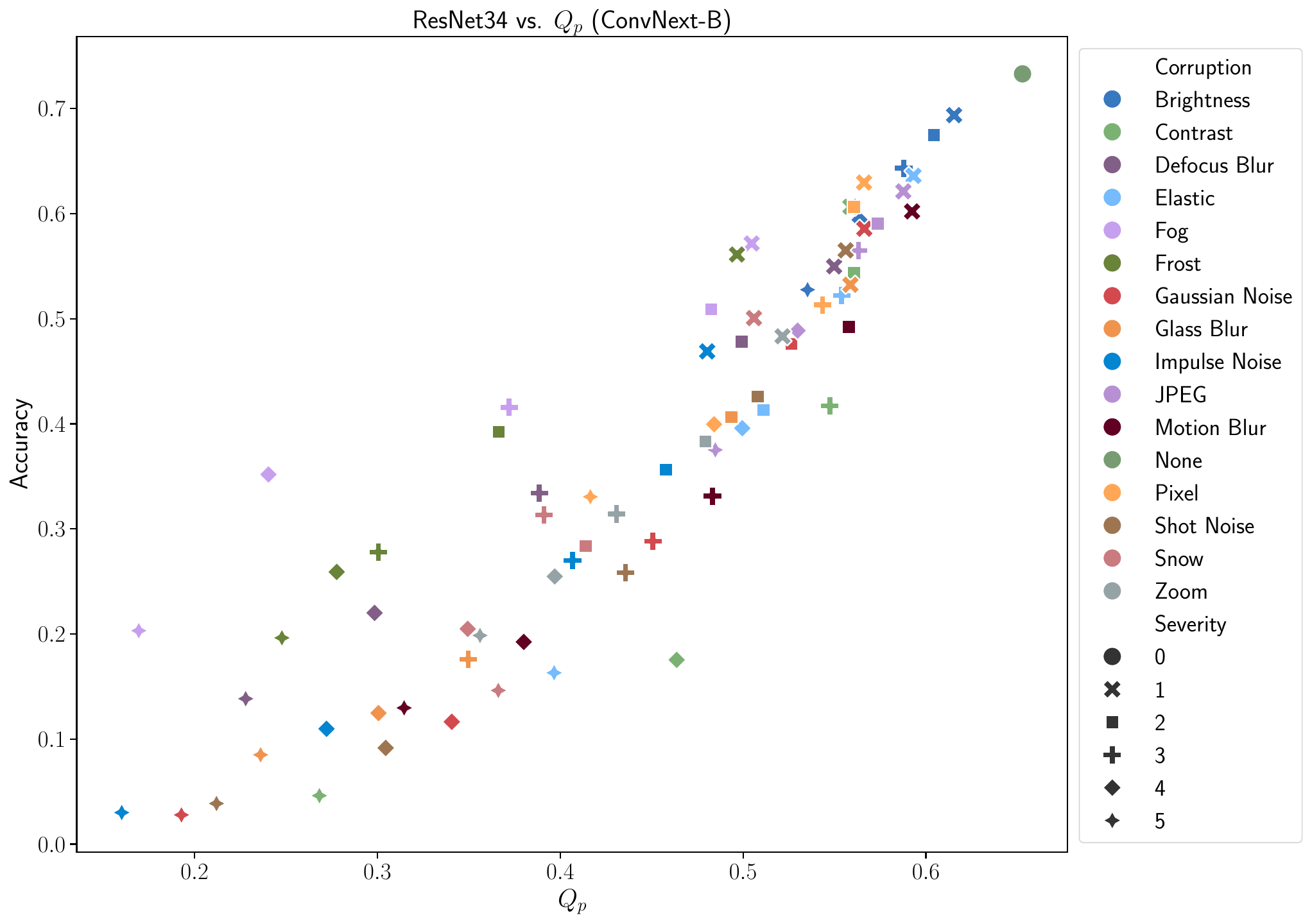}
    \includegraphics[width=0.48\linewidth]{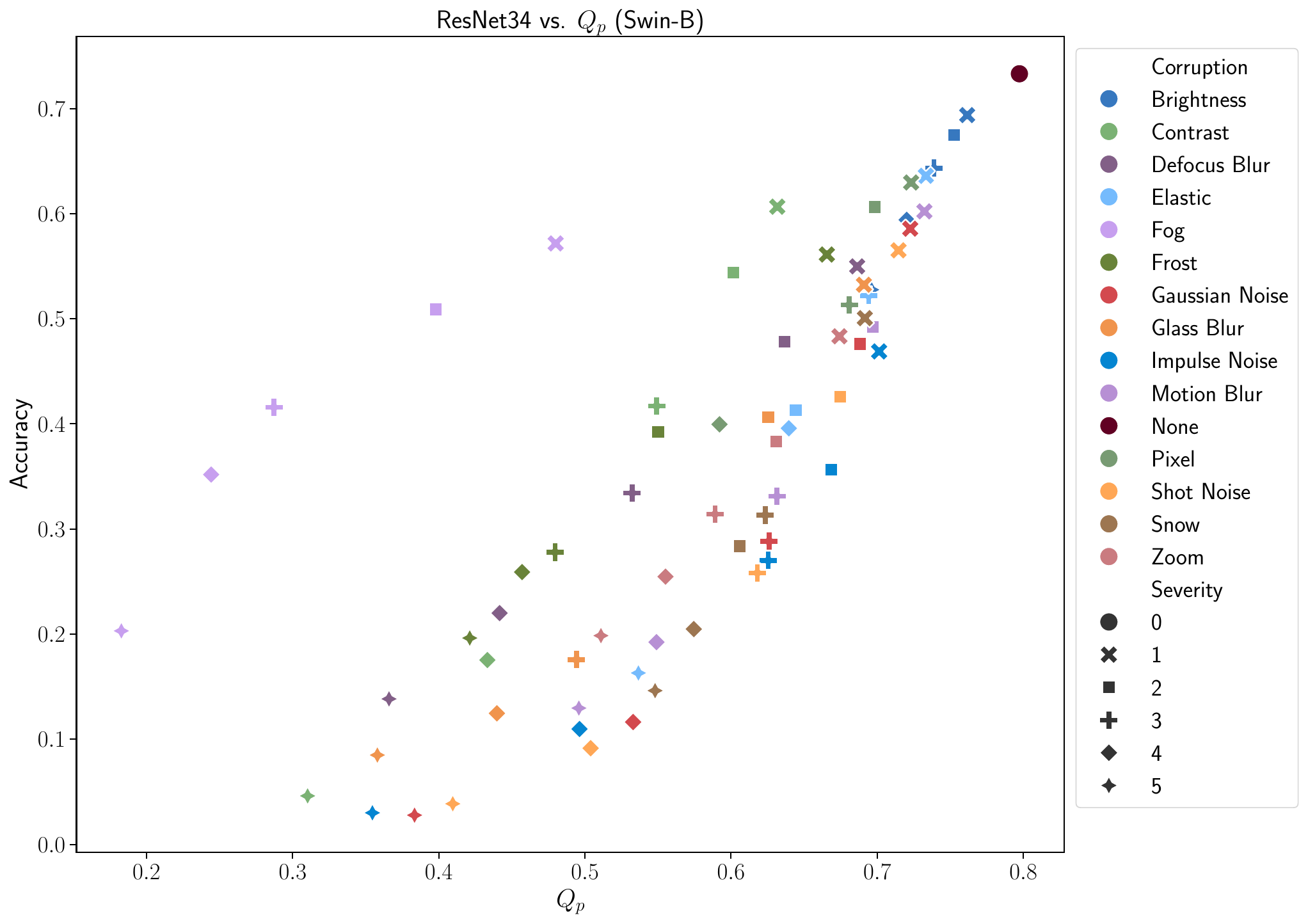} \\
    \caption{Comparison of ResNet34 accuracy with (row) $Q_h, Q_l, Q_p$ computed using (col) ConvNext-B, Swin-B. High correlation is observed between each IQA metric and accuracy. }
    \label{fig:app-stg-resnet}
\end{figure}

In Figures~\ref{fig:app-stg-convnext},~\ref{fig:app-stg-swin},~\ref{fig:app-stg-resnet}, we examine the relationship between DNN performance and the strong task-guided metrics ($Q_p, Q_h, Q_l$) described in \S\ref{sec:strong-tg-iqa}. Each figure pairs the task DNN under consideration with a pre-trained task model used to compute the quality metric.

Tables~\ref{tab:app-acc-iq-strong-tg-cnn},~\ref{tab:app-acc-iq-strong-tg-eff}, and~\ref{tab:app-acc-iq-strong-tg-tx} also show the point-wise predictability results for the strong task-guided IQA case.  These tables extend Table~\ref{tab:acc-iq-strong-tg} for additional task DNNs.  Results here show that strong task-guided IQA metrics are highly correlated with DNN performance and that predictability remains high regardless of whether the pre-trained DNN used to compute $Q$ is the same DNN used to obtain $M$.  

\begin{table}[h!]
    \centering
    \caption{\textbf{Strong Task-Guidance: } Correlation between IQ and accuracy for convolutional task DNN architectures. KRCC, SRCC, PLCC computed using average accuracy for each (corruption, severity). {\small AUC and CE based on point-wise predictions (95\% CI within $\pm0.001$). KRCC, SRCC, PLCC values have $p < 0.05$.}}
    \resizebox{0.75\linewidth}{!}{%
    \begin{tabular}{cl|ll||lll}
    \toprule
    Model & IQA Metric & AUC $\uparrow$ & CE $\downarrow$ & $\mid PLCC \mid$ $\uparrow$ & $\mid SRCC \mid$ $\uparrow$ \\
    \midrule
    \multirow[c]{18}{*}{ConvNext-B} & ConvNext-B $Q_h$ & 0.772 & 0.562 & 0.660$\pm$0.070 & 0.822$\pm$0.070 & 0.854$\pm$0.063 \\
 & ConvNext-B $Q_l$ & 0.778 & 0.555 & 0.660$\pm$0.067 & 0.826$\pm$0.067 & 0.854$\pm$0.063 \\
 & ConvNext-B $Q_p$ & 0.826 & 0.504 & 0.738$\pm$0.045 & 0.888$\pm$0.045 & 0.910$\pm$0.044 \\
 & EfficientNet-V2-M $Q_h$ & 0.805 & 0.529 & 0.855$\pm$0.030 & 0.945$\pm$0.030 & 0.969$\pm$0.019 \\
 & EfficientNet-V2-M $Q_l$ & 0.801 & 0.538 & 0.846$\pm$0.038 & 0.927$\pm$0.038 & 0.965$\pm$0.020 \\
 & EfficientNet-V2-M $Q_p$ & 0.831 & 0.500 & 0.867$\pm$0.025 & 0.956$\pm$0.025 & 0.973$\pm$0.018 \\
 & MobileNet-V3-L $Q_h$ & 0.746 & 0.581 & 0.847$\pm$0.028 & 0.927$\pm$0.028 & 0.966$\pm$0.022 \\
 & MobileNet-V3-L $Q_l$ & 0.743 & 0.586 & 0.840$\pm$0.039 & 0.909$\pm$0.039 & 0.962$\pm$0.026 \\
 & MobileNet-V3-L $Q_p$ & 0.751 & 0.575 & 0.858$\pm$0.025 & 0.929$\pm$0.025 & 0.968$\pm$0.020 \\
 & ResNet34 $Q_h$ & 0.725 & 0.601 & 0.770$\pm$0.069 & 0.859$\pm$0.069 & 0.924$\pm$0.045 \\
 & ResNet34 $Q_l$ & 0.717 & 0.603 & 0.776$\pm$0.051 & 0.866$\pm$0.051 & 0.925$\pm$0.043 \\
 & ResNet34 $Q_p$ & 0.719 & 0.601 & 0.775$\pm$0.051 & 0.875$\pm$0.051 & 0.926$\pm$0.044 \\
 & Swin-B $Q_h$ & 0.760 & 0.579 & 0.601$\pm$0.165 & 0.624$\pm$0.165 & 0.724$\pm$0.170 \\
 & Swin-B $Q_l$ & 0.724 & 0.601 & 0.554$\pm$0.166 & 0.604$\pm$0.166 & 0.686$\pm$0.176 \\
 & Swin-B $Q_p$ & 0.791 & 0.547 & 0.672$\pm$0.132 & 0.742$\pm$0.132 & 0.797$\pm$0.139 \\
 & ViT-L16 $Q_h$ & 0.781 & 0.562 & 0.615$\pm$0.117 & 0.716$\pm$0.117 & 0.808$\pm$0.100 \\
 & ViT-L16 $Q_l$ & 0.757 & 0.577 & 0.571$\pm$0.125 & 0.696$\pm$0.125 & 0.772$\pm$0.103 \\
 & ViT-L16 $Q_p$ & 0.800 & 0.540 & 0.671$\pm$0.093 & 0.786$\pm$0.093 & 0.855$\pm$0.075 \\
\cline{1-7}

\multirow[c]{18}{*}{ResNet34} & ConvNext-B $Q_h$ & 0.767 & 0.557 & 0.720$\pm$0.060 & 0.858$\pm$0.060 & 0.889$\pm$0.055 \\
 & ConvNext-B $Q_l$ & 0.760 & 0.563 & 0.733$\pm$0.059 & 0.853$\pm$0.059 & 0.896$\pm$0.055 \\
 & ConvNext-B $Q_p$ & 0.801 & 0.522 & 0.785$\pm$0.044 & 0.904$\pm$0.044 & 0.930$\pm$0.041 \\
 & EfficientNet-V2-M $Q_h$ & 0.796 & 0.527 & 0.786$\pm$0.035 & 0.903$\pm$0.035 & 0.936$\pm$0.033 \\
 & EfficientNet-V2-M $Q_l$ & 0.784 & 0.539 & 0.785$\pm$0.041 & 0.877$\pm$0.041 & 0.934$\pm$0.035 \\
 & EfficientNet-V2-M $Q_p$ & 0.814 & 0.508 & 0.789$\pm$0.031 & 0.913$\pm$0.031 & 0.939$\pm$0.032 \\
 & MobileNet-V3-L $Q_h$ & 0.814 & 0.506 & 0.864$\pm$0.015 & 0.958$\pm$0.015 & 0.968$\pm$0.022 \\
 & MobileNet-V3-L $Q_l$ & 0.807 & 0.515 & 0.858$\pm$0.019 & 0.937$\pm$0.019 & 0.966$\pm$0.022 \\
 & MobileNet-V3-L $Q_p$ & 0.823 & 0.495 & 0.872$\pm$0.016 & 0.970$\pm$0.016 & 0.971$\pm$0.021 \\
 & ResNet34 $Q_h$ & 0.848 & 0.470 & 0.862$\pm$0.028 & 0.930$\pm$0.028 & 0.969$\pm$0.023 \\
 & ResNet34 $Q_l$ & 0.827 & 0.492 & 0.870$\pm$0.015 & 0.951$\pm$0.015 & 0.973$\pm$0.020 \\
 & ResNet34 $Q_p$ & 0.850 & 0.461 & 0.886$\pm$0.015 & 0.960$\pm$0.015 & 0.977$\pm$0.021 \\
 & Swin-B $Q_h$ & 0.751 & 0.574 & 0.636$\pm$0.153 & 0.643$\pm$0.153 & 0.774$\pm$0.140 \\
 & Swin-B $Q_l$ & 0.709 & 0.600 & 0.624$\pm$0.146 & 0.628$\pm$0.146 & 0.754$\pm$0.145 \\
 & Swin-B $Q_p$ & 0.774 & 0.551 & 0.673$\pm$0.129 & 0.747$\pm$0.129 & 0.825$\pm$0.112 \\
 & ViT-L16 $Q_h$ & 0.782 & 0.546 & 0.687$\pm$0.102 & 0.771$\pm$0.102 & 0.854$\pm$0.083 \\
 & ViT-L16 $Q_l$ & 0.751 & 0.567 & 0.655$\pm$0.100 & 0.757$\pm$0.100 & 0.834$\pm$0.089 \\
 & ViT-L16 $Q_p$ & 0.794 & 0.534 & 0.721$\pm$0.081 & 0.825$\pm$0.081 & 0.883$\pm$0.070 \\
\cline{1-7}
    \bottomrule
    \end{tabular}
    }%
    \label{tab:app-acc-iq-strong-tg-cnn}
\end{table}

\begin{table}[h!]
    \centering
    \caption{\textbf{Strong Task-Guidance: } Correlation between IQ and accuracy for efficient task DNN architectures. KRCC, SRCC, PLCC computed using average accuracy for each (corruption, severity). {\small AUC and CE based on point-wise predictions (95\% CI within $\pm0.001$). KRCC, SRCC, PLCC values have $p < 0.05$.}}
    \resizebox{0.75\linewidth}{!}{%
    \begin{tabular}{cl|ll||lll}
    \toprule
    Model & IQA Metric & AUC $\uparrow$ & CE $\downarrow$ & $\mid PLCC \mid$ $\uparrow$ & $\mid SRCC \mid$ $\uparrow$ \\
    \midrule
    \multirow[c]{18}{*}{EfficientNet-V2-M} & ConvNext-B $Q_h$ & 0.753 & 0.578 & 0.604$\pm$0.102 & 0.742$\pm$0.102 & 0.802$\pm$0.086 \\
 & ConvNext-B $Q_l$ & 0.753 & 0.578 & 0.601$\pm$0.101 & 0.740$\pm$0.101 & 0.803$\pm$0.085 \\
 & ConvNext-B $Q_p$ & 0.798 & 0.534 & 0.679$\pm$0.075 & 0.820$\pm$0.075 & 0.867$\pm$0.062 \\
 & EfficientNet-V2-M $Q_h$ & 0.831 & 0.497 & 0.888$\pm$0.020 & 0.956$\pm$0.020 & 0.981$\pm$0.012 \\
 & EfficientNet-V2-M $Q_l$ & 0.831 & 0.496 & 0.876$\pm$0.029 & 0.937$\pm$0.029 & 0.977$\pm$0.014 \\
 & EfficientNet-V2-M $Q_p$ & 0.862 & 0.456 & 0.900$\pm$0.014 & 0.968$\pm$0.014 & 0.984$\pm$0.010 \\
 & MobileNet-V3-L $Q_h$ & 0.749 & 0.578 & 0.827$\pm$0.035 & 0.905$\pm$0.035 & 0.958$\pm$0.024 \\
 & MobileNet-V3-L $Q_l$ & 0.745 & 0.583 & 0.819$\pm$0.045 & 0.885$\pm$0.045 & 0.953$\pm$0.027 \\
 & MobileNet-V3-L $Q_p$ & 0.752 & 0.573 & 0.827$\pm$0.034 & 0.908$\pm$0.034 & 0.958$\pm$0.024 \\
 & ResNet34 $Q_h$ & 0.731 & 0.595 & 0.766$\pm$0.066 & 0.865$\pm$0.066 & 0.921$\pm$0.043 \\
 & ResNet34 $Q_l$ & 0.724 & 0.597 & 0.762$\pm$0.051 & 0.864$\pm$0.051 & 0.920$\pm$0.042 \\
 & ResNet34 $Q_p$ & 0.723 & 0.597 & 0.761$\pm$0.052 & 0.873$\pm$0.052 & 0.921$\pm$0.043 \\
 & Swin-B $Q_h$ & 0.748 & 0.589 & 0.549$\pm$0.189 & 0.533$\pm$0.189 & 0.673$\pm$0.187 \\
 & Swin-B $Q_l$ & 0.711 & 0.609 & 0.503$\pm$0.184 & 0.508$\pm$0.184 & 0.632$\pm$0.197 \\
 & Swin-B $Q_p$ & 0.778 & 0.558 & 0.621$\pm$0.162 & 0.663$\pm$0.162 & 0.750$\pm$0.157 \\
 & ViT-L16 $Q_h$ & 0.760 & 0.579 & 0.565$\pm$0.151 & 0.607$\pm$0.151 & 0.749$\pm$0.122 \\
 & ViT-L16 $Q_l$ & 0.735 & 0.593 & 0.521$\pm$0.155 & 0.582$\pm$0.155 & 0.709$\pm$0.132 \\
 & ViT-L16 $Q_p$ & 0.779 & 0.560 & 0.620$\pm$0.134 & 0.684$\pm$0.134 & 0.801$\pm$0.099 \\
\cline{1-7}
\multirow[c]{18}{*}{MobileNet-V3-L} & ConvNext-B $Q_h$ & 0.778 & 0.561 & 0.678$\pm$0.074 & 0.842$\pm$0.074 & 0.861$\pm$0.069 \\
 & ConvNext-B $Q_l$ & 0.773 & 0.567 & 0.679$\pm$0.072 & 0.840$\pm$0.072 & 0.862$\pm$0.072 \\
 & ConvNext-B $Q_p$ & 0.815 & 0.520 & 0.749$\pm$0.051 & 0.901$\pm$0.051 & 0.910$\pm$0.050 \\
 & EfficientNet-V2-M $Q_h$ & 0.815 & 0.521 & 0.836$\pm$0.024 & 0.938$\pm$0.024 & 0.966$\pm$0.018 \\
 & EfficientNet-V2-M $Q_l$ & 0.804 & 0.534 & 0.833$\pm$0.032 & 0.913$\pm$0.032 & 0.964$\pm$0.019 \\
 & EfficientNet-V2-M $Q_p$ & 0.834 & 0.497 & 0.847$\pm$0.018 & 0.951$\pm$0.018 & 0.969$\pm$0.016 \\
 & MobileNet-V3-L $Q_h$ & 0.833 & 0.496 & 0.906$\pm$0.019 & 0.973$\pm$0.019 & 0.984$\pm$0.013 \\
 & MobileNet-V3-L $Q_l$ & 0.835 & 0.496 & 0.895$\pm$0.031 & 0.951$\pm$0.031 & 0.982$\pm$0.015 \\
 & MobileNet-V3-L $Q_p$ & 0.853 & 0.470 & 0.922$\pm$0.010 & 0.983$\pm$0.010 & 0.988$\pm$0.010 \\
 & ResNet34 $Q_h$ & 0.783 & 0.558 & 0.778$\pm$0.061 & 0.872$\pm$0.061 & 0.925$\pm$0.044 \\
 & ResNet34 $Q_l$ & 0.769 & 0.563 & 0.783$\pm$0.046 & 0.891$\pm$0.046 & 0.926$\pm$0.045 \\
 & ResNet34 $Q_p$ & 0.778 & 0.555 & 0.787$\pm$0.044 & 0.904$\pm$0.044 & 0.930$\pm$0.043 \\
 & Swin-B $Q_h$ & 0.756 & 0.587 & 0.613$\pm$0.180 & 0.587$\pm$0.180 & 0.716$\pm$0.181 \\
 & Swin-B $Q_l$ & 0.714 & 0.614 & 0.583$\pm$0.175 & 0.567$\pm$0.175 & 0.685$\pm$0.183 \\
 & Swin-B $Q_p$ & 0.785 & 0.556 & 0.670$\pm$0.149 & 0.710$\pm$0.149 & 0.783$\pm$0.144 \\
 & ViT-L16 $Q_h$ & 0.788 & 0.557 & 0.655$\pm$0.122 & 0.745$\pm$0.122 & 0.819$\pm$0.104 \\
 & ViT-L16 $Q_l$ & 0.757 & 0.578 & 0.619$\pm$0.121 & 0.731$\pm$0.121 & 0.793$\pm$0.111 \\
 & ViT-L16 $Q_p$ & 0.804 & 0.539 & 0.706$\pm$0.103 & 0.807$\pm$0.103 & 0.858$\pm$0.085 \\
\cline{1-7}\bottomrule
    \end{tabular}
    }%
    \label{tab:app-acc-iq-strong-tg-eff}
\end{table}

\begin{table}[h!]
    \centering
    \caption{\textbf{Strong Task-Guidance: } Correlation between IQ and accuracy for transformer task DNN architectures. KRCC, SRCC, PLCC computed using average accuracy for each (corruption, severity). {\small AUC and CE based on point-wise predictions (95\% CI within $\pm0.001$). KRCC, SRCC, PLCC values have $p < 0.05$.}}
    \resizebox{0.75\linewidth}{!}{%
    \begin{tabular}{cl|ll||lll}
    \toprule
    Model & IQA Metric & AUC $\uparrow$ & CE $\downarrow$ & $\mid PLCC \mid$ $\uparrow$ & $\mid SRCC \mid$ $\uparrow$ \\
    \midrule
         \multirow[c]{18}{*}{Swin-B} & ConvNext-B $Q_h$ & 0.744 & 0.586 & 0.574$\pm$0.129 & 0.706$\pm$0.129 & 0.768$\pm$0.098 \\
 & ConvNext-B $Q_l$ & 0.746 & 0.586 & 0.573$\pm$0.127 & 0.709$\pm$0.127 & 0.768$\pm$0.099 \\
 & ConvNext-B $Q_p$ & 0.791 & 0.542 & 0.649$\pm$0.102 & 0.788$\pm$0.102 & 0.834$\pm$0.079 \\
 & EfficientNet-V2-M $Q_h$ & 0.797 & 0.537 & 0.835$\pm$0.026 & 0.943$\pm$0.026 & 0.964$\pm$0.019 \\
 & EfficientNet-V2-M $Q_l$ & 0.793 & 0.547 & 0.826$\pm$0.037 & 0.923$\pm$0.037 & 0.960$\pm$0.020 \\
 & EfficientNet-V2-M $Q_p$ & 0.822 & 0.510 & 0.853$\pm$0.018 & 0.956$\pm$0.018 & 0.970$\pm$0.016 \\
 & MobileNet-V3-L $Q_h$ & 0.742 & 0.583 & 0.812$\pm$0.039 & 0.911$\pm$0.039 & 0.950$\pm$0.028 \\
 & MobileNet-V3-L $Q_l$ & 0.739 & 0.589 & 0.804$\pm$0.049 & 0.893$\pm$0.049 & 0.945$\pm$0.031 \\
 & MobileNet-V3-L $Q_p$ & 0.748 & 0.576 & 0.814$\pm$0.033 & 0.913$\pm$0.033 & 0.952$\pm$0.025 \\
 & ResNet34 $Q_h$ & 0.722 & 0.603 & 0.724$\pm$0.078 & 0.828$\pm$0.078 & 0.896$\pm$0.053 \\
 & ResNet34 $Q_l$ & 0.713 & 0.604 & 0.721$\pm$0.061 & 0.831$\pm$0.061 & 0.892$\pm$0.053 \\
 & ResNet34 $Q_p$ & 0.716 & 0.602 & 0.727$\pm$0.062 & 0.845$\pm$0.062 & 0.897$\pm$0.052 \\
 & Swin-B $Q_h$ & 0.766 & 0.578 & 0.532$\pm$0.207 & 0.483$\pm$0.207 & 0.654$\pm$0.174 \\
 & Swin-B $Q_l$ & 0.732 & 0.597 & 0.485$\pm$0.203 & 0.458$\pm$0.203 & 0.611$\pm$0.181 \\
 & Swin-B $Q_p$ & 0.807 & 0.529 & 0.603$\pm$0.184 & 0.620$\pm$0.184 & 0.732$\pm$0.142 \\
 & ViT-L16 $Q_h$ & 0.768 & 0.576 & 0.542$\pm$0.177 & 0.584$\pm$0.177 & 0.730$\pm$0.123 \\
 & ViT-L16 $Q_l$ & 0.744 & 0.589 & 0.502$\pm$0.176 & 0.565$\pm$0.176 & 0.692$\pm$0.130 \\
 & ViT-L16 $Q_p$ & 0.789 & 0.551 & 0.596$\pm$0.158 & 0.663$\pm$0.158 & 0.782$\pm$0.105 \\
\cline{1-7}
\multirow[c]{18}{*}{ViT-L16} & ConvNext-B $Q_h$ & 0.736 & 0.590 & 0.635$\pm$0.086 & 0.780$\pm$0.086 & 0.834$\pm$0.064 \\
 & ConvNext-B $Q_l$ & 0.739 & 0.591 & 0.642$\pm$0.085 & 0.784$\pm$0.085 & 0.837$\pm$0.063 \\
 & ConvNext-B $Q_p$ & 0.781 & 0.551 & 0.709$\pm$0.065 & 0.850$\pm$0.065 & 0.887$\pm$0.053 \\
 & EfficientNet-V2-M $Q_h$ & 0.781 & 0.553 & 0.804$\pm$0.028 & 0.939$\pm$0.028 & 0.947$\pm$0.028 \\
 & EfficientNet-V2-M $Q_l$ & 0.776 & 0.565 & 0.800$\pm$0.037 & 0.920$\pm$0.037 & 0.944$\pm$0.028 \\
 & EfficientNet-V2-M $Q_p$ & 0.803 & 0.530 & 0.815$\pm$0.022 & 0.949$\pm$0.022 & 0.953$\pm$0.027 \\
 & MobileNet-V3-L $Q_h$ & 0.732 & 0.589 & 0.807$\pm$0.033 & 0.926$\pm$0.033 & 0.950$\pm$0.024 \\
 & MobileNet-V3-L $Q_l$ & 0.729 & 0.595 & 0.800$\pm$0.046 & 0.908$\pm$0.046 & 0.946$\pm$0.027 \\
 & MobileNet-V3-L $Q_p$ & 0.740 & 0.580 & 0.813$\pm$0.025 & 0.931$\pm$0.025 & 0.954$\pm$0.022 \\
 & ResNet34 $Q_h$ & 0.718 & 0.604 & 0.772$\pm$0.074 & 0.861$\pm$0.074 & 0.923$\pm$0.045 \\
 & ResNet34 $Q_l$ & 0.707 & 0.607 & 0.771$\pm$0.057 & 0.868$\pm$0.057 & 0.922$\pm$0.047 \\
 & ResNet34 $Q_p$ & 0.713 & 0.602 & 0.777$\pm$0.055 & 0.882$\pm$0.055 & 0.926$\pm$0.045 \\
 & Swin-B $Q_h$ & 0.742 & 0.594 & 0.572$\pm$0.164 & 0.558$\pm$0.164 & 0.715$\pm$0.139 \\
 & Swin-B $Q_l$ & 0.706 & 0.613 & 0.533$\pm$0.163 & 0.535$\pm$0.163 & 0.677$\pm$0.154 \\
 & Swin-B $Q_p$ & 0.774 & 0.562 & 0.630$\pm$0.138 & 0.685$\pm$0.138 & 0.787$\pm$0.110 \\
 & ViT-L16 $Q_h$ & 0.799 & 0.545 & 0.633$\pm$0.124 & 0.690$\pm$0.124 & 0.819$\pm$0.092 \\
 & ViT-L16 $Q_l$ & 0.780 & 0.556 & 0.592$\pm$0.124 & 0.672$\pm$0.124 & 0.786$\pm$0.092 \\
 & ViT-L16 $Q_p$ & 0.828 & 0.506 & 0.694$\pm$0.104 & 0.765$\pm$0.104 & 0.864$\pm$0.075 \\
\cline{1-7}
\bottomrule
    \end{tabular}
    }%
    \label{tab:app-acc-iq-strong-tg-tx}
\end{table}

\section{Relationship of weak task-guided IQA and DNN performance metrics}
\begin{figure}
    \centering
    \includegraphics[width=0.75\linewidth]{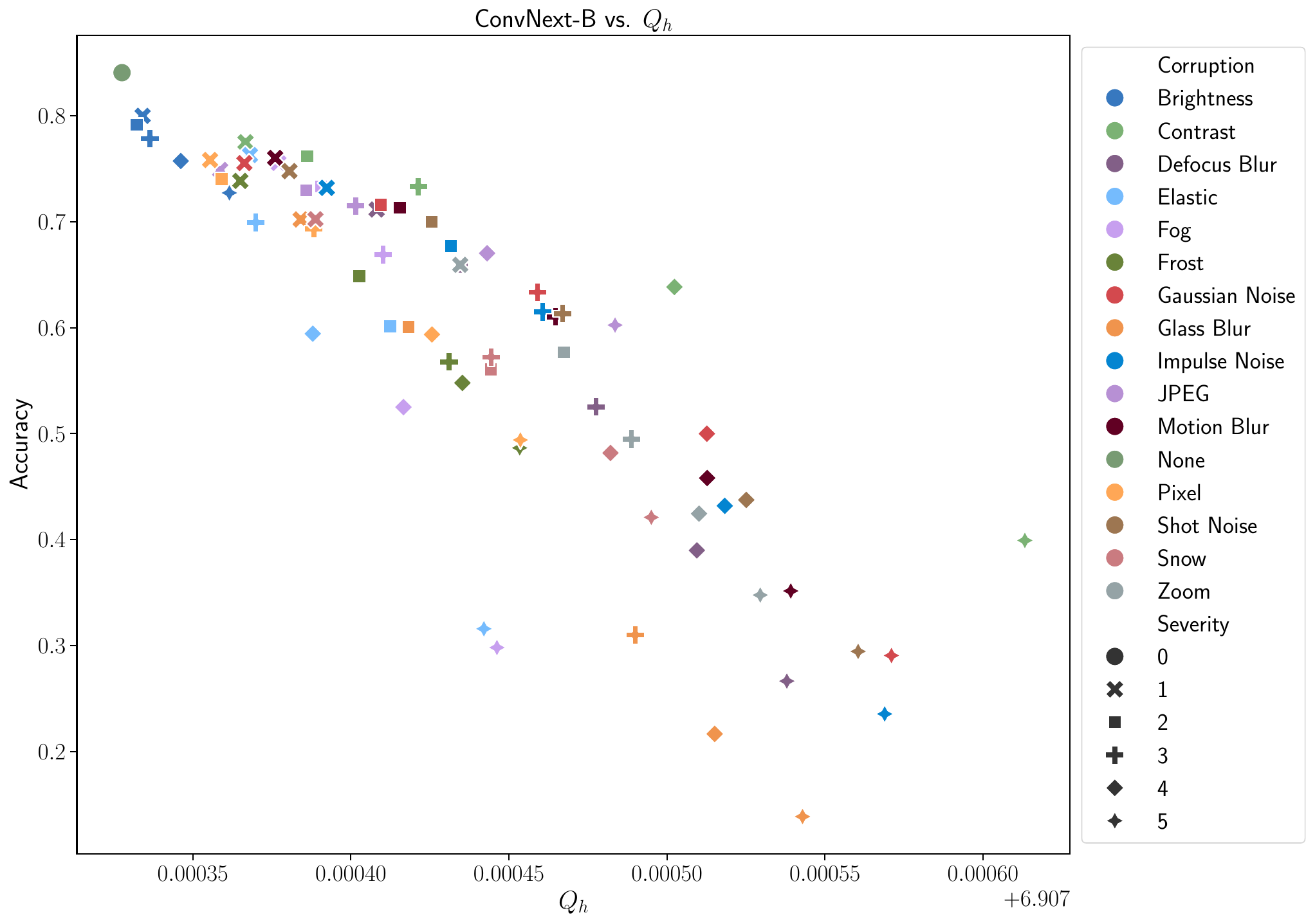} \\
    \includegraphics[width=0.75\linewidth]{figures/Figure-6-acc-convnext-zsclip-logit.pdf} \\
    \includegraphics[width=0.75\linewidth]{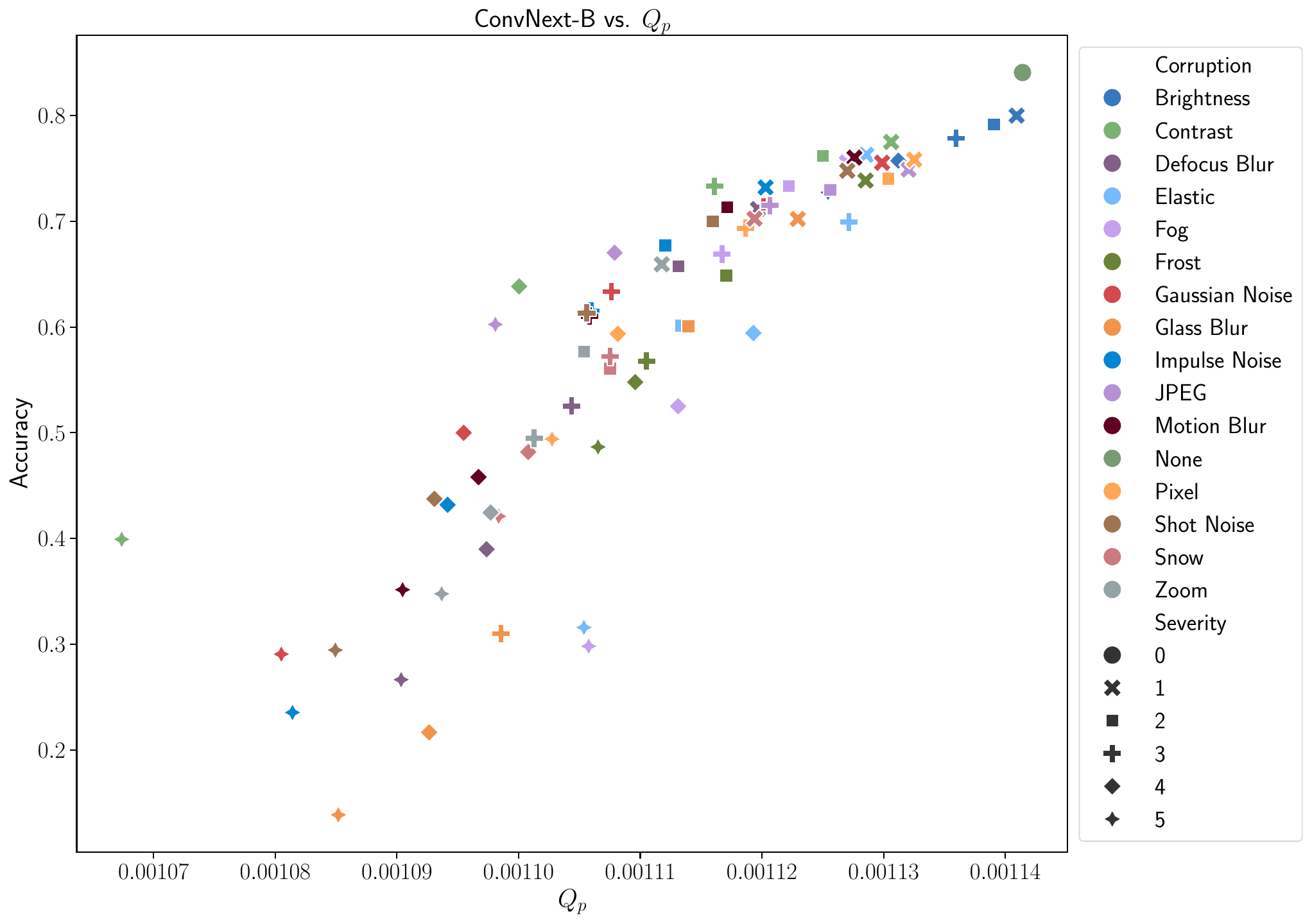}
    \caption{Comparison of ConvNext-B accuracy with (top to bottom) $Q_h, Q_l, Q_p$ based on ZSCLIP-IQA. High correlation is observed between each ZSCLIP-IQA variant and accuracy.}
    \label{fig:app-wtg-convnext}
\end{figure}

\begin{figure}
    \centering
    \includegraphics[width=0.75\linewidth]{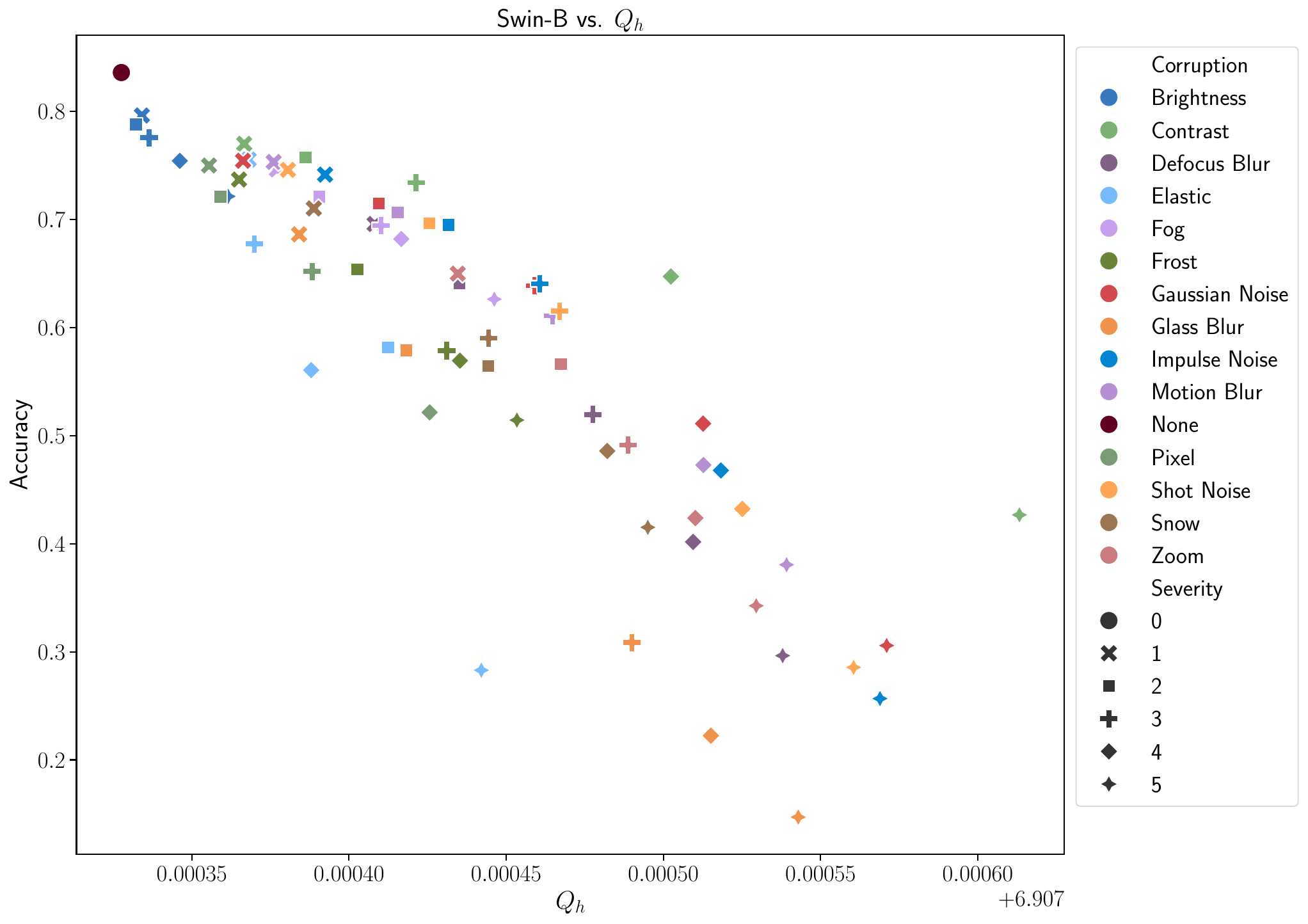} \\
    \includegraphics[width=0.75\linewidth]{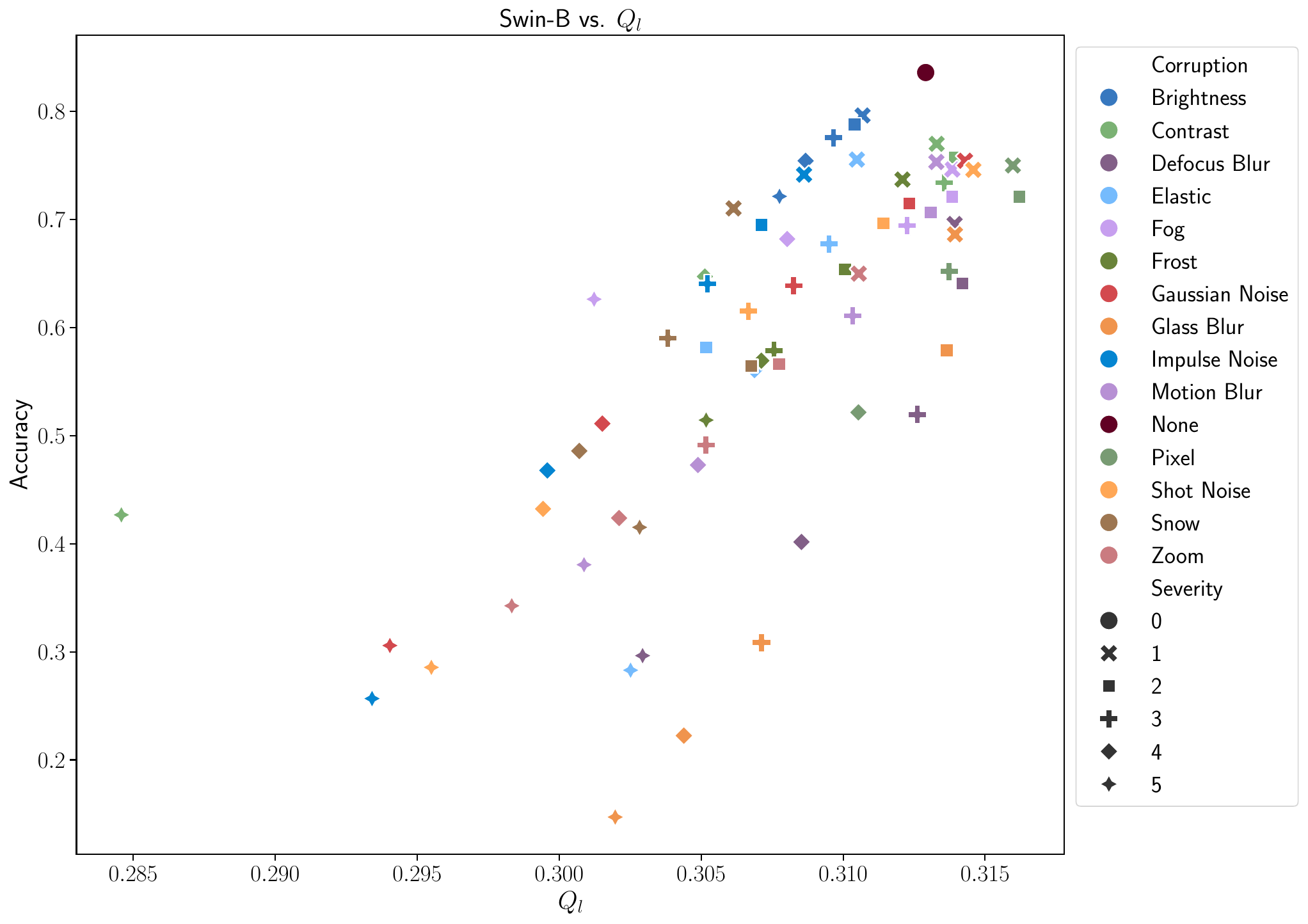} \\
    \includegraphics[width=0.75\linewidth]{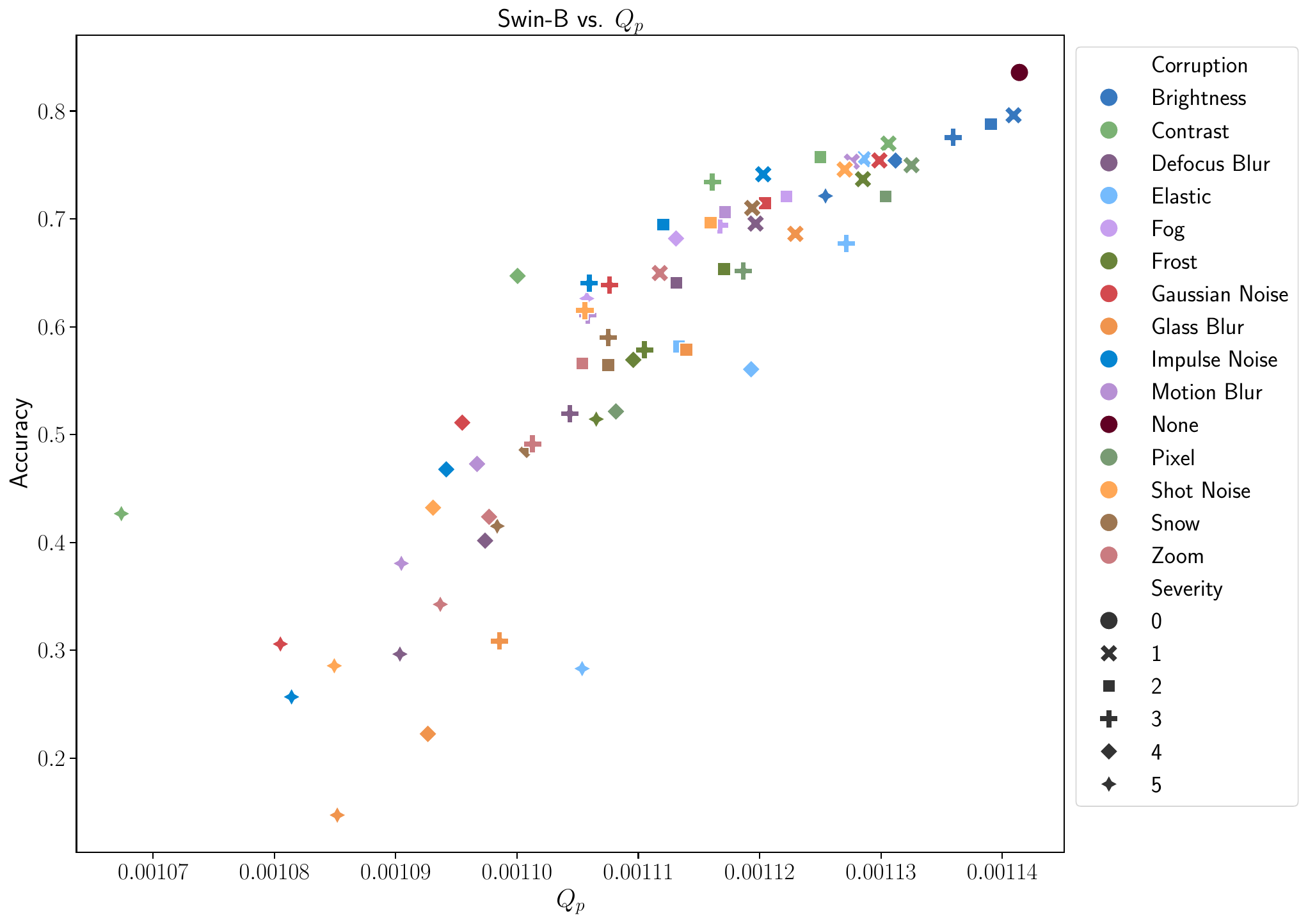}
    \caption{Comparison of Swin-B accuracy with (top to bottom) $Q_h, Q_l, Q_p$ based on ZSCLIP-IQA. Some correlation is observed between each ZSCLIP-IQA variant and accuracy.}
    \label{fig:app-wtg-swin}
\end{figure}

\begin{figure}
    \centering
    \includegraphics[width=0.75\linewidth]{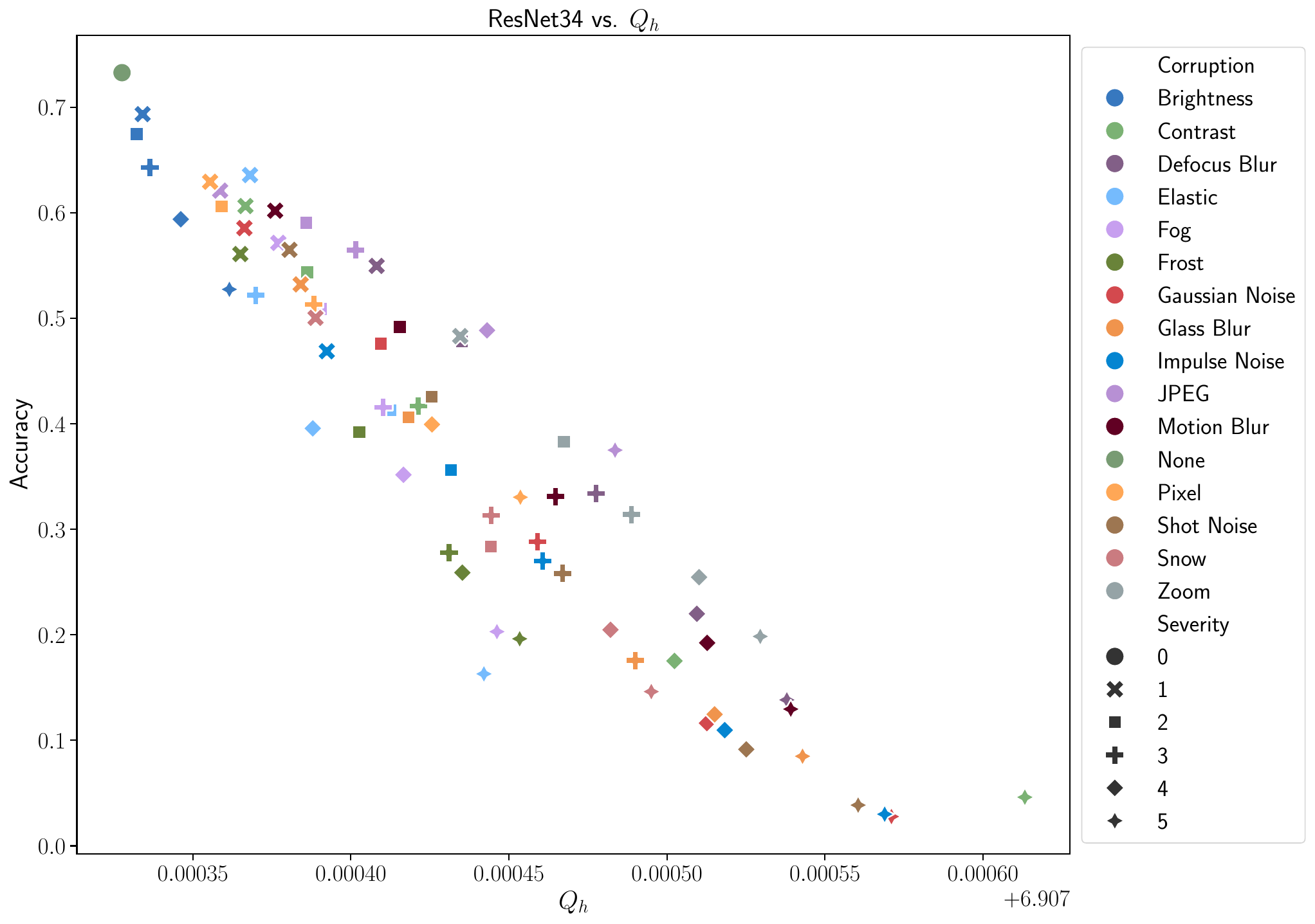} \\
    \includegraphics[width=0.75\linewidth]{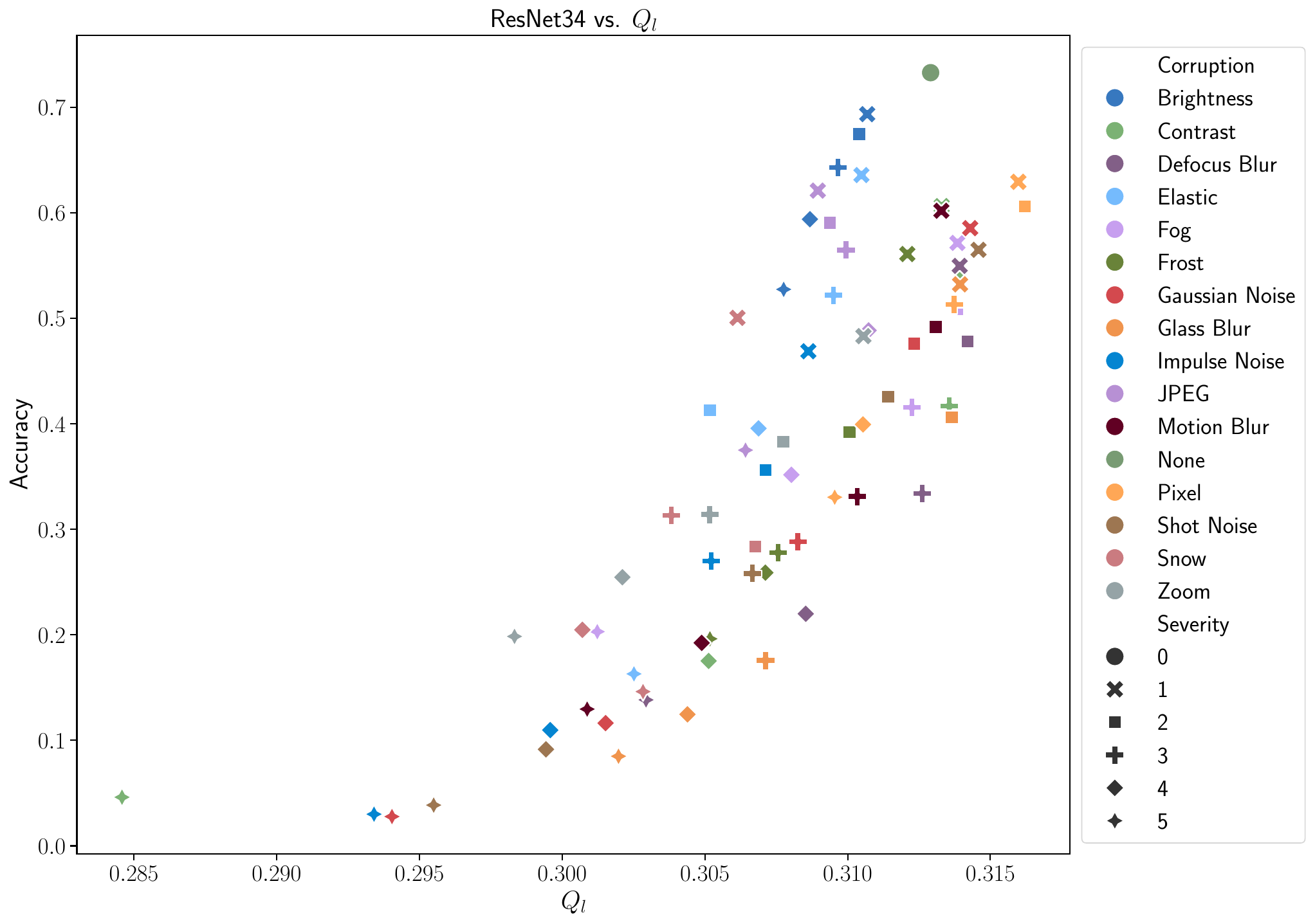} \\
    \includegraphics[width=0.75\linewidth]{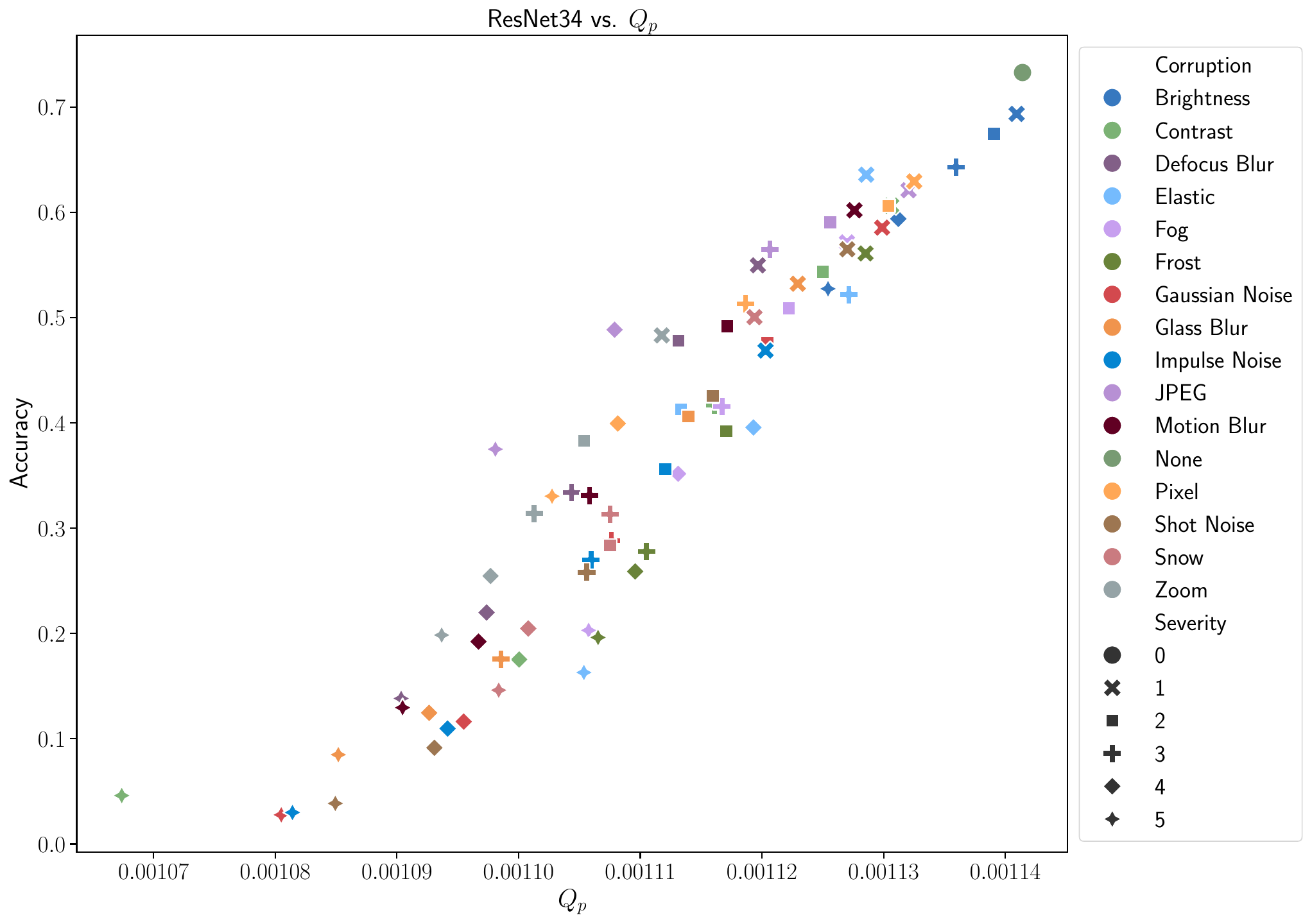}
    \caption{Comparison of ResNet34 accuracy with (top to bottom) $Q_h, Q_l, Q_p$ based on ZSCLIP-IQA. High correlation is observed between each ZSCLIP-IQA variant and accuracy.}
    \label{fig:app-wtg-resnet}
\end{figure}

We provide Figures~\ref{fig:app-wtg-convnext},~\ref{fig:app-wtg-swin}, ~\ref{fig:app-wtg-resnet} showing the relationship between DNN performance the weak task-guided ZSCLIP-IQA metric from \S\ref{sec:weak-tg-iqa}.

\section{Controlling for image content when evaluating predictability}
\label{app:control-for-content}
The experiments of \S\ref{sec:exp1}-\ref{sec:weak-tg-iqa} examined predictability by modeling $P(M|Q,X)$.  Since the content of $X$ may be a confounder for both $M,Q$, we attempt to control for it in two ways.  

\begin{figure}[h!]
    \centering    
    \begin{subfigure}[b]{0.48\textwidth}
    \includegraphics[width=\textwidth]{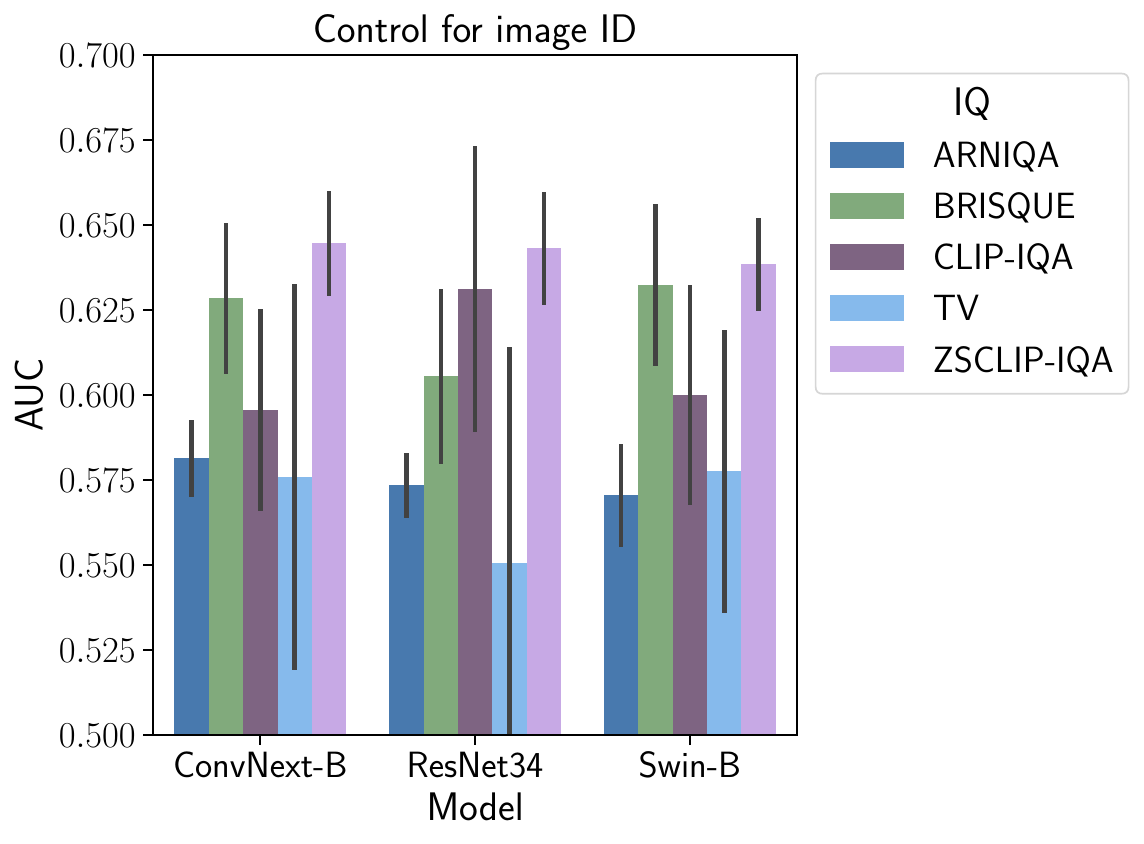}
    \caption{}
    \label{fig:app-control-id}
    \end{subfigure}
    \hfill
    \begin{subfigure}[b]{0.48\textwidth}
    \includegraphics[width=\textwidth]{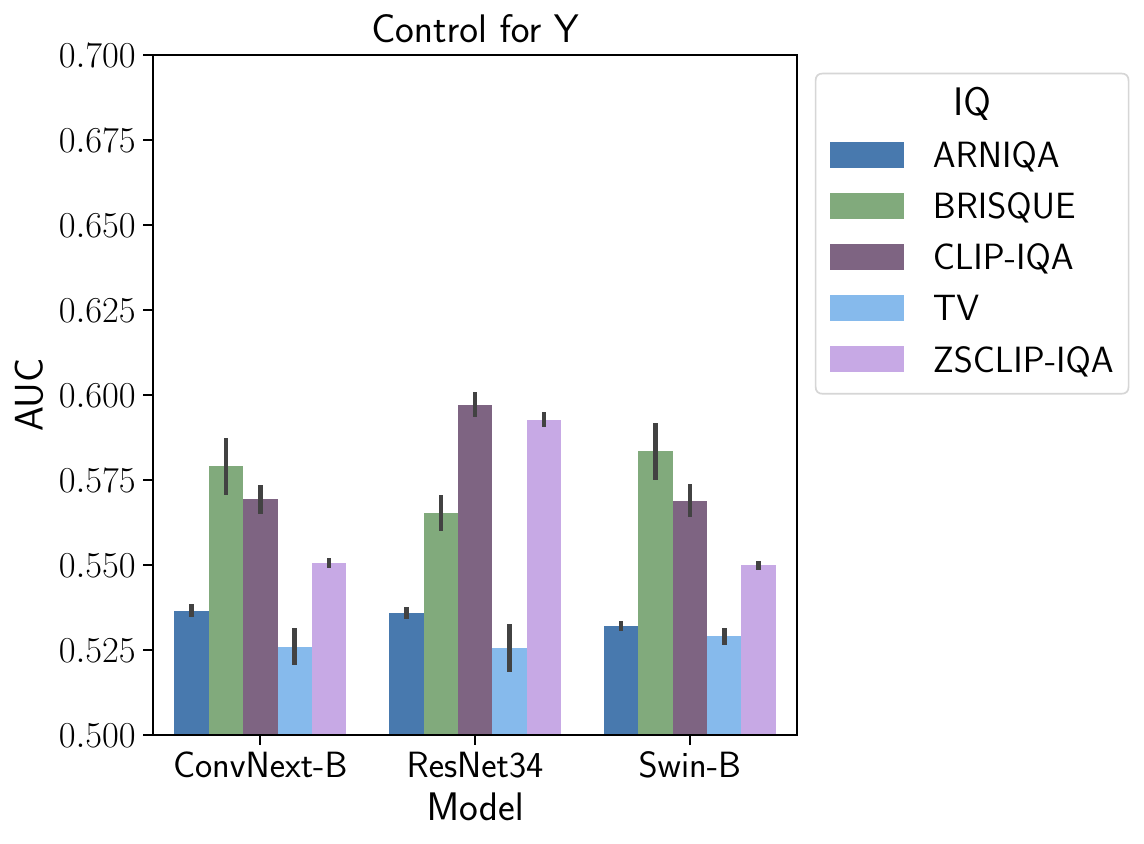}
    \caption{}
    \label{fig:app-control-y}
    \end{subfigure}
    \caption{Mean AUC (mAUC) for classifiers trained (a) per image ID ($X_i$) to model $P(M|Q,X_i)$ and (b) per label $Y$ to model $P(M|Q,Y)$. Averages are taken over all images/labels respectively and cross-validation folds with error bars indicating one standard deviation. Higher variance in (a) is attributed to lower sample sizes for training the logistic regression classifier.}
    \label{fig:app-control}
\end{figure}

In the first case, we take advantage of the synthetic nature of the IN-C dataset by looking at the predictability of $M$ from $Q$ for each image in the dataset separately which allows us to control the content precisely and only change the quality characteristics.  For this experiment, we train a logistic regression classifier to predict $P(M|Q)$ for individual image IDs trained using only $M, Q$ computed from the set of distorted variants of each specific image ID. Given the original clean image and 15 corruptions with 5 severity levels each, we run 5-fold cross-validation with an 80/20 train/test split of the 76 total images.  We repeat this for all 50k image IDs in the ImageNet validation set.  The results shown in Figure~\ref{fig:app-control-id} are an average of the AUC over all image IDs and folds.

We see that even when controlling for the image content the weak task-guided IQA generally achieves the highest $mAUC$ with the lowest variance.   Overall, this supports our hypothesis and causal analysis that weak task-guidance provides a means to associate $M, Q$ even when conditioning on the image directly.  

In the second case, we adjust for image content by controlling for the image label $Y$. Here, we train a separate classifier to model $P(M|Q)$ for each of the 1000 labels in the ImageNet dataset.  Each classifier is trained on the aggregate of 50 images per label along with all 15 corruptions at 5 severity levels (3751 total per label).  We again use an 80/20 train/test split and perform 5-fold cross-validation.  The results shown in Figure~\ref{fig:app-control-y} are an average of the $AUC$ over all labels and folds.

We find here that across all IQ metrics evaluated, $mAUC$ is barely above chance.  For the traditional NR-IQA metrics, this supports our analysis and main experiments which show little correlation between $M,Q$.  For the ZSCLIP-IQA metric, we refer again to the causal model (Fig.~\ref{fig:zsclip-dag}) and see that while the task selection variable ensures the association between $M, Q$, it is the conditioning on $Y$ (and $X$), as done here, that blocks all paths between $M, Q$ and once again removes the association.

\section{Predictability of DNN performance for mildly corrupted datasets}
\label{app:pred-real-world-data}
\begin{figure}[h!]
    \centering
    \includegraphics[width=0.47\linewidth]{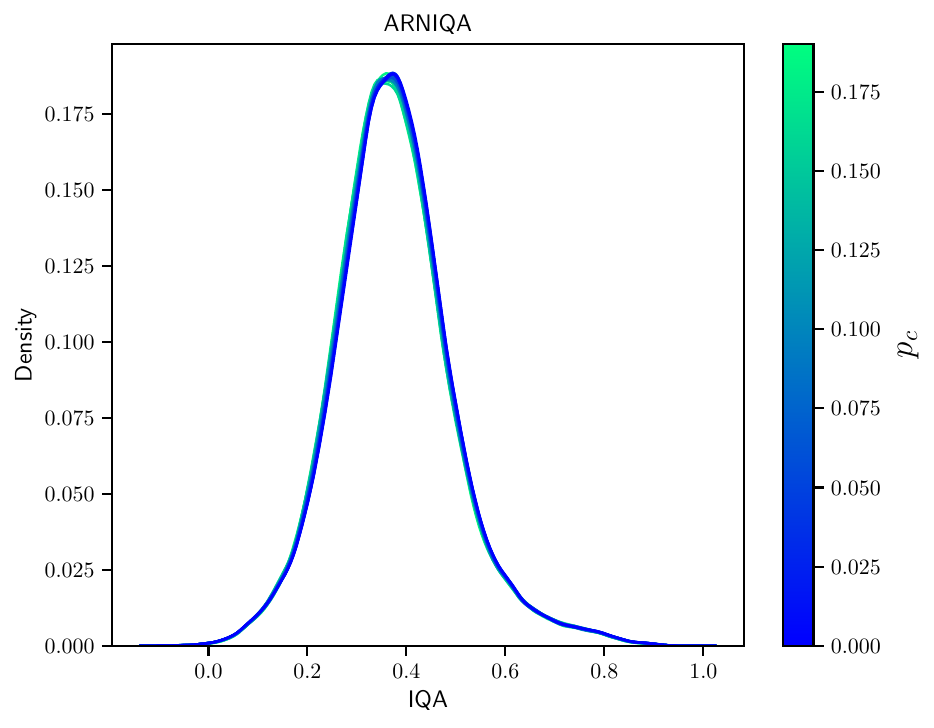}
    \includegraphics[width=0.47\linewidth]{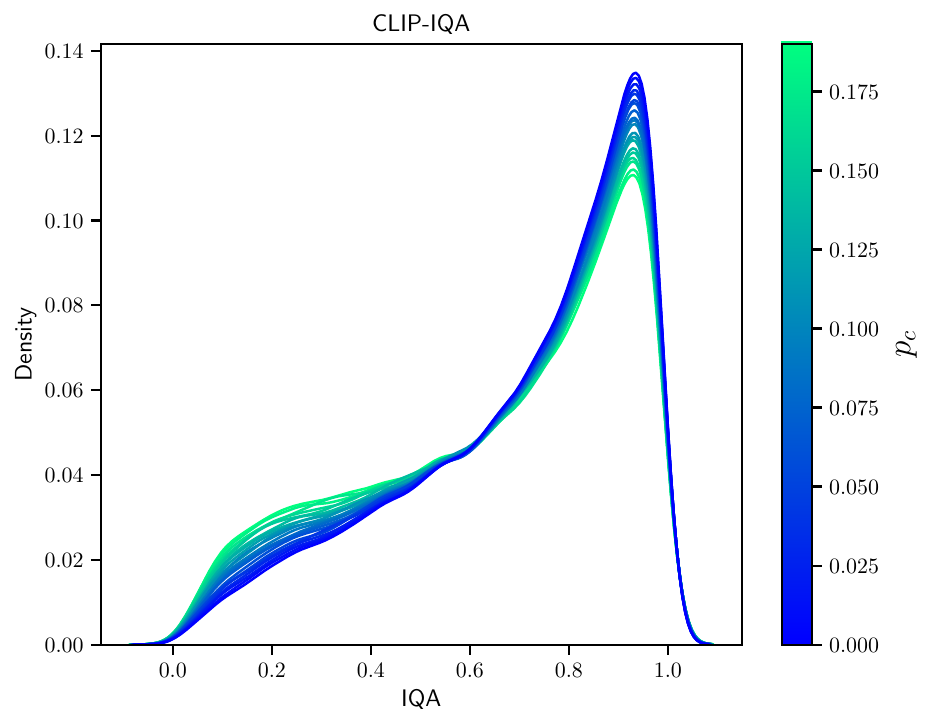} \\
    \includegraphics[width=0.47\linewidth]{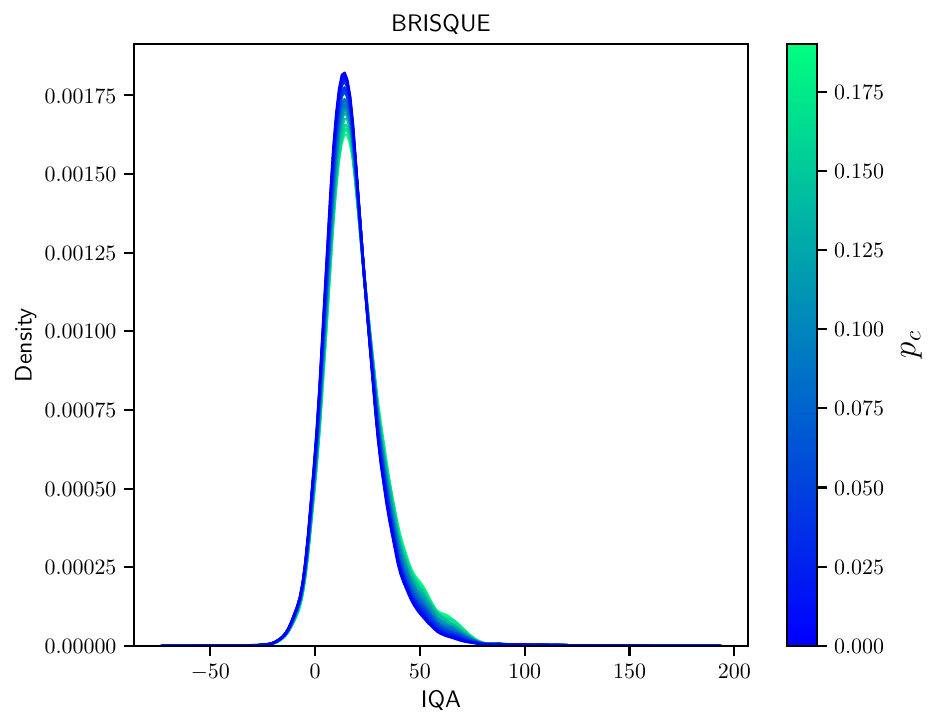}
    \includegraphics[width=0.47\linewidth]{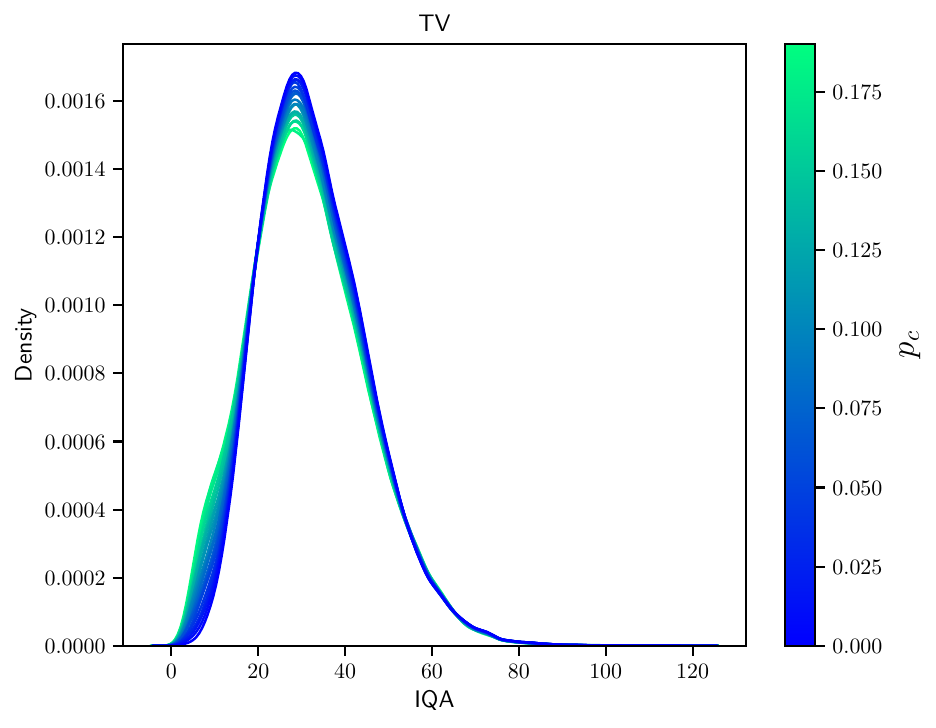} \\
    \includegraphics[width=0.47\linewidth]{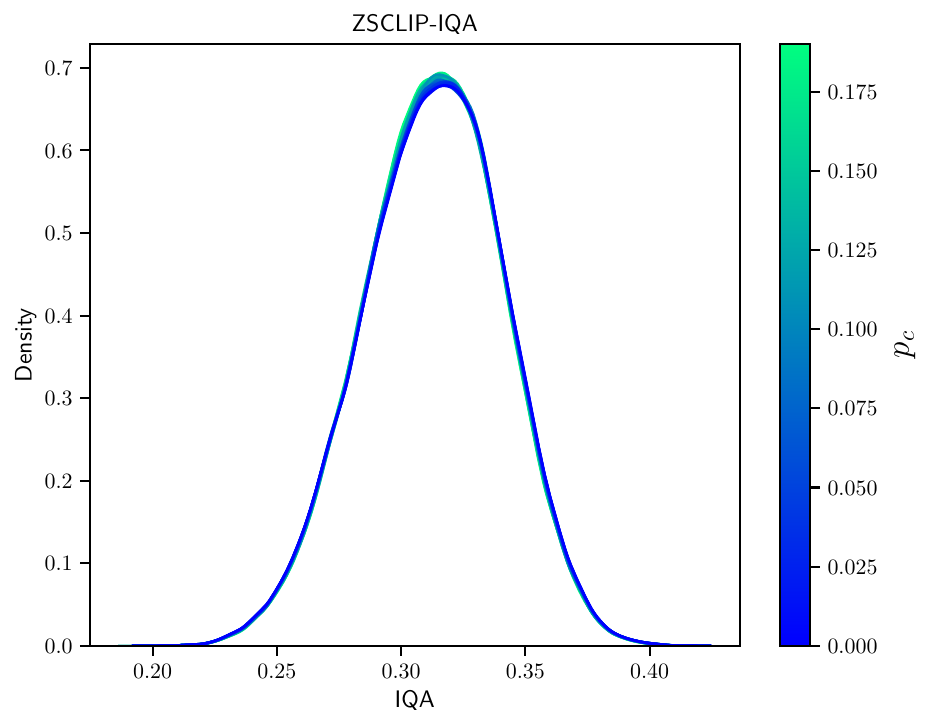}
    \caption{Distribution of IQA for each mildly corrupted variant of IN-val. Line color indicates the likelihood of image corruption $p_c$ for each variant. Note that the amount of similarity/difference in the IQ distribution across variants does not explain the predictability which is determined by the causal DAG such as in \S\ref{sec:causal-iqa}, \S\ref{sec:strong-tg-iqa}, and \S\ref{sec:weak-tg-iqa}.  See Figure~\ref{fig:app-auc-vs-p_corrupt} for predictability results.}
    \label{fig:app-distr-iqa}
\end{figure}

To show that $D1$ is satisfied even in the case of mildly corrupted data, we plot the distributions of $Q$ in Figure~\ref{fig:app-distr-iqa}. Across all variants, even in cases where the likelihood is low, each IQA metric exhibits sensitivity to corruption (\textbf{D1}). 

\begin{figure}[h!]
    \centering
    \includegraphics[width=0.75\linewidth]{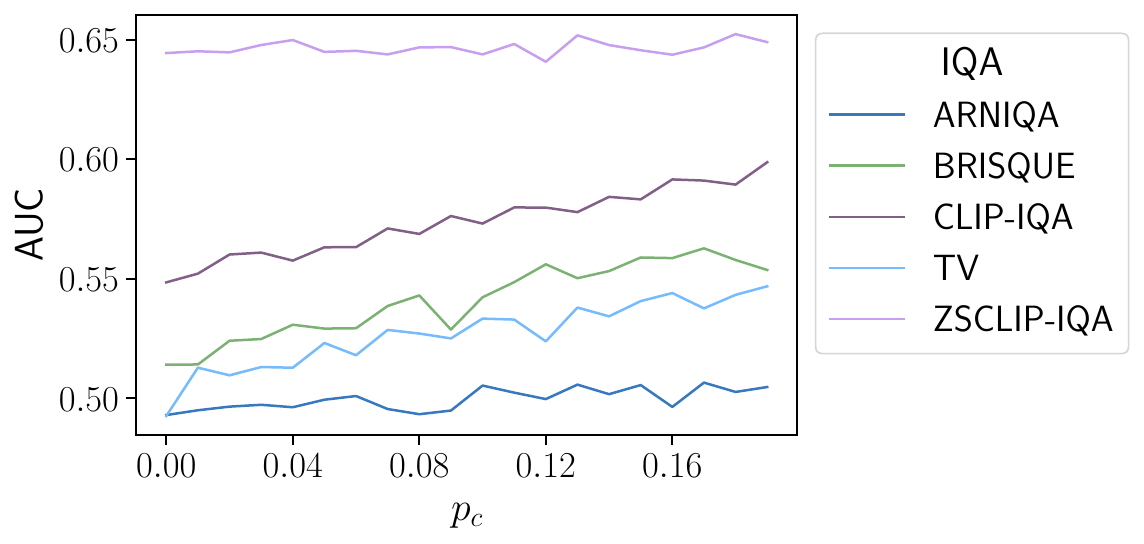} \\
    \includegraphics[width=0.75\linewidth]{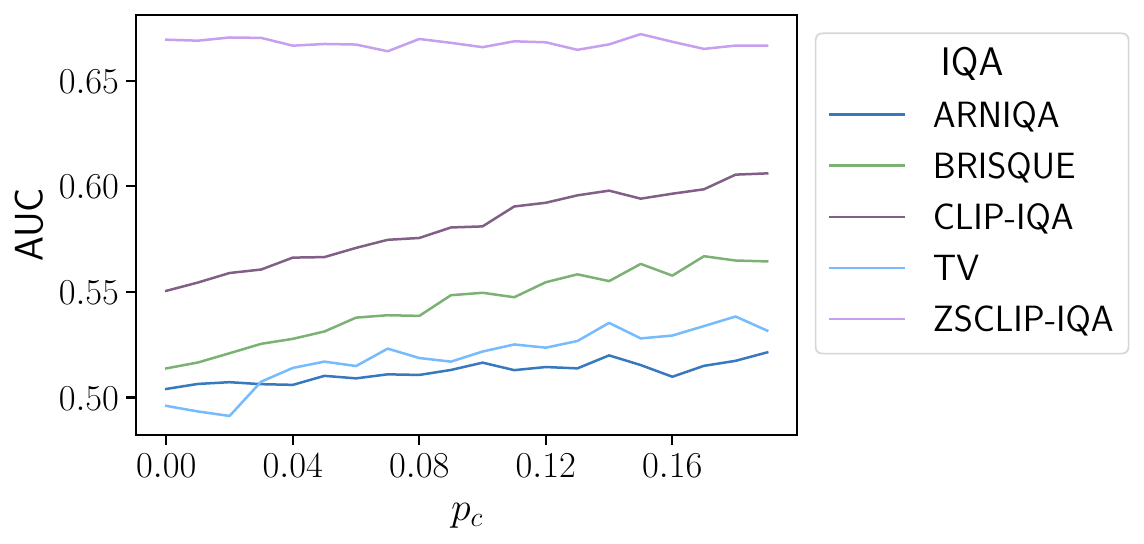} \\
    \includegraphics[width=0.75\linewidth]{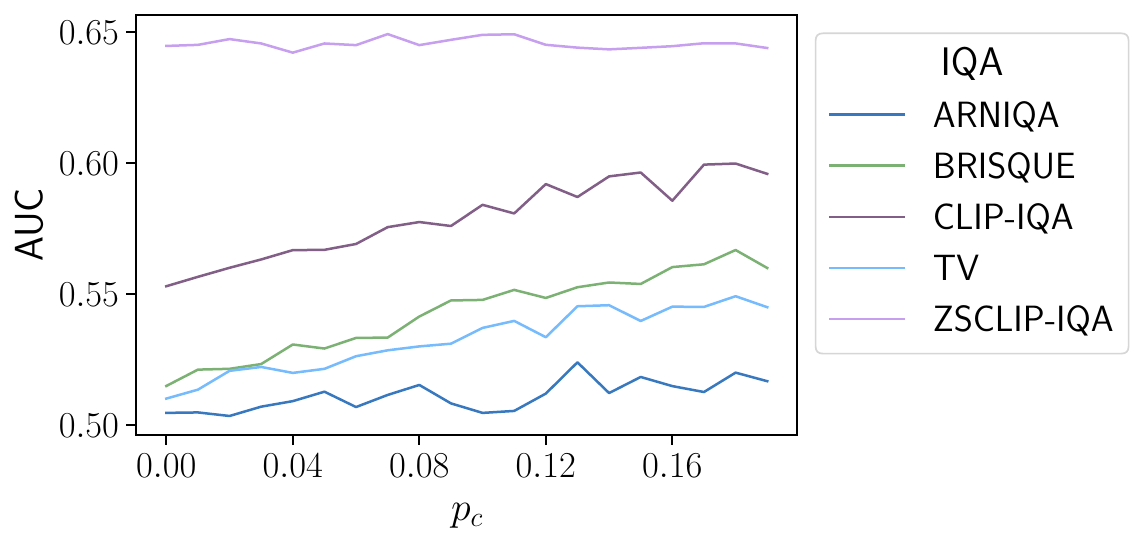} \\
    \caption{AUC vs. $p_c$ where $p_c$ represents the fraction of images in the test set that are mildly corrupted. Results are listed top to bottom: ConvNet-B, ResNet34, Swin-B. Predictability for ZSCLIP-IQA is relatively insensitive to the proportion of corrupted images whereas other metrics only improve as the proportion and diversity of corruptions increases.}
    \label{fig:app-auc-vs-p_corrupt}
\end{figure}

While some IQA metrics are more sensitive to the overall image corruption, this does not necessarily translate to higher predictability.  In fact, ZSCLIP-IQA appears to have smaller differences in IQA distribution across variants compared to other IQA metrics, yet the highest predictability of $M$.  Figure~\ref{fig:app-auc-vs-p_corrupt} shows the predictability of $M$ from $Q$ for variants of the IN-C benchmark created as described in \S\ref{sub:exp3a-mild-corrupt}.  

Results show that weak task-guided IQA metrics are able to achieve higher $AUC$ even when the number of corrupted images in the dataset is low.  In comparison, conventional NR-IQA metrics achieve lower $AUC$ and are more sensitive to the total level of corruption.

\section{Compute resources}
\label{app:compute}
All experiments were run using a single NVIDIA A40 GPU with 48GB of memory. Predictability analysis can be conducted on CPU-only machine with at least 8 cores. 

\end{document}